\documentclass{article} % For LaTeX2e
\usepackage{iclr2024_conference,times}

\usepackage{amsmath,amsfonts,bm}

\def\eqref#1{equation~\ref{#1}}
\def\1{\bm{1}}

\DeclareMathAlphabet{\mathsfit}{\encodingdefault}{\sfdefault}{m}{sl}
\SetMathAlphabet{\mathsfit}{bold}{\encodingdefault}{\sfdefault}{bx}{n}

\usepackage[utf8]{inputenc} % allow utf-8 input
\usepackage[T1]{fontenc}    % use 8-bit T1 fonts
\usepackage{titletoc}
\usepackage{hyperref}       % hyperlinks
\usepackage{url}            % simple URL typesetting
\usepackage{booktabs}       % professional-quality tables
\usepackage{amsfonts}       % blackboard math symbols
\usepackage{nicefrac}       % compact symbols for 1/2, etc.
\usepackage{microtype}      % microtypography
\usepackage{longtable}
\usepackage{listings}
\usepackage{amsmath}
\usepackage{tabularx}

\usepackage{booktabs} % For better spacing in tables
\usepackage{graphicx} % For images and including the red arrow

\renewcommand{\cite}[1]{\citep{#1}}

\definecolor{demphcolor}{RGB}{144,144,144}
\newcommand{\demph}[1]{\textcolor{demphcolor}{#1}}

\usepackage{mdframed}
\newmdenv[
  nobreak=false,
  font=\ttfamily\small,
  linewidth=0.5pt,
  innerleftmargin=10pt,
  innerrightmargin=10pt,
  innertopmargin=10pt,
  innerbottommargin=10pt,
]{monobox}

\definecolor{darkblue}{rgb}{0.0, 0.0, 0.6}
\definecolor{medblue}{rgb}{0.0, 0.0, 0.8}
\definecolor{forestgreen}{rgb}{0.13, 0.55, 0.13}
\definecolor{brickred}{rgb}{0.8, 0.25, 0.33}
\definecolor{darkcyan}{rgb}{0.0, 0.55, 0.55}
\definecolor{gold}{rgb}{0.83, 0.69, 0.22}
\definecolor{magenta}{rgb}{1.0, 0.0, 1.0}

\colorlet{punct}{red!60!black}
\definecolor{background}{HTML}{FFFFFF}  % Here we set the background to white
\definecolor{delim}{RGB}{20,105,176}
\colorlet{numb}{brickred}

\lstdefinelanguage{json}{
    basicstyle=\small\ttfamily,
    numbers=left,
    numberstyle=\scriptsize,
    stepnumber=1,
    numbersep=8pt,
    showstringspaces=false,
    breaklines=true,
    frame=lines,
    backgroundcolor=\color{background},
    literate=
     *{0}{{{\color{numb}0}}}{1}
      {1}{{{\color{numb}1}}}{1}
      {2}{{{\color{numb}2}}}{1}
      {3}{{{\color{numb}3}}}{1}
      {4}{{{\color{numb}4}}}{1}
      {5}{{{\color{numb}5}}}{1}
      {6}{{{\color{numb}6}}}{1}
      {7}{{{\color{numb}7}}}{1}
      {8}{{{\color{numb}8}}}{1}
      {9}{{{\color{numb}9}}}{1}
      {:}{{{\color{punct}{:}}}}{1}
      {,}{{{\color{punct}{,}}}}{1}
      {'}{{{\color{delim}{'}}}}{1}
      {"}{{{\color{delim}{"}}}}{1}
      {\{}{{{\color{delim}{\{}}}}{1}
      {\}}{{{\color{delim}{\}}}}}{1}
      {[}{{{\color{delim}{[}}}}{1}
      {]}{{{\color{delim}{]}}}}{1},
}

\newcommand{\Xmat}{\mathbf{X}}
\newcommand{\Imat}{\mathbf{I}}
\newcommand{\Jmat}{\mathbf{J}}
\newcommand{\Cmat}{\mathbf{C}}

\newcommand{\thetav}{\boldsymbol{\theta}}
\newcommand{\xv}{\boldsymbol{x}}
\definecolor{mygreen}{HTML}{3cb44b}
\newcommand{\PredSty}[1]{\textnormal{\ttfamily\color{forestgreen}#1}\unskip}
\usepackage{graphicx}
\usepackage{tcolorbox}
\newcommand{\VarSty}[1]{\textnormal{\ttfamily\color{darkcyan}#1}\unskip}

\usepackage{booktabs}
\usepackage{multirow}
\usepackage{graphicx}
\usepackage[normalem]{ulem}
\useunder{\uline}{\ul}{}

\usepackage{amsthm}
\usepackage{amsmath}
\usepackage{amsfonts}
\usepackage{xcolor}
\usepackage{mdframed}

\newcommand{\OurModel}{SparklesChat}
\newcommand{\OurEval}{SparklesEval}
\newcommand{\OurData}{SparklesDialogue}
\newcommand{\OurDataCC}{SparklesDialogueCC}
\newcommand{\OurDataVG}{SparklesDialogueVG}

\usepackage{textcomp}  % Required for encoding \textbigcircle
\usepackage{scalerel}  % Required for emoji \scalerel

\def\staricon{\scalerel*{\includegraphics{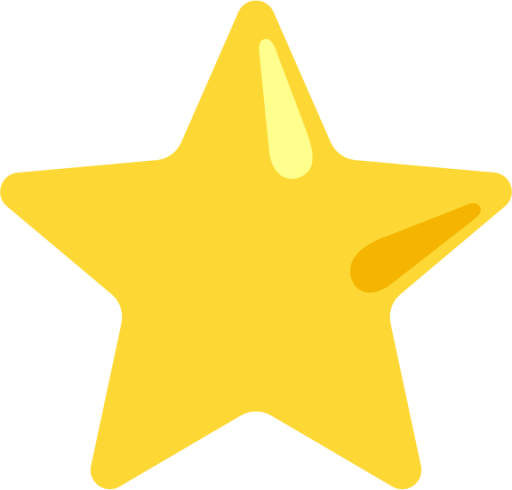}}{\textrm{\textbigcircle}}}

\title{Sparkles: Unlocking Chats Across Multiple Images for Multimodal Instruction-Following Models}

\author{Antiquus S.~Hippocampus, Natalia Cerebro \& Amelie P. Amygdale \thanks{ Use footnote for providing further information
about author (webpage, alternative address)---\emph{not} for acknowledging
funding agencies.  Funding acknowledgements go at the end of the paper.} \\
Department of Computer Science\\
Cranberry-Lemon University\\
Pittsburgh, PA 15213, USA \\
\texttt{\{hippo,brain,jen\}@cs.cranberry-lemon.edu} \\
\And
Ji Q. Ren \& Yevgeny LeNet \\
Department of Computational Neuroscience \\
University of the Witwatersrand \\
Joburg, South Africa \\
\texttt{\{robot,net\}@wits.ac.za} \\
\AND
Coauthor \\
Affiliation \\
Address \\
\texttt{email}
}

\author{%
  Yupan Huang$^{12}$\, Zaiqiao Meng$^{23}$, Fangyu Liu$^{2}$, Yixuan Su$^{2}$, Nigel Collier$^{2}$, Yutong Lu$^{1}$ \\
  $^{1}$Sun Yat-sen University \;\;\;   $^{2}$University of Cambridge \;\;\; $^{3}$University of Glasgow \\
  \texttt{\{huangyp28@mail2,luyutong@mail\}.sysu.edu.cn} \\ \texttt{\{zm324,fl399,ys484,nhc30\}@cam.ac.uk}
}

\iclrfinalcopy % Uncomment for camera-ready version, but NOT for submission.
\begin{document}

\maketitle

\begin{abstract}
Large language models exhibit enhanced zero-shot performance on various tasks when fine-tuned with instruction-following data. Multimodal instruction-following models extend these capabilities by integrating both text and images. However, existing models such as MiniGPT-4 and LLaVA face challenges in maintaining dialogue coherence in scenarios involving multiple images. A primary reason is the lack of a specialized dataset for this critical application. To bridge these gaps, we introduce \textbf{\OurData{}}, the first machine-generated dialogue dataset tailored for word-level interleaved multi-image and text interactions. Furthermore, we construct \textbf{\OurEval{}}, a GPT-assisted benchmark for quantitatively assessing a model's conversational competence across multiple images and dialogue turns. We then present \textbf{\OurModel{}}, a multimodal instruction-following model for open-ended dialogues across multiple images. Our experiments validate the effectiveness of training \OurModel{} with \OurData{} based on MiniGPT-4 and LLaVA-v1.5, which enhances comprehension across multiple images and dialogue turns, and does not compromise single-image understanding capabilities. Qualitative evaluations further demonstrate \OurModel{}'s generality in handling real-world applications. All resources related to this study are publicly available at \url{https://github.com/HYPJUDY/Sparkles}.
\end{abstract}

\begin{figure*}[ht]
\centering
\includegraphics[width=0.98\textwidth]{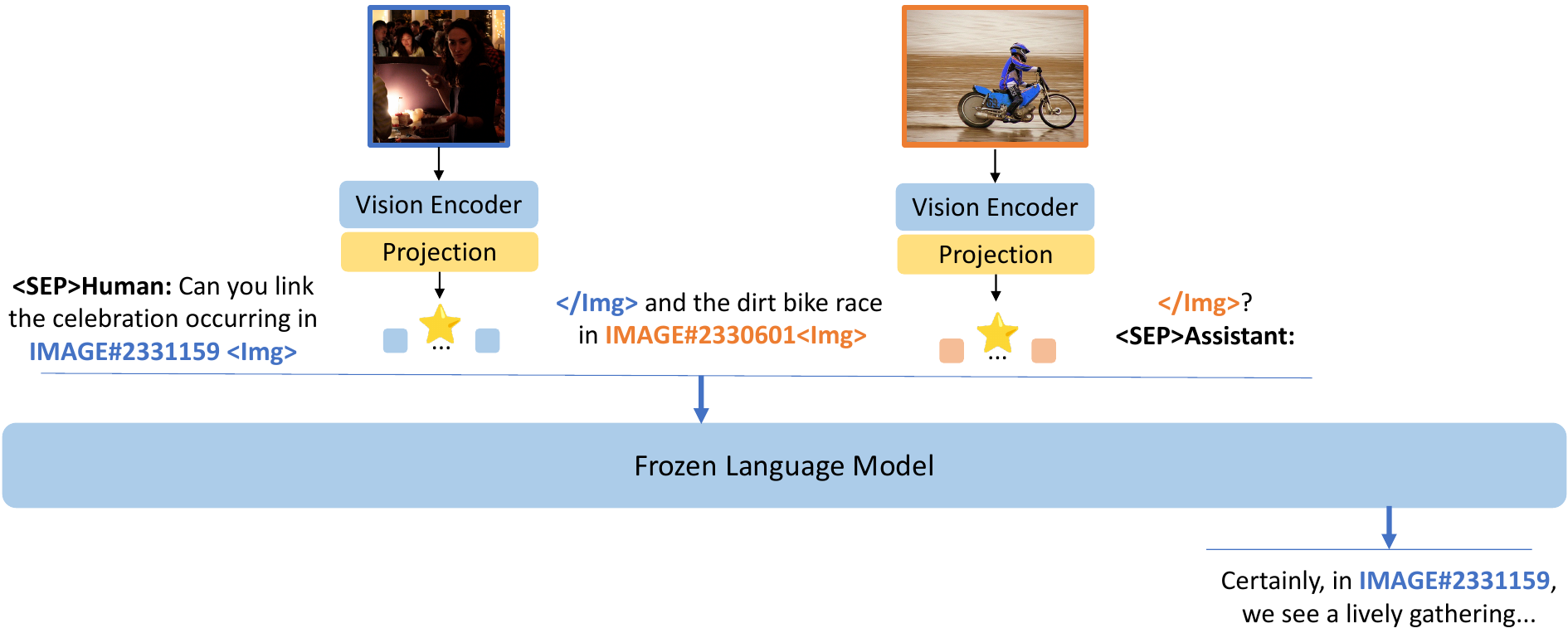}
\caption{
The architecture of \OurModel{}. \OurModel{} integrates multiple images at the word level within the dialogue, facilitating a fine-grained and human-like multimodal interaction.
}
\label{fig:SparklesChat}
\end{figure*}

\begin{figure*}[htbp]
\centering
\includegraphics[width=0.98\textwidth]{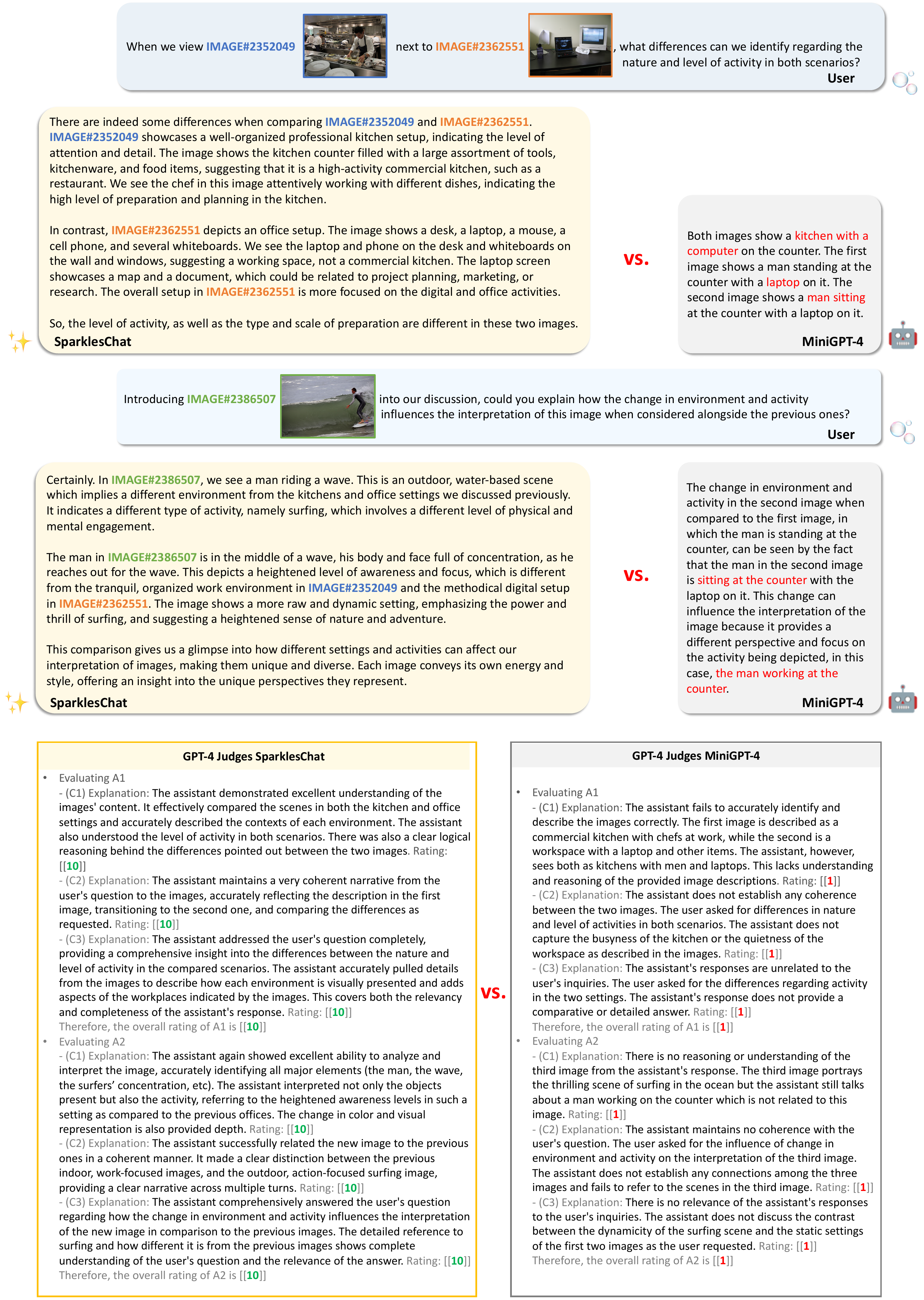}
\caption{{Comparison between our \OurModel{} (left) and MiniGPT-4 (right) on an example from \OurEval{}.} We adapt MiniGPT-4 to accept multiple images as input. \OurModel{} shows conversational competence in open dialogues across three criteria: (C1) image understanding and reasoning, (C2) maintaining cross-image and cross-turn coherence, and (C3) generating relevant and complete responses. In contrast, MiniGPT-4 faces challenges in these aspects, leading to difficulty following user instructions across various images and dialogue turns.}
\label{fig:demo_SparklesEval}
\end{figure*}

\section{Introduction}
\label{sec:intro}
Large language models (LLMs) have shown remarkable progress in zero-shot performance across a variety of tasks when fine-tuned using instruction-following data~\cite{ouyang2022training,openai2023gpt4,llama,vicuna2023,wei2022finetuned,wang2022self,yin2023survey}. In the multimodal domain, instruction-following models such as MiniGPT-4~\cite{zhu2023minigpt4} and LLaVA~\cite{liu2023visual} extend these capabilities by integrating pretrained vision encoders with instruction-following LLMs using projection layers. 
These models learn alignments between individual images and sentences by training on image-text pairs but struggle to capture interactions among multiple images and text. 
This capability is crucial for user-assistant conversations, where users often refer to multiple images with text snippets to convey their instructions in detail. As shown in \autoref{fig:demo_SparklesEval}, MiniGPT-4 mixes up the content of multiple images, fails to establish coherence between images, and consequently falls short in following user instructions during open dialogues.

One key limitation hindering progress in this area is the lack of specialized datasets designed for multimodal dialogues that involve multiple images and fine-grained, word-level text interactions. Existing models such as Flamingo can adapt to various image understanding tasks when prompted with a few relevant examples due to their training on image-text interleaved web data~\cite{alayrac2022flamingo}. However, these models often fall short in following intricate human instructions because they are trained to predict the next word on a large web dataset rather than perform the task the user wants~\cite{ouyang2022training}.

To address these gaps, we introduce \textbf{\OurData{}}, the first machine-generated dialogue dataset designed for word-level interleaved multi-image and text interactions. Notably, \OurData{} was generated by using OpenAI's GPT-4~\cite{openai2023gpt4} to simulate user-assistant conversations with visual capabilities based on detailed image descriptions. By leveraging two different image and description sources, we curated two distinct subsets, namely \OurDataCC{} and \OurDataVG{}, which ensure enhanced robustness and diversity.

Furthermore, we introduce \textbf{\OurEval{}}, a GPT-assisted benchmark to quantitatively evaluate a model's conversational competence in open-ended dialogues across multiple images and dialogue turns. \OurEval{} features a comprehensive and interpretable scoring system based on three criteria: \textit{Image Understanding and Reasoning}, \textit{Cross-Image and Cross-Turn Coherence}, and \textit{Relevance and Completeness of Responses}.

We then present \textbf{\OurModel{}}, a multimodal instruction-following model for open-ended dialogues across multiple images. Unlike previous approaches such as MiniGPT-4
and LLaVA~\cite{liu2023visual,liu2023improved} that take the concatenation of a single image with sentence-level text as input (e.g., "\staricon{} Can you describe this image as detailed as possible?" - where `\staricon{}' denotes a single image), \OurModel{}, as shown in \autoref{fig:SparklesChat}, integrates multiple images at the word level (e.g., "Can you link the celebration occurring in IMAGE\#2331159\staricon{} and the dirt bike race in IMAGE\#2330601\staricon?"). This innovation enables fine-grained integration of multiple images and context tokens, mimicking natural human communication closely.

For quantitative evaluation, we validate the effectiveness of our proposed \OurModel{} and \OurData{} through extensive experiments. These include evaluations for both multi-image and single-image understanding based on the MiniGPT-4 and LLaVA-v1.5 architectures. For multi-image understanding, we perform evaluations on binary image selection on the BISON dataset~\cite{hu2019evaluating} and visual reasoning on the NLVR2 dataset~\cite{suhr2019corpus}. With the BISON dataset, \OurModel{} achieves 10.7\% and 12.6\% improvement in accuracy for MiniGPT-4 and LLaVA-v1.5, respectively. Similarly, on the NLVR2 dataset, \OurModel{} enhances accuracy by 6.7\% and 3.4\% for MiniGPT-4 and LLaVA-v1.5, respectively. In our \OurEval{}, \OurModel{} scores 8.56 out of 10, outperforming the MiniGPT-4 (3.91) and LLaVA-v1.5 (2.75) scores.
Furthermore, our experiments indicate that training \OurModel{} with \OurData{} not only maintains but potentially enhances the single-image understanding capabilities of multimodal instruction-following models on some benchmarks. Qualitative evaluations further demonstrate \OurModel{}'s applicability in real-world scenarios to handle multi-turn dialogues, with each turn involving multiple images.

\begin{figure*}[t]
\centering
\includegraphics[width=0.98\textwidth]{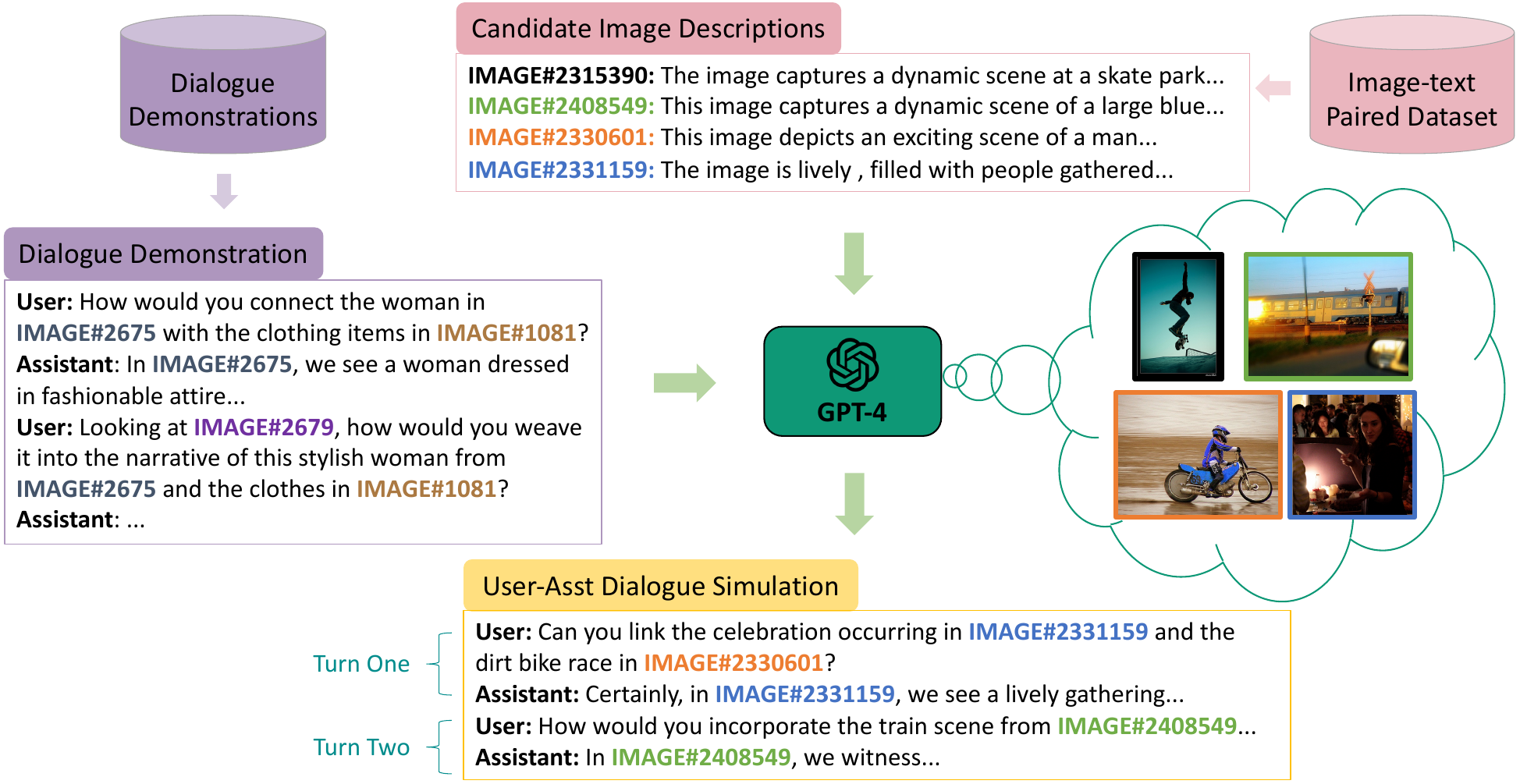}
\caption{The GPT-assisted data construction process. GPT-4 simulates dialogues between a user and an assistant using multiple images. Dialogue Demonstrations act as learning examples for generating well-formatted dialogues, and Candidate Image Descriptions provide a pool of images for discussion. No visual images are sent to GPT-4.}
\label{fig:SparklesDialogue}
\end{figure*}

\section{Related Works}
\label{section:related_works}

We provide a concise summary herein and direct readers to \autoref{appendix:related_works} for a detailed discussion.

\paragraph{Multimodal alignment datasets.}
Various datasets such as Visual Genome~\cite{krishna2017visual} and Conceptual Captions~\cite{sharma2018conceptual} align images with corresponding descriptions, forming the foundation for multimodal alignment. Advancements such as the Common Crawl Interleaved data~\cite{huang2023language} and the Multimodal C4 dataset~\cite{zhu2023multimodal} expand conventional datasets by integrating multiple images and sentences from web corpora. Models including Flamingo~\cite{alayrac2022flamingo} and Kosmos-1~\cite{huang2023language} trained on them can adapt to various tasks using multiple image-text examples. However, they fall short in following intricate instructions as they are trained to predict the next word on a web dataset rather than perform the task the user wants~\cite{ouyang2022training}.

\paragraph{Multimodal dialogue datasets.}
Datasets such as Visual Dialog~\cite{das2017visual} created by crowd workers, and LLaVA data~\cite{liu2023visual} generated by LLMs, focus on image-driven conversations inquiring about image attributes or factual knowledge. Conversely, datasets such as OpenViDial~\cite{meng2020openvidial} and PhotoChat~\cite{zang2021photochat} integrate images within daily human conversations sparsely. Nonetheless, these datasets are not designed for instructive, in-depth multi-image analysis dialogues, posing challenges in dealing with real-world analytical scenarios.

\paragraph{Multimodal instruction tuning.}
Multimodal instruction tuning developed with datasets like MultiInstruct~\cite{xu2022multiinstruct} offering benchmarks for diverse multimodal tasks and models like MiniGPT-4~\cite{zhu2023minigpt4} being fine-tuned on detailed image descriptions to align better with user intentions. Techniques such as LLaVA~\cite{liu2023visual,liu2023improved} and SVIT~\cite{zhao2023svit} leverage LLMs to interpret image annotations and generate instruction-following datasets. Our dataset and model build upon these developments and explore complex interactions between multiple images and word-level textual content.

\begin{figure*}[t]
\centering
	\begin{tabular}{c c}
\includegraphics[height=5.cm]{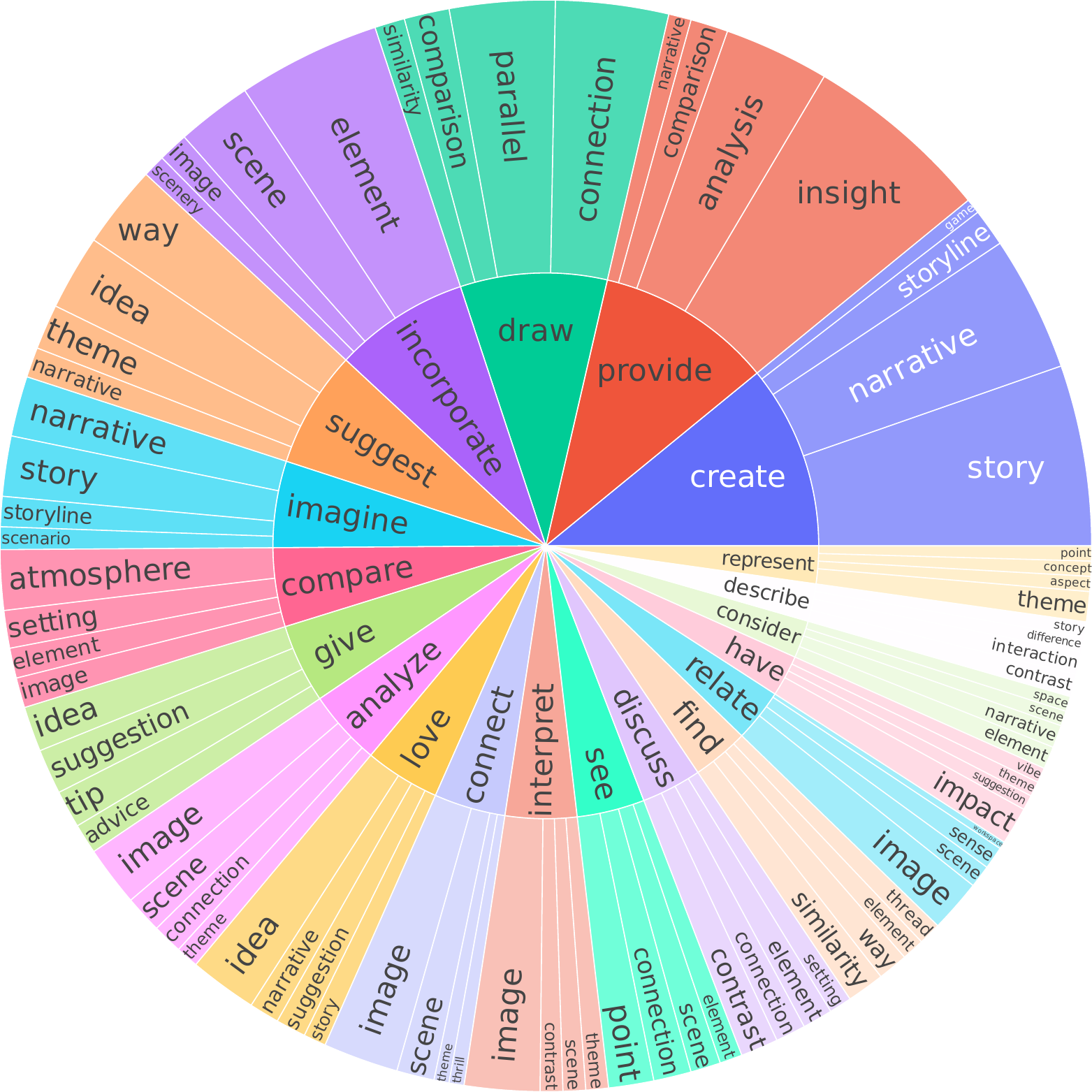} & 
\includegraphics[height=5.cm]{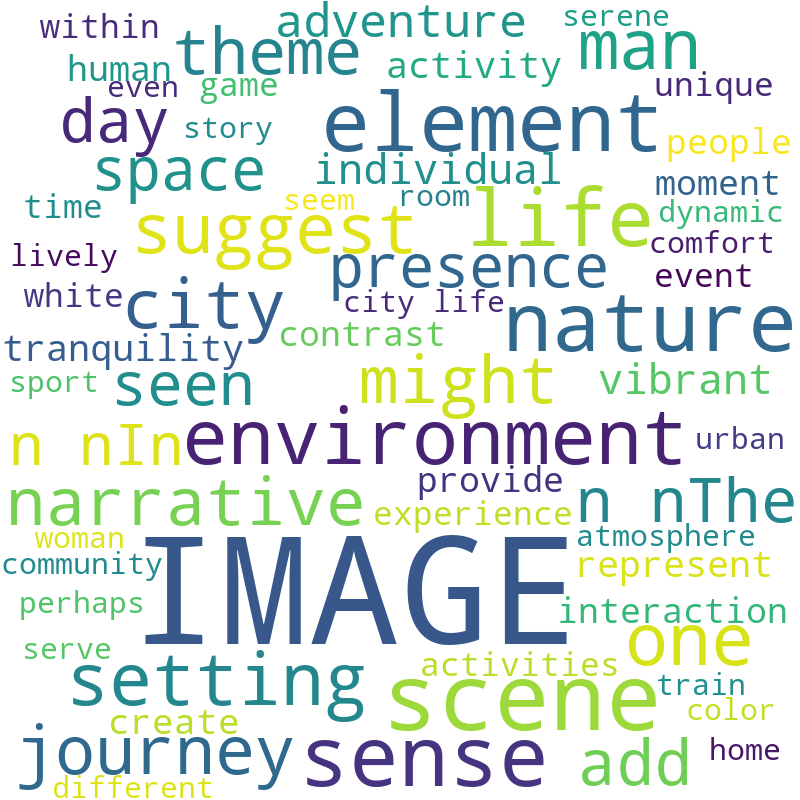}  \\
		(a) Root verb-noun pairs in user messages.
		&
		(b) Word cloud of assistant messages.  \\
\includegraphics[height=4.0cm]{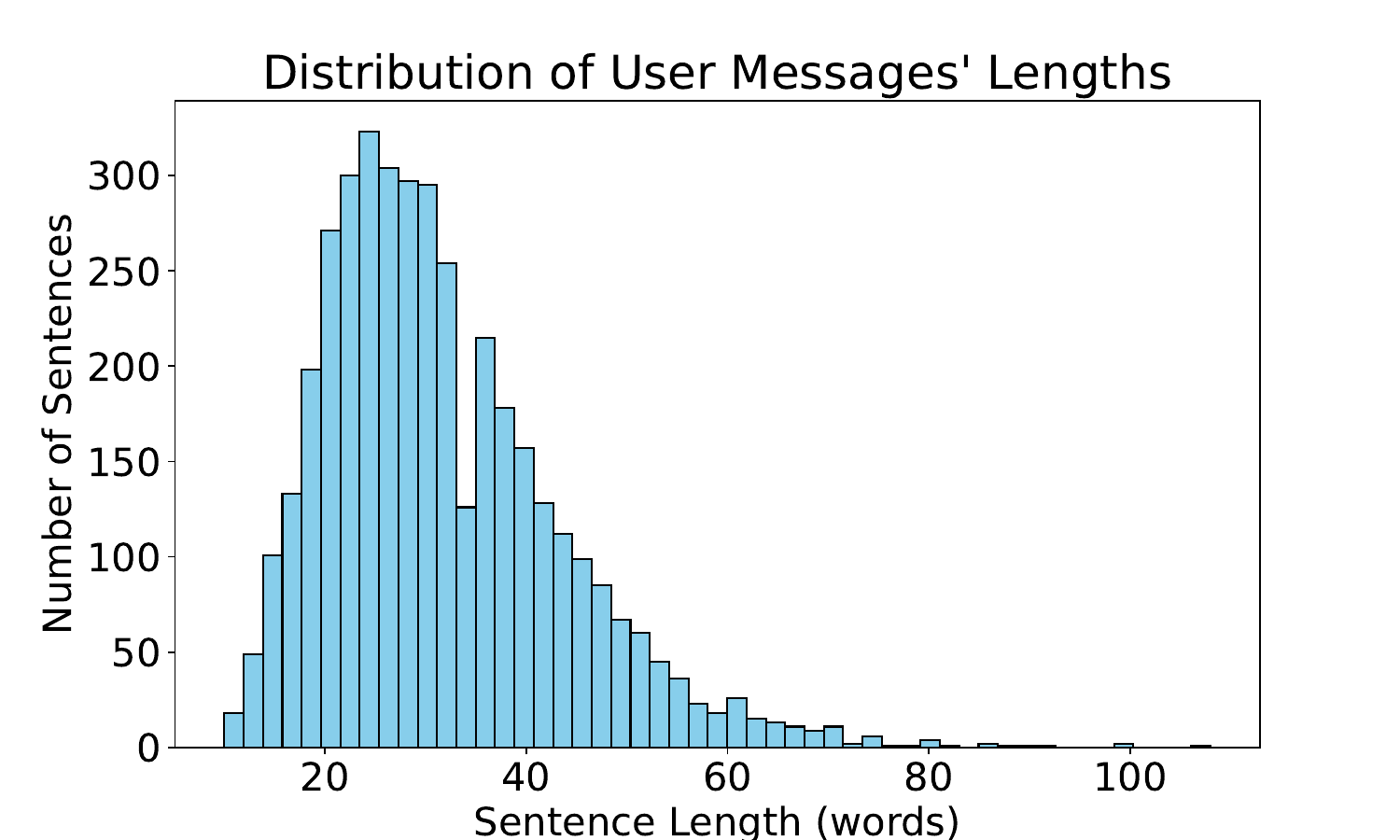} & 
\includegraphics[height=4.0cm]{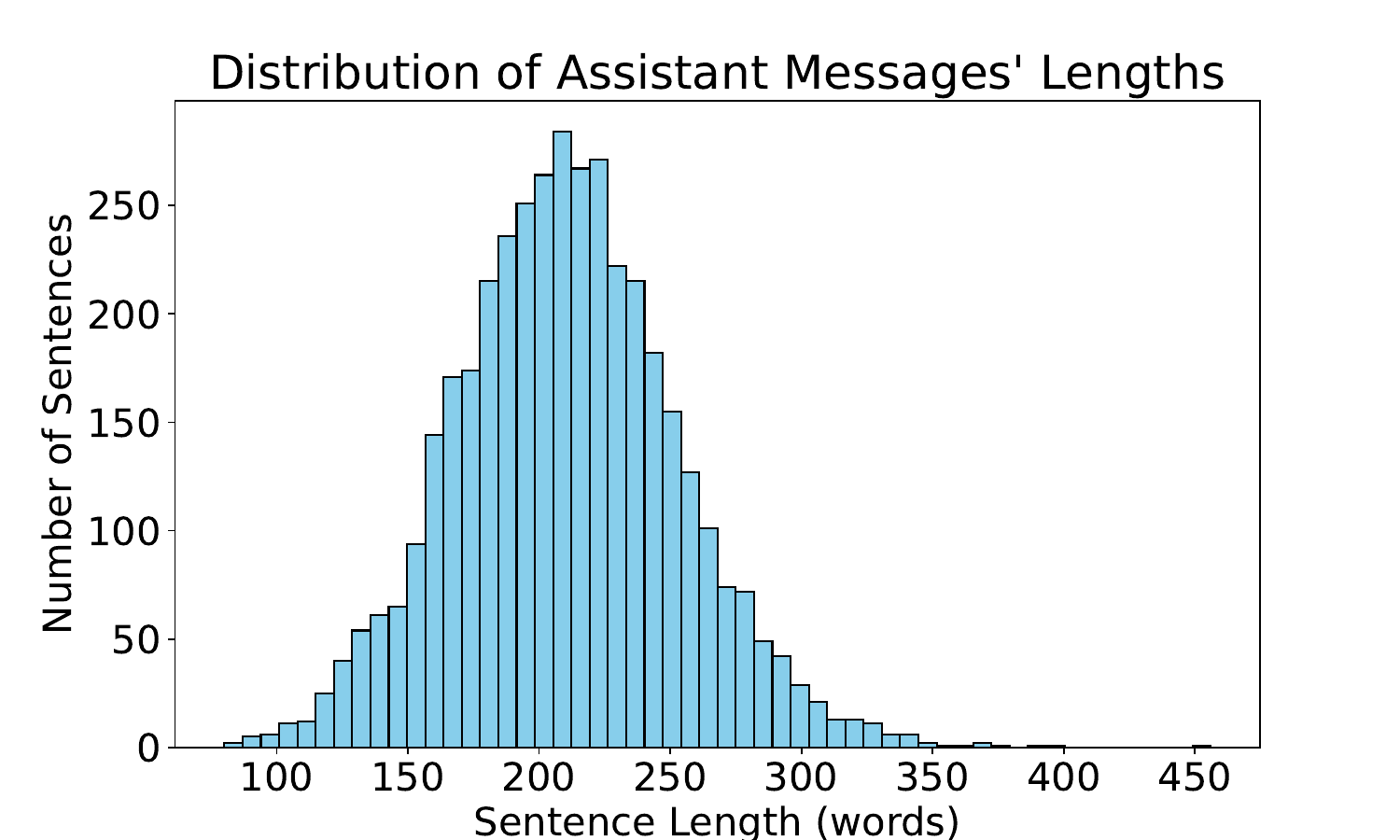}  \\
		(c) Distribution of word lengths in user
		&
		(d) Distribution of word lengths in assistant \\ messages with an average of 31.5 words.&
   messages with an average of 210.5 words. \\
  & 
   \end{tabular}
    \caption{{Characteristics of \OurDataVG{}.}}
    \label{fig:vg_data_stats}
\end{figure*}

\section{\OurData{}}
We introduce \OurData{} to enhance the conversational abilities of multimodal models across multiple images and dialogue turns.

\paragraph{GPT-assisted data construction.}\label{subsection:data_construction}
We aim to construct a multimodal dialogue dataset that offers fine-grained interactions between multiple images and words, mimicking user-assistant conversations. These dialogues should cover real-world concepts, objects, and entities, spanning scenarios that involve generating text materials, seeking advice, guidance, assistance, and much more.
GPT-4 is used as the primary tool for advanced instruction-following ability and a broad knowledge base.
The data collection process is visualized in \autoref{fig:SparklesDialogue}.
We instruct GPT-4 to simulate realistic and diverse dialogues between a user and an assistant discussing multiple images through a sequence of turns.
Initially, the user gives a reasonable and creative message regarding some images, to which the assistant provides a detailed answer that includes comprehensive reasoning regarding the visual content. Subsequent turns introduce new images for further discussion, ensuring responses are helpful with comprehensive reasoning to better align with human preferences.
For prompt templates and examples, refer to \autoref{appendix:gpt_assisted_dialogue_generation}.

\texttt{Dialogue Demonstration} and \texttt{Candidate Image Descriptions} are crucial components in this process.
\texttt{Dialogue Demonstrations} provide GPT-4 with examples to guide the generation of well-formatted and diverse responses. We curated hundreds of such dialogues and manually checked their quality. A small subset of these dialogues is randomly selected as demonstrations for every dialogue generation task.
{\texttt{Candidate Image Descriptions}} serves as a candidate pool for relevant image selection. 
Given that the existing multimodal models like GPT-4-Vision have limitations in accurate image understanding such as inaccurate spatial reasoning and counting, we represent image content with detailed descriptions sourced from various image annotations such as image captions, bounding boxes, and region descriptions~\cite{zhu2023minigpt4,zhao2023svit,liu2023visual}.
For each dialogue generation instance, candidate images are randomly selected from an image-text paired dataset, the data sources of which are provided following.

\begin{table*}[t]
\small
\caption{Statistics of \OurData{} and \OurEval{}.}
\tabcolsep=0.06cm
\centering
\begin{tabular}{cccccccc}
\toprule
\textbf{Dataset Name} &
  \textbf{\begin{tabular}[c]{@{}c@{}}Image\\ Source\end{tabular}} &
  \textbf{\begin{tabular}[c]{@{}c@{}}Caption\\ Source\end{tabular}} &
  \textbf{\#Dialogue} &
  \textbf{\begin{tabular}[c]{@{}c@{}}\#Image\\ Turn one\end{tabular}} &
  \textbf{\begin{tabular}[c]{@{}c@{}}\#Image\\ Turn two\end{tabular}} &
  \multicolumn{2}{c}{\textbf{\#Unique/Total Image}} \\ \midrule
\multirow{3}{*}{\OurDataCC{}} & \multirow{3}{*}{CC} & \multirow{3}{*}{MiniGPT-4} & 1,653 & 1 & 1 & 2,067/3,306 & \multirow{3}{*}{3,373/12,979} \\
                                    &                     &                            & 1,799 & 2 & 1 & 2,642/5,397 &                               \\
                                    &                     &                            & 1,069 & 3 & 1 & 2,408/4,276 &                               \\ \hline
\multirow{2}{*}{\OurDataVG{}} & \multirow{2}{*}{VG} & \multirow{2}{*}{SVIT}      & 1,000 & 2 & 1 & 3,000/3,000 & \multirow{2}{*}{7,000/7,000}  \\
                                    &                     &                            & 1,000 & 3 & 1 & 4,000/4,000 &                               \\ \hline
\multirow{3}{*}{\OurEval{}} & \multirow{3}{*}{VG} & \multirow{3}{*}{SVIT}      & 50    & 2 & 1 & 150/150     & \multirow{3}{*}{550/550}      \\
                                    &                     &                            & 50    & 2 & 2 & 200/200     &                               \\
                                    &                     &                            & 50    & 3 & 1 & 200/200     &                               \\ \bottomrule
\end{tabular}
\label{tab:data_statistics}
\end{table*}

\paragraph{Data sources and subsets.}
Following the above data construction process, we construct {\OurData{}}, the first machine-generated multimodal dialogue dataset with fine-grained interactions between multiple images and words. In particular, {\OurData{}} consists of two subsets: \OurDataCC{} and \OurDataVG{}, which were generated based on two different image sources, namely Conceptual Captions (CC)~\cite{sharma2018conceptual} and Visual Genome (VG)~\cite{krishna2017visual} respectively.
\OurDataVG{} is of high quality as the VG image descriptions generated by GPT-4 benefit from human-annotated captions, objects, and regions~\cite{zhao2023svit}. On the other hand, \OurDataCC{} enriches \OurData{} by drawing from a more extensive set of images -- 3.3 million in CC compared to 0.1 million in VG. However, the CC image descriptions are generated by a multimodal model with image features but not human annotations and are more prone to object hallucination issues~\cite{zhu2023minigpt4}. These different sources ensure enhanced robustness and diversity across different downstream applications. Our ablation study elaborated in \autoref{subsection:ablation_effect_subset} demonstrates that combining these two subsets improves models' capacity for understanding and reasoning across images and text.

\paragraph{Statistics and characteristics.}
\label{subsec:statistics_and_characteristics}
\autoref{tab:data_statistics} presents the statistics for \OurData{}. Specifically, \OurDataCC{} comprises 4.5K dialogues, each consisting of at least two images spanning two conversational turns. \OurDataVG{} includes 2K dialogues, each with at least three distinct images across two turns.
\autoref{fig:vg_data_stats} shows the characteristics of our dataset using \OurDataVG{} as a representative subset. 
The visualization of root verb-noun pairs indicates a wide range of user queries, from generating text materials to seeking advice or discussing image relationships, such as comparison and connection. The word cloud reveals that dialogues span various real-world topics, including the environment, nature, life, cities, etc. 
The high average word count in assistant messages suggests that the responses in \OurData{} are thorough and detailed.
For details on the characteristics of \OurDataCC{} and the method for extracting root verb-noun pairs and their visualization based on image count in each turn, please see \autoref{appendix:turn_level_verb_noun_distribution}.

\section{\OurEval{}}
While previous research, such as visual storytelling, has leaned toward human evaluations as superior to quantitative measures, these evaluations are often subjective, costly, and time-consuming~\cite{huang2016visual}. Inspired by the consistency of recent LLMs with human assessment in evaluating output quality~\cite{zheng2023judging}, we developed \OurEval{}, a GPT-assisted benchmark to quantitatively assess a model's conversational competence across multiple images and dialogue turns. 

\OurEval{} evaluates dialogues that include questions from the benchmark and the model's generated responses, considering both the current question and any preceding dialogue history.
In this evaluation, a judge model (e.g., GPT-4) is presented with the complete dialogue but is only required to assess the model-generated responses. Image descriptions related to the dialogue are provided to support the assessment. Each assessment is based on three distinct criteria: {Image Understanding and Reasoning}, {Cross-Image and Cross-Turn Coherence}, and {Relevance and Completeness of Responses}, with reasons and ratings on a scale of 1 to 10 for each criterion. 
Moreover, we calculate average scores for each dialogue turn and an overall score derived from the average of these turn-specific scores.
The evaluation prompt and score computation process are elaborated in \autoref{appendix:sparkleseval}.

Our evaluation approach differs from prior GPT-assisted evaluations in two aspects.
First, it employs a combined score for a more comprehensive and interpretable assessment instead of a singular one~\cite{liu2023visual}.
Second, it is less biased and more efficient by assessing a single dialogue per prompt rather than contrasting multiple dialogues within one prompt~\cite{zheng2023judging}. Our approach eliminates position bias - the potential favor to certain positions when multiple dialogues are assessed within a prompt~\cite{zheng2023judging}. It enhances efficiency by avoiding the recalculation of combined scores for multiple dialogues.

\autoref{tab:data_statistics} provides the data statistics for \OurEval{}. \OurEval{} emphasizes more on accuracy and is thus constructed using the detailed SVIT image descriptions sourced from human annotations~\cite{zhao2023svit}. To encourage diversity, \OurEval{} was curated by analyzing the verb-noun distribution in user questions and selecting those that appear only once. \OurEval{} includes 150 dialogues, with one-third containing two images in the first and second conversational turns.

\begin{table*}[t]
  \small
  \caption{Architectures and instruction-tuning data comparisons for MiniGPT-4 and LLaVA-v1.5.}
  \label{tab:compare_minigpt4_llava}
\tabcolsep=0.1cm
  \centering
\begin{tabular}{lllll}
\toprule
\textbf{Model} & \textbf{Res.} & \textbf{Vision Encoder} & \textbf{Language Decoder} & \textbf{Instruction-tuning Data (size)}                                 \\
\midrule
MiniGPT-4      & 224                 & EVA-ViT-G               & Vicuna-v0-7B            & Description (3.5K)                                      \\
LLaVA-v1.5     & 336                 & CLIP-ViT-L              & Vicuna-v1.5-7B          & Description, reasoning, conversation, VQA (665K)\\
\bottomrule
\end{tabular}
\end{table*}

\begin{table*}[ht]
  \small
  \caption{Comparison of model performance on multi-image and single-image understanding benchmarks. \OurModel{}, adaptable to architectures like MiniGPT-4 or LLaVA-v1.5, supports multiple image inputs and benefits from training with \OurData{}.
  Scores for \OurEval{} range from 1 to 10, while other benchmarks use accuracy as the metric. LLaVA data includes a mix of description, reasoning, conversation, and VQA data.}
  \label{tab:compare_models}
\tabcolsep=0.1cm
  \centering
\begin{tabular}{llllllll}
\toprule
\multirow{2}{*}{\textbf{Architecture}} &
  \multirow{2}{*}{\textbf{\begin{tabular}[c]{@{}c@{}}Instruction-tuning Data \end{tabular}}} &
  \multicolumn{3}{c}{\textbf{Multi-image Understanding}} &
  \multicolumn{3}{c}{\textbf{Single-image Understanding}} \\
  \cmidrule(lr){3-5} \cmidrule(lr){6-8} 
 &
   &
  {BISON} &
  {NLVR2} &
  {\begin{tabular}[c]{@{}c@{}}Sparkles\\ Eval\end{tabular}} &
  {MMMU} &
  {\begin{tabular}[c]{@{}c@{}}Science\\QA\end{tabular}} &
  {\begin{tabular}[c]{@{}c@{}}Hallusion\\ Bench\end{tabular}}\\ 
  \midrule
\multirow{2}{*}{MiniGPT-4} &
  Description (3.5K) &
  46.0 &
  51.3 &
  3.91 &
  23.6 &
  39.6 &
  52.4 \\
 &
  \textbf{SparklesDialogue (6.5K)} &
  {56.7} \textcolor{red}{\scriptsize{(+10.7)}} &
  {58.0} \textcolor{red}{\scriptsize{(+6.7)}} &
  {8.56} \textcolor{red}{\scriptsize{(+4.7)}} &
  {28.4} \textcolor{red}{\scriptsize{(+4.8)}} &
  {42.4} \textcolor{red}{\scriptsize{(+2.8)}} &
  {54.4} \textcolor{red}{\scriptsize{(+2.0)}} \\
  \hline
\multirow{2}{*}{LLaVA-v1.5} &
  Mixture (665K) &
  52.7 &
  53.3 &
  2.75 &
  {36.2} &
  {68.9} &
  48.3 \\
 &
  \textbf{SparklesDialogue (6.5K)} &
  {65.3} \textcolor{red}{\scriptsize{(+12.6)}} &
  {56.7} \textcolor{red}{\scriptsize{(+3.4)}} &
  {7.93} \textcolor{red}{\scriptsize{(+5.2)}} &
  35.1 \textcolor{blue}{\scriptsize{(-1.1)}} &
  67.5 \textcolor{blue}{\scriptsize{(-1.4)}} &
  {49.5} \textcolor{red}{\scriptsize{(+1.2)}} \\
\bottomrule
\end{tabular}
\end{table*}

\section{\OurModel{}}
We present a multimodal instruction-following model \OurModel{} to foster interactions between users and AI assistants across multiple images and illustrate the framework in \autoref{fig:SparklesChat}.

\paragraph{Architecture.}
\OurModel{} can be based on different architectures such as MiniGPT-4 or LLaVA, which connects a pretrained vision encoder and a pretrained language decoder with a projection layer~\cite{zhu2023minigpt4,liu2023visual,liu2023improved}, as compared in \autoref{tab:compare_minigpt4_llava}.
For the original MiniGPT-4 or LLaVA, the input to the language model is a single image representation followed by a sentence embedding of the image description.
In \OurModel{}, image representations of different images are embedded between text according to their positions in dialogues.
More details are in \autoref{appendix:implementation_details}.

\paragraph{Instruction-tuning.} We simplify the representation of a $T$-turn dialogue $\Xmat^i$ into question-answer pairs for each turn. Training samples are constructed by sequencing these pairs with a predefined system prompt. The prompt $\Xmat_{\texttt{prompt}}^{i, t}$ and response $\Xmat_{\texttt{response}}^{i, t}$ at turn $t$ are formulated to incorporate the system prompt and the dialogue content up to that turn. The model is trained using an auto-regressive training objective, focusing on predicting the target responses based on the prompts. A detailed description is available in the \autoref{appendix:architecutre_training_details}.

\section{Experiments}
\subsection{Comparison of model performance}
\autoref{tab:compare_models} compares the performance of MiniGPT-4, LLaVA-v1.5, and \OurModel{} on multi-image and single-image understanding. 
\OurModel{} is built upon MiniGPT-4 and LLaVA-v1.5 by supporting multi-image training on SparklesDialogue.
Please find more model comparisons in \autoref{appendix:detail_model_compare_multiple_image} and experimental details in \autoref{appendix:implementation_details}.

\paragraph{Multi-image understanding.}
We evaluate models' conversational competence on \OurEval{} and zero-shot understanding and reasoning across images through binary image selection on BISON and visual reasoning with natural language on NLVR2~\cite{hu2019evaluating,suhr2019corpus}. 
For a comprehensive understanding of evaluation protocol and prompt design, please refer to \autoref{appendix:evaluation_vltasks}.

MiniGPT-4 provides baseline results on these tasks.
\OurModel{}, when adopting the MiniGPT-4 architecture and training on our \OurData{}, outperforms MiniGPT-4 in three evaluation sets involving multiple images. Specifically, \OurModel{} improves accuracies of 10.7\% and 6.7\% on BISON and NLVR2, respectively, reflecting its efficacy in handling tasks that require fine-grained visual grounding and compositional visual reasoning over two images. Moreover, \OurModel{} excels in the \OurEval{} benchmark, scoring 8.56 out of 10. 
Side-by-side comparisons of example outputs for \OurModel{} and MiniGPT-4 are in \autoref{fig:demo_SparklesEval} and \autoref{appendix:evaluation_vltasks}.

LLaVA-v1.5 has the advantages of higher image resolution and a larger training set than MiniGPT-4, and outperforms MiniGPT-4 on BISON and NLVR2. However, LLaVA-v1.5 shows weaker results on SparklesEval, which may be due to its training data primarily focusing on closed-set multimodal tasks such as VQA, TextCaps, and RefCOCO, while lacking in open-ended dialogue training. 
After fine-tuning with \OurData{} using the low-resource technique LoRA, \OurModel{} based on LLaVA-v1.5 improved in open-ended dialogue tasks and enhanced BISON and NLVR2 as well. These results validate the adaptability of our method in unlocking chats across multiple images for multimodal instruction-following models with minimal additional training cost.

\paragraph{Single image understanding.}
We evaluate models on MMMU validation split~\cite{yue2023mmmu}, ScienceQA test split~\cite{lu2022learn}, and HallusionBench~\cite{guan2023hallusionbench} with OpenCompass-VLMEvalKit~\cite{2023opencompass}. 
Models utilizing the LLaVA-v1.5 architecture outperform those based on MiniGPT-4, which is attributable to LLaVA-v1.5's higher image resolution and its training on more comprehensive instruction-following data.
Incorporation of the SparklesDialogue training data further refines performance, with MiniGPT-4 showing consistent improvements across benchmarks, while LLaVA-v1.5 exhibits a marginal reduction in some scores. This suggests that while the SparklesDialogue data enhances multi-image comprehension, it does not compromise and may even augment single-image understanding in multimodal models.

\begin{table}[t]
  \caption{Ablation studies analyzing the impact of dialogue turn ratios and subsets from \OurData{} on training \OurModel{} for multi-image understanding.
  The evaluation metric is accuracy for BISON and NLVR2; \OurEval{} is rated 1-10.
  }
  \label{tab:ablation}
\tabcolsep=0.1cm
  \centering
\begin{tabular}{ccccccc}
\toprule
\textbf{Instruction Data} &
  \textbf{\begin{tabular}[c]{@{}c@{}}Turns\\ Ratio\end{tabular}} &
  \textbf{BISON} &
  \textbf{NLVR2} &
  \textbf{\begin{tabular}[c]{@{}c@{}}Sparkles\\ Eval\end{tabular}} \\
                                          \midrule
\multirow{6}{*}{\begin{tabular}[c]{@{}c@{}}SparklesDialogue\\ (CC+VG)\end{tabular}} 
  & 1:0 & \underline{57.3} & \underline{55.3} & 8.50 \\
  & 0:1 & 50.7 & 46.7 & 8.24 \\ 
  & 1:1 & \textbf{59.3} & 51.3 & \textbf{8.73} \\
  & 1:2 & 49.3 & 51.3 & 8.43 \\
  & \textbf{2:1} & 56.7 & \textbf{58.0} & 8.56 \\
  & 3:1 & 50.7 & 48.7 & 8.45 \\ \hline
SparklesDialogueCC & 2:1 & 44.7 & 53.3 & 8.18  \\
SparklesDialogueVG & 2:1 & 54.7 & 52.0 & \underline{8.59} \\ \bottomrule
\end{tabular}
\end{table}

\subsection{Ablation Studies}\label{subsection:ablation_studies}
We investigate the impact of dialogue turn ratios and subsets from \OurData{} on \OurModel{}'s performance and show results in \autoref{tab:ablation}.
For comprehensive scores from \OurEval{} across dialogue turns and criteria, see \autoref{section:ablation_studies_detail}.
\vspace{-2mm}

\paragraph{Effect of dialogue turns in \OurData{}.}\label{subsection:ablation_effect_turns}
We first train models with individual dialogue turns. The model trained solely on the first turn (turns ratio ‘1:0’) outperforms the one trained on the second turn (turns ratio ‘0:1’).
This could stem from the extended prompts in the second turn, which includes the content of the first turn, thus deviating from the short prompt format favored by many end tasks.
A balanced sampling of dialogue turns, indicated by the turn ratio ‘1:1’, yields the highest performance on BISON and \OurEval{}, albeit with a decrease in performance on NLVR2.
An increase in the sampling ratios of the second turn data (turns ratio ‘1:2’) predictably results in a performance drop.
Therefore, we increase the sampling ratio of the first-turn data to enhance performance. We finally adopt a 2:1 ratio for the first turn to the second turn as our default setting as it achieves balanced good performance across all benchmarks.

\paragraph{Effect of subsets of \OurData{}.}\label{subsection:ablation_effect_subset}
Our analysis extended to training the model on two distinct subsets of \OurData{}: \OurDataCC{} and \OurDataVG{}. Training on \OurDataVG{} outperformed \OurDataCC{} in BISON and \OurEval{} assessments, with similar performance observed on the NLVR2 test.
The superior performance of \OurDataVG{} can be attributed to its higher-quality, human-annotated data, as discussed in \autoref{subsec:statistics_and_characteristics}. Notably, \OurDataVG{} and \OurEval{} share image and caption sources, likely contributing to \OurDataVG{}'s enhanced \OurEval{} scores.
Combining both subsets yields higher or comparable performance than using either subset alone. 
This suggests that combining \OurDataVG{}'s high-quality data and \OurDataCC{}'s diverse data results in a more robust and versatile dataset for enhancing models' capabilities in image-text understanding.

\subsection{Demonstrations and applications}
We conducted qualitative demonstrations to showcase \OurModel{}'s broad applications in free-form scenarios by asking questions such as: 
``Create a story that takes place in \staricon for the characters depicted in \staricon.'', ``Imagine a dialogue between Harry Potter and \staricon that takes place in the scene of \staricon.'', ``Create a song where the scene twists from \staricon to \staricon.'', ``Create a title for this song that takes inspiration from \staricon.''.
These scenarios cover dialogues involving two or three-turn dialogues, {with each turn involving images from one to five.}
The visualization {and analysis} of results are shown in \autoref{appendix:demo}.

\vspace{-1mm}
\section{Conclusion and Limitations}
{In conclusion}, this work unlocks multimodal instruction-following models' capabilities in open-ended dialogues involving multiple images. We introduced {\OurData{}}, the first machine-generated dialogue dataset tailored for multi-image and word-level text interactions. Furthermore, we proposed {\OurEval{}}, a specialized benchmark for quantitatively assessing a model's multimodal conversational competence. We also presented \OurModel{}, a model designed to handle word-level text interactions in a multimodal context, offering natural conversational flow and direct context awareness. 
Experimental results demonstrated the effectiveness of training \OurModel{} with \OurData{} based on MiniGPT-4 and LLaVA-v1.5 architectures in both multi-image and single-image understanding benchmarks.
We also conducted qualitative demonstrations to showcase the model's broad applications in free-form scenarios.

We discuss some {limitations} of this work to inspire {future research} in this field.
First, \OurModel{} shares common drawbacks with large language models, such as being out-of-date in its knowledge, sometimes providing inaccurate information, and having limited context length and inference speed~\cite{openai2023gpt4}. Potential solutions may include regular updates to the model's knowledge base and fine-tuning with more reliable data sources.
Second, \OurModel{} inherits weaknesses from vision models, such as inaccurate object recognition, people/places identification, or visual relationships reasoning~\cite{li2023blip2}. This calls for a more powerful visual perception model, and training on more well-aligned image-text datasets.
Third, \OurModel{} occasionally encounters difficulties maintaining multi-image and multi-turn consistency. Specifically, the model may lose the context of prior images after several dialogue turns or mix up the contents of different images. Potential solutions involve advanced model designs in position encoding and attention mechanisms to enhance the model's consistency in recalling historical images and dialogues.
Fourth, \OurData{} primarily concentrates on natural images, which limits its versatility in handling text-rich images such as charts, tables, and receipts, as well as domain-specific images such as medical scans, math illustrations, and satellite photos. Moreover, the dialogues in \OurData{} do not cover all possible user scenarios. Therefore, broadening the dataset to cover more diverse image types and user cases is a direction for future work.
Lastly, the reliability of \OurEval{} is tied to the capabilities of current GPT models. This limitation can be mitigated by incorporating more robust judge models and the assistance of human evaluators.
Future works addressing these issues should make for a more reliable and robust system.

\clearpage

\bibliography{iclr2024_conference}

\begin{thebibliography}{62}
\providecommand{\natexlab}[1]{#1}
\providecommand{\url}[1]{\texttt{#1}}
\expandafter\ifx\csname urlstyle\endcsname\relax
  \providecommand{\doi}[1]{doi: #1}\else
  \providecommand{\doi}{doi: \begingroup \urlstyle{rm}\Url}\fi

\bibitem[Aghajanyan et~al.(2022)Aghajanyan, Huang, Ross, Karpukhin, Xu, Goyal, Okhonko, Joshi, Ghosh, Lewis, et~al.]{aghajanyan2022cm3}
Armen Aghajanyan, Bernie Huang, Candace Ross, Vladimir Karpukhin, Hu~Xu, Naman Goyal, Dmytro Okhonko, Mandar Joshi, Gargi Ghosh, Mike Lewis, et~al.
\newblock Cm3: A causal masked multimodal model of the internet.
\newblock \emph{arXiv preprint arXiv:2201.07520}, 2022.

\bibitem[Alayrac et~al.(2022)Alayrac, Donahue, Luc, Miech, Barr, Hasson, Lenc, Mensch, Millican, Reynolds, et~al.]{alayrac2022flamingo}
Jean-Baptiste Alayrac, Jeff Donahue, Pauline Luc, Antoine Miech, Iain Barr, Yana Hasson, Karel Lenc, Arthur Mensch, Katherine Millican, Malcolm Reynolds, et~al.
\newblock Flamingo: a visual language model for few-shot learning.
\newblock In \emph{NeurIPS}, 2022.

\bibitem[Awadalla et~al.(2023)Awadalla, Gao, Gardner, Hessel, Hanafy, Zhu, Marathe, Bitton, Gadre, Sagawa, et~al.]{awadalla2023openflamingo}
Anas Awadalla, Irena Gao, Josh Gardner, Jack Hessel, Yusuf Hanafy, Wanrong Zhu, Kalyani Marathe, Yonatan Bitton, Samir Gadre, Shiori Sagawa, et~al.
\newblock Openflamingo: An open-source framework for training large autoregressive vision-language models.
\newblock \emph{arXiv preprint arXiv:2308.01390}, 2023.

\bibitem[Changpinyo et~al.(2021)Changpinyo, Sharma, Ding, and Soricut]{changpinyo2021conceptual}
Soravit Changpinyo, Piyush Sharma, Nan Ding, and Radu Soricut.
\newblock Conceptual 12m: Pushing web-scale image-text pre-training to recognize long-tail visual concepts.
\newblock In \emph{CVPR}, 2021.

\bibitem[Chen et~al.(2023)Chen, Liu, Dai, and Wang]{chen2023visual}
Delong Chen, Jianfeng Liu, Wenliang Dai, and Baoyuan Wang.
\newblock Visual instruction tuning with polite flamingo.
\newblock \emph{arXiv preprint arXiv:2307.01003}, 2023.

\bibitem[Chiang et~al.(2023)Chiang, Li, Lin, Sheng, Wu, Zhang, Zheng, Zhuang, Zhuang, Gonzalez, Stoica, and Xing]{vicuna2023}
Wei-Lin Chiang, Zhuohan Li, Zi~Lin, Ying Sheng, Zhanghao Wu, Hao Zhang, Lianmin Zheng, Siyuan Zhuang, Yonghao Zhuang, Joseph~E. Gonzalez, Ion Stoica, and Eric~P. Xing.
\newblock Vicuna: An open-source chatbot impressing gpt-4 with 90\%* chatgpt quality, March 2023.
\newblock URL \url{https://vicuna.lmsys.org}.

\bibitem[Contributors(2023)]{2023opencompass}
OpenCompass Contributors.
\newblock Opencompass: A universal evaluation platform for foundation models.
\newblock \url{https://github.com/open-compass/opencompass}, 2023.

\bibitem[Dai et~al.(2023)Dai, Li, Li, Tiong, Zhao, Wang, Li, Fung, and Hoi]{dai2023instructblip}
Wenliang Dai, Junnan Li, Dongxu Li, Anthony Meng~Huat Tiong, Junqi Zhao, Weisheng Wang, Boyang Li, Pascale Fung, and Steven Hoi.
\newblock Instructblip: Towards general-purpose vision-language models with instruction tuning.
\newblock \emph{arXiv preprint arXiv:2305.06500}, 2023.

\bibitem[Das et~al.(2017)Das, Kottur, Gupta, Singh, Yadav, Moura, Parikh, and Batra]{das2017visual}
Abhishek Das, Satwik Kottur, Khushi Gupta, Avi Singh, Deshraj Yadav, Jos{\'e}~MF Moura, Devi Parikh, and Dhruv Batra.
\newblock Visual dialog.
\newblock In \emph{CVPR}, 2017.

\bibitem[Dosovitskiy et~al.(2021)Dosovitskiy, Beyer, Kolesnikov, Weissenborn, Zhai, Unterthiner, Dehghani, Minderer, Heigold, Gelly, Uszkoreit, and Houlsby]{dosovitskiy2021vit}
Alexey Dosovitskiy, Lucas Beyer, Alexander Kolesnikov, Dirk Weissenborn, Xiaohua Zhai, Thomas Unterthiner, Mostafa Dehghani, Matthias Minderer, Georg Heigold, Sylvain Gelly, Jakob Uszkoreit, and Neil Houlsby.
\newblock An image is worth 16x16 words: Transformers for image recognition at scale.
\newblock In \emph{ICLR}, 2021.

\bibitem[Fang et~al.(2022)Fang, Wang, Xie, Sun, Wu, Wang, Huang, Wang, and Cao]{fang2022eva}
Yuxin Fang, Wen Wang, Binhui Xie, Quan Sun, Ledell Wu, Xinggang Wang, Tiejun Huang, Xinlong Wang, and Yue Cao.
\newblock Eva: Exploring the limits of masked visual representation learning at scale.
\newblock \emph{arXiv preprint arXiv:2211.07636}, 2022.

\bibitem[Feng et~al.(2023)Feng, Sun, Xu, Zhao, Yang, Tao, Zhao, and Lin]{feng-etal-2023-mmdialog}
Jiazhan Feng, Qingfeng Sun, Can Xu, Pu~Zhao, Yaming Yang, Chongyang Tao, Dongyan Zhao, and Qingwei Lin.
\newblock {MMD}ialog: A large-scale multi-turn dialogue dataset towards multi-modal open-domain conversation.
\newblock In \emph{ACL}, 2023.

\bibitem[Gao et~al.(2023)Gao, Han, Zhang, Lin, Geng, Zhou, Zhang, Lu, He, Yue, et~al.]{gao2023llama}
Peng Gao, Jiaming Han, Renrui Zhang, Ziyi Lin, Shijie Geng, Aojun Zhou, Wei Zhang, Pan Lu, Conghui He, Xiangyu Yue, et~al.
\newblock Llama-adapter v2: Parameter-efficient visual instruction model.
\newblock \emph{arXiv preprint arXiv:2304.15010}, 2023.

\bibitem[Gong et~al.(2023)Gong, Lyu, Zhang, Wang, Zheng, Zhao, Liu, Zhang, Luo, and Chen]{gong2023multimodal}
Tao Gong, Chengqi Lyu, Shilong Zhang, Yudong Wang, Miao Zheng, Qian Zhao, Kuikun Liu, Wenwei Zhang, Ping Luo, and Kai Chen.
\newblock Multimodal-gpt: A vision and language model for dialogue with humans.
\newblock \emph{arXiv preprint arXiv:2305.04790}, 2023.

\bibitem[Guan et~al.(2023)Guan, Liu, Li, Wang, Yacoob, and Zhou]{guan2023hallusionbench}
Tianrui Guan, Fuxiao Liu, Xiyang Wu Ruiqi Xian~Zongxia Li, Xiaoyu Liu~Xijun Wang, Lichang Chen Furong Huang~Yaser Yacoob, and Dinesh Manocha~Tianyi Zhou.
\newblock Hallusionbench: An advanced diagnostic suite for entangled language hallucination \& visual illusion in large vision-language models.
\newblock \emph{arXiv e-prints}, 2023.

\bibitem[Hu et~al.(2019)Hu, Misra, and Van Der~Maaten]{hu2019evaluating}
Hexiang Hu, Ishan Misra, and Laurens Van Der~Maaten.
\newblock Evaluating text-to-image matching using binary image selection (bison).
\newblock In \emph{Proceedings of the IEEE/CVF International Conference on Computer Vision Workshops}, 2019.

\bibitem[Huang et~al.(2023)Huang, Dong, Wang, Hao, Singhal, Ma, Lv, Cui, Mohammed, Liu, et~al.]{huang2023language}
Shaohan Huang, Li~Dong, Wenhui Wang, Yaru Hao, Saksham Singhal, Shuming Ma, Tengchao Lv, Lei Cui, Owais~Khan Mohammed, Qiang Liu, et~al.
\newblock Language is not all you need: Aligning perception with language models.
\newblock \emph{arXiv preprint arXiv:2302.14045}, 2023.

\bibitem[Huang et~al.(2016)Huang, Ferraro, Mostafazadeh, Misra, Agrawal, Devlin, Girshick, He, Kohli, Batra, et~al.]{huang2016visual}
Ting-Hao Huang, Francis Ferraro, Nasrin Mostafazadeh, Ishan Misra, Aishwarya Agrawal, Jacob Devlin, Ross Girshick, Xiaodong He, Pushmeet Kohli, Dhruv Batra, et~al.
\newblock Visual storytelling.
\newblock In \emph{NAACL}, 2016.

\bibitem[Huang et~al.(2021{\natexlab{a}})Huang, Xue, Liu, and Lu]{huang2021unifying}
Yupan Huang, Hongwei Xue, Bei Liu, and Yutong Lu.
\newblock Unifying multimodal transformer for bi-directional image and text generation.
\newblock In \emph{ACM Multimedia}, 2021{\natexlab{a}}.

\bibitem[Huang et~al.(2021{\natexlab{b}})Huang, Zeng, Huang, Liu, Fu, and Fu]{huang2021seeing}
Zhicheng Huang, Zhaoyang Zeng, Yupan Huang, Bei Liu, Dongmei Fu, and Jianlong Fu.
\newblock Seeing out of the box: End-to-end pre-training for vision-language representation learning.
\newblock In \emph{CVPR}, 2021{\natexlab{b}}.

\bibitem[Jia et~al.(2021)Jia, Yang, Xia, Chen, Parekh, Pham, Le, Sung, Li, and Duerig]{jia2021scaling}
Chao Jia, Yinfei Yang, Ye~Xia, Yi-Ting Chen, Zarana Parekh, Hieu Pham, Quoc Le, Yun-Hsuan Sung, Zhen Li, and Tom Duerig.
\newblock Scaling up visual and vision-language representation learning with noisy text supervision.
\newblock In \emph{ICML}, 2021.

\bibitem[Kitaev \& Klein(2018)Kitaev and Klein]{kitaev-klein-2018-constituency}
Nikita Kitaev and Dan Klein.
\newblock Constituency parsing with a self-attentive encoder.
\newblock In \emph{ACL}, 2018.

\bibitem[Kojima et~al.(2022)Kojima, Gu, Reid, Matsuo, and Iwasawa]{kojima2022large}
Takeshi Kojima, Shixiang~Shane Gu, Machel Reid, Yutaka Matsuo, and Yusuke Iwasawa.
\newblock Large language models are zero-shot reasoners.
\newblock \emph{NeurIPS}, 2022.

\bibitem[Krishna et~al.(2017)Krishna, Zhu, Groth, Johnson, Hata, Kravitz, Chen, Kalantidis, Li, Shamma, et~al.]{krishna2017visual}
Ranjay Krishna, Yuke Zhu, Oliver Groth, Justin Johnson, Kenji Hata, Joshua Kravitz, Stephanie Chen, Yannis Kalantidis, Li-Jia Li, David~A Shamma, et~al.
\newblock Visual genome: Connecting language and vision using crowdsourced dense image annotations.
\newblock \emph{International Journal of Computer Vision}, 123\penalty0 (1):\penalty0 32--73, 2017.

\bibitem[Lee et~al.(2021)Lee, Shin, Choo, Choi, and Myaeng]{lee2021constructing}
Nyoungwoo Lee, Suwon Shin, Jaegul Choo, Ho-Jin Choi, and Sung-Hyon Myaeng.
\newblock Constructing multi-modal dialogue dataset by replacing text with semantically relevant images.
\newblock In \emph{Proceedings of the 59th Annual Meeting of the Association for Computational Linguistics and the 11th International Joint Conference on Natural Language Processing (Volume 2: Short Papers)}, 2021.

\bibitem[Lee et~al.(2022)Lee, Ko, Kim, and Choi]{lee2022dialogcc}
Young-Jun Lee, Byungsoo Ko, Han-Gyu Kim, and Ho-Jin Choi.
\newblock Dialogcc: Large-scale multi-modal dialogue dataset.
\newblock \emph{arXiv preprint arXiv:2212.04119}, 2022.

\bibitem[Li et~al.(2023{\natexlab{a}})Li, Zhang, Chen, Wang, Pu, Yang, Li, and Liu]{li2023mimic}
Bo~Li, Yuanhan Zhang, Liangyu Chen, Jinghao Wang, Fanyi Pu, Jingkang Yang, Chunyuan Li, and Ziwei Liu.
\newblock Mimic-it: Multi-modal in-context instruction tuning.
\newblock \emph{arXiv preprint arXiv:2306.05425}, 2023{\natexlab{a}}.

\bibitem[Li et~al.(2023{\natexlab{b}})Li, Zhang, Chen, Wang, Yang, and Liu]{li2023otter}
Bo~Li, Yuanhan Zhang, Liangyu Chen, Jinghao Wang, Jingkang Yang, and Ziwei Liu.
\newblock Otter: A multi-modal model with in-context instruction tuning.
\newblock \emph{arXiv preprint arXiv:2305.03726}, 2023{\natexlab{b}}.

\bibitem[Li et~al.(2023{\natexlab{c}})Li, Li, Savarese, and Hoi]{li2023blip2}
Junnan Li, Dongxu Li, Silvio Savarese, and Steven Hoi.
\newblock Blip-2: Bootstrapping language-image pre-training with frozen image encoders and large language models.
\newblock \emph{arXiv preprint arXiv:2301.12597}, 2023{\natexlab{c}}.

\bibitem[Lin et~al.(2014)Lin, Maire, Belongie, Hays, Perona, Ramanan, Doll{\'a}r, and Zitnick]{lin2014microsoft}
Tsung-Yi Lin, Michael Maire, Serge Belongie, James Hays, Pietro Perona, Deva Ramanan, Piotr Doll{\'a}r, and C~Lawrence Zitnick.
\newblock Microsoft coco: Common objects in context.
\newblock In \emph{ECCV}, 2014.

\bibitem[Liu et~al.(2023{\natexlab{a}})Liu, Lin, Li, Wang, Yacoob, and Wang]{liu2023aligning}
Fuxiao Liu, Kevin Lin, Linjie Li, Jianfeng Wang, Yaser Yacoob, and Lijuan Wang.
\newblock Aligning large multi-modal model with robust instruction tuning.
\newblock \emph{arXiv preprint arXiv:2306.14565}, 2023{\natexlab{a}}.

\bibitem[Liu et~al.(2023{\natexlab{b}})Liu, Li, Li, and Lee]{liu2023improved}
Haotian Liu, Chunyuan Li, Yuheng Li, and Yong~Jae Lee.
\newblock Improved baselines with visual instruction tuning.
\newblock \emph{arXiv preprint arXiv:2310.03744}, 2023{\natexlab{b}}.

\bibitem[Liu et~al.(2023{\natexlab{c}})Liu, Li, Wu, and Lee]{liu2023visual}
Haotian Liu, Chunyuan Li, Qingyang Wu, and Yong~Jae Lee.
\newblock Visual instruction tuning.
\newblock In \emph{NeurIPS}, 2023{\natexlab{c}}.

\bibitem[Lu et~al.(2022)Lu, Mishra, Xia, Qiu, Chang, Zhu, Tafjord, Clark, and Kalyan]{lu2022learn}
Pan Lu, Swaroop Mishra, Tony Xia, Liang Qiu, Kai-Wei Chang, Song-Chun Zhu, Oyvind Tafjord, Peter Clark, and Ashwin Kalyan.
\newblock Learn to explain: Multimodal reasoning via thought chains for science question answering.
\newblock In \emph{NeurIPS}, 2022.

\bibitem[Meng et~al.(2020)Meng, Wang, Han, Sun, Wu, Yan, and Li]{meng2020openvidial}
Yuxian Meng, Shuhe Wang, Qinghong Han, Xiaofei Sun, Fei Wu, Rui Yan, and Jiwei Li.
\newblock Openvidial: A large-scale, open-domain dialogue dataset with visual contexts.
\newblock \emph{arXiv preprint arXiv:2012.15015}, 2020.

\bibitem[Mostafazadeh et~al.(2017)Mostafazadeh, Brockett, Dolan, Galley, Gao, Spithourakis, and Vanderwende]{mostafazadeh-etal-2017-image}
Nasrin Mostafazadeh, Chris Brockett, Bill Dolan, Michel Galley, Jianfeng Gao, Georgios Spithourakis, and Lucy Vanderwende.
\newblock Image-grounded conversations: Multimodal context for natural question and response generation.
\newblock In \emph{Proceedings of the Eighth International Joint Conference on Natural Language Processing (Volume 1: Long Papers)}, 2017.

\bibitem[Openai(2023)]{openai2023gpt4}
Openai.
\newblock {GPT-4} technical report.
\newblock \emph{arXiv preprint arXiv:2303.08774}, 2023.

\bibitem[Ouyang et~al.(2022)Ouyang, Wu, Jiang, Almeida, Wainwright, Mishkin, Zhang, Agarwal, Slama, Ray, et~al.]{ouyang2022training}
Long Ouyang, Jeffrey Wu, Xu~Jiang, Diogo Almeida, Carroll Wainwright, Pamela Mishkin, Chong Zhang, Sandhini Agarwal, Katarina Slama, Alex Ray, et~al.
\newblock Training language models to follow instructions with human feedback.
\newblock In \emph{NeurIPS}, 2022.

\bibitem[Radford et~al.(2021)Radford, Kim, Hallacy, Ramesh, Goh, Agarwal, Sastry, Askell, Mishkin, Clark, et~al.]{radford2021learning}
Alec Radford, Jong~Wook Kim, Chris Hallacy, Aditya Ramesh, Gabriel Goh, Sandhini Agarwal, Girish Sastry, Amanda Askell, Pamela Mishkin, Jack Clark, et~al.
\newblock Learning transferable visual models from natural language supervision.
\newblock In \emph{ICML}, 2021.

\bibitem[Rombach et~al.(2022)Rombach, Blattmann, Lorenz, Esser, and Ommer]{Rombach_2022_CVPR}
Robin Rombach, Andreas Blattmann, Dominik Lorenz, Patrick Esser, and Bj\"orn Ommer.
\newblock High-resolution image synthesis with latent diffusion models.
\newblock In \emph{CVPR}, 2022.

\bibitem[Schuhmann et~al.(2021)Schuhmann, Vencu, Beaumont, Kaczmarczyk, Mullis, Katta, Coombes, Jitsev, and Komatsuzaki]{schuhmann2021laion}
Christoph Schuhmann, Richard Vencu, Romain Beaumont, Robert Kaczmarczyk, Clayton Mullis, Aarush Katta, Theo Coombes, Jenia Jitsev, and Aran Komatsuzaki.
\newblock Laion-400m: Open dataset of clip-filtered 400 million image-text pairs.
\newblock \emph{arXiv preprint arXiv:2111.02114}, 2021.

\bibitem[Sharma et~al.(2018)Sharma, Ding, Goodman, and Soricut]{sharma2018conceptual}
Piyush Sharma, Nan Ding, Sebastian Goodman, and Radu Soricut.
\newblock Conceptual captions: A cleaned, hypernymed, image alt-text dataset for automatic image captioning.
\newblock In \emph{ACL}, 2018.

\bibitem[Shuster et~al.(2020)Shuster, Humeau, Bordes, and Weston]{shuster2020image}
Kurt Shuster, Samuel Humeau, Antoine Bordes, and Jason Weston.
\newblock Image-chat: Engaging grounded conversations.
\newblock In \emph{ACL}, 2020.

\bibitem[Su et~al.(2023)Su, Lan, Li, Xu, Wang, and Cai]{su2023pandagpt}
Yixuan Su, Tian Lan, Huayang Li, Jialu Xu, Yan Wang, and Deng Cai.
\newblock Pandagpt: One model to instruction-follow them all.
\newblock \emph{arXiv preprint arXiv:2305.16355}, 2023.

\bibitem[Suhr et~al.(2019)Suhr, Zhou, Zhang, Zhang, Bai, and Artzi]{suhr2019corpus}
Alane Suhr, Stephanie Zhou, Ally Zhang, Iris Zhang, Huajun Bai, and Yoav Artzi.
\newblock A corpus for reasoning about natural language grounded in photographs.
\newblock In \emph{ACL}, 2019.

\bibitem[Sun et~al.(2023)Sun, Yu, Cui, Zhang, Zhang, Wang, Gao, Liu, Huang, and Wang]{sun2023generative}
Quan Sun, Qiying Yu, Yufeng Cui, Fan Zhang, Xiaosong Zhang, Yueze Wang, Hongcheng Gao, Jingjing Liu, Tiejun Huang, and Xinlong Wang.
\newblock Generative pretraining in multimodality.
\newblock \emph{arXiv preprint arXiv:2307.05222}, 2023.

\bibitem[Touvron et~al.(2023)Touvron, Lavril, Izacard, Martinet, Lachaux, Lacroix, Rozi{\`e}re, Goyal, Hambro, Azhar, et~al.]{llama}
Hugo Touvron, Thibaut Lavril, Gautier Izacard, Xavier Martinet, Marie-Anne Lachaux, Timoth{\'e}e Lacroix, Baptiste Rozi{\`e}re, Naman Goyal, Eric Hambro, Faisal Azhar, et~al.
\newblock Llama: Open and efficient foundation language models.
\newblock \emph{arXiv preprint arXiv:2302.13971}, 2023.

\bibitem[Viktor \& Denis(2023)Viktor and Denis]{viktor2023imad}
Moskvoretskii Viktor and Kuznetsov Denis.
\newblock Imad: Image-augmented multi-modal dialogue.
\newblock \emph{arXiv preprint arXiv:2305.10512}, 2023.

\bibitem[Wang et~al.(2021)Wang, Meng, Li, Sun, Ouyang, and Li]{wang2021openvidial}
Shuhe Wang, Yuxian Meng, Xiaoya Li, Xiaofei Sun, Rongbin Ouyang, and Jiwei Li.
\newblock Openvidial 2.0: A larger-scale, open-domain dialogue generation dataset with visual contexts.
\newblock \emph{arXiv preprint arXiv:2109.12761}, 2021.

\bibitem[Wang et~al.(2022)Wang, Kordi, Mishra, Liu, Smith, Khashabi, and Hajishirzi]{wang2022self}
Yizhong Wang, Yeganeh Kordi, Swaroop Mishra, Alisa Liu, Noah~A Smith, Daniel Khashabi, and Hannaneh Hajishirzi.
\newblock Self-instruct: Aligning language model with self generated instructions.
\newblock \emph{arXiv preprint arXiv:2212.10560}, 2022.

\bibitem[Wei et~al.(2022)Wei, Bosma, Zhao, Guu, Yu, Lester, Du, Dai, and Le]{wei2022finetuned}
Jason Wei, Maarten Bosma, Vincent Zhao, Kelvin Guu, Adams~Wei Yu, Brian Lester, Nan Du, Andrew~M. Dai, and Quoc~V Le.
\newblock Finetuned language models are zero-shot learners.
\newblock In \emph{ICLR}, 2022.

\bibitem[Xu et~al.(2022)Xu, Shen, and Huang]{xu2022multiinstruct}
Zhiyang Xu, Ying Shen, and Lifu Huang.
\newblock Multiinstruct: Improving multi-modal zero-shot learning via instruction tuning.
\newblock \emph{arXiv preprint arXiv:2212.10773}, 2022.

\bibitem[Ye et~al.(2023)Ye, Xu, Xu, Ye, Yan, Zhou, Wang, Hu, Shi, Shi, et~al.]{ye2023mplug}
Qinghao Ye, Haiyang Xu, Guohai Xu, Jiabo Ye, Ming Yan, Yiyang Zhou, Junyang Wang, Anwen Hu, Pengcheng Shi, Yaya Shi, et~al.
\newblock mplug-owl: Modularization empowers large language models with multimodality.
\newblock \emph{arXiv preprint arXiv:2304.14178}, 2023.

\bibitem[Yin et~al.(2023{\natexlab{a}})Yin, Fu, Zhao, Li, Sun, Xu, and Chen]{yin2023survey}
Shukang Yin, Chaoyou Fu, Sirui Zhao, Ke~Li, Xing Sun, Tong Xu, and Enhong Chen.
\newblock A survey on multimodal large language models.
\newblock \emph{arXiv preprint arXiv:2306.13549}, 2023{\natexlab{a}}.

\bibitem[Yin et~al.(2023{\natexlab{b}})Yin, Wang, Cao, Shi, Liu, Li, Sheng, Bai, Huang, Wang, et~al.]{yin2023lamm}
Zhenfei Yin, Jiong Wang, Jianjian Cao, Zhelun Shi, Dingning Liu, Mukai Li, Lu~Sheng, Lei Bai, Xiaoshui Huang, Zhiyong Wang, et~al.
\newblock Lamm: Language-assisted multi-modal instruction-tuning dataset, framework, and benchmark.
\newblock \emph{arXiv preprint arXiv:2306.06687}, 2023{\natexlab{b}}.

\bibitem[Yue et~al.(2023)Yue, Ni, Zhang, Zheng, Liu, Zhang, Stevens, Jiang, Ren, Sun, Wei, Yu, Yuan, Sun, Yin, Zheng, Yang, Liu, Huang, Sun, Su, and Chen]{yue2023mmmu}
Xiang Yue, Yuansheng Ni, Kai Zhang, Tianyu Zheng, Ruoqi Liu, Ge~Zhang, Samuel Stevens, Dongfu Jiang, Weiming Ren, Yuxuan Sun, Cong Wei, Botao Yu, Ruibin Yuan, Renliang Sun, Ming Yin, Boyuan Zheng, Zhenzhu Yang, Yibo Liu, Wenhao Huang, Huan Sun, Yu~Su, and Wenhu Chen.
\newblock Mmmu: A massive multi-discipline multimodal understanding and reasoning benchmark for expert agi.
\newblock \emph{arXiv preprint arXiv:2311.16502}, 2023.

\bibitem[Zang et~al.(2021)Zang, Liu, Wang, Song, Zhang, and Chen]{zang2021photochat}
Xiaoxue Zang, Lijuan Liu, Maria Wang, Yang Song, Hao Zhang, and Jindong Chen.
\newblock Photochat: A human-human dialogue dataset with photo sharing behavior for joint image-text modeling.
\newblock In \emph{Proceedings of the 59th Annual Meeting of the Association for Computational Linguistics and the 11th International Joint Conference on Natural Language Processing (Volume 1: Long Papers)}, 2021.

\bibitem[Zhao et~al.(2023)Zhao, Wu, and Huang]{zhao2023svit}
Bo~Zhao, Boya Wu, and Tiejun Huang.
\newblock Svit: Scaling up visual instruction tuning.
\newblock \emph{arXiv preprint arXiv:2307.04087}, 2023.

\bibitem[Zheng et~al.(2023)Zheng, Chiang, Sheng, Zhuang, Wu, Zhuang, Lin, Li, Li, Xing, et~al.]{zheng2023judging}
Lianmin Zheng, Wei-Lin Chiang, Ying Sheng, Siyuan Zhuang, Zhanghao Wu, Yonghao Zhuang, Zi~Lin, Zhuohan Li, Dacheng Li, Eric Xing, et~al.
\newblock Judging llm-as-a-judge with mt-bench and chatbot arena.
\newblock \emph{arXiv preprint arXiv:2306.05685}, 2023.

\bibitem[Zheng et~al.(2022)Zheng, Chen, Liu, and Sun]{zheng2022MMChat}
Yinhe Zheng, Guanyi Chen, Xin Liu, and Jian Sun.
\newblock {MMC}hat: Multi-modal chat dataset on social media.
\newblock In \emph{Proceedings of The 13th Language Resources and Evaluation Conference}, 2022.

\bibitem[Zhu et~al.(2023{\natexlab{a}})Zhu, Chen, Shen, Li, and Elhoseiny]{zhu2023minigpt4}
Deyao Zhu, Jun Chen, Xiaoqian Shen, Xiang Li, and Mohamed Elhoseiny.
\newblock Minigpt-4: Enhancing vision-language understanding with advanced large language models.
\newblock \emph{arXiv preprint arXiv:2304.10592}, 2023{\natexlab{a}}.

\bibitem[Zhu et~al.(2023{\natexlab{b}})Zhu, Hessel, Awadalla, Gadre, Dodge, Fang, Yu, Schmidt, Wang, and Choi]{zhu2023multimodal}
Wanrong Zhu, Jack Hessel, Anas Awadalla, Samir~Yitzhak Gadre, Jesse Dodge, Alex Fang, Youngjae Yu, Ludwig Schmidt, William~Yang Wang, and Yejin Choi.
\newblock {Multimodal C4}: An open, billion-scale corpus of images interleaved with text.
\newblock \emph{arXiv preprint arXiv:2304.06939}, 2023{\natexlab{b}}.

\end{thebibliography}
\bibliographystyle{iclr2024_conference}

\clearpage
\appendix

\begin{center}
{\LARGE \textbf{Appendix}}
\end{center}

\startcontents[appendices]
{\hypersetup{hidelinks}
\printcontents[appendices]{l}{1}{\setcounter{tocdepth}{2}}
}

\section{Instruction-tuning details}
\label{appendix:architecutre_training_details}

We represent an $i$-th $T$-turn dialogue as $\Xmat^i=\left(\Xmat_{\texttt{q}}^{i, 1}, \Xmat_{\texttt{a}}^{i, 1}, \cdots, \Xmat_{\texttt{q}}^{i, T}, \Xmat_{\texttt{a}}^{i, T} \right)$, where each pair of $\left(\Xmat_{\texttt{q}}^{i, t}, \Xmat_{\texttt{a}}^{i, t}\right)$ includes a question from the user and an answer from the assistant in turn-$t$. 
For each $\Xmat^i$, we construct $T$ training samples by organizing each pair of questions and answers as a sequence. Given a predefined system prompt $\Xmat_{\texttt{system}}$, the prompt $\Xmat_{\texttt{prompt}}^{i, t} $ and response $\Xmat_{\texttt{response}}^{i, t}$ at the $t$-th turn are defined as the following:
\begin{equation}
\begin{aligned}
\Xmat_{\texttt{prompt}}^{i, t} =
\begin{cases} 
\Xmat_{\texttt{system}}\texttt{<SEP>} \texttt{Human}: \Xmat_{\texttt{q}}^{i, 1}\texttt{<SEP>} \\ \texttt{Assistant}:, \text{if } t = 1,\\
\Xmat_{\texttt{prompt}}^{i, t-1}\texttt{<SEP>} \texttt{Human}: \Xmat_{\texttt{q}}^{i, t}\texttt{<SEP>} \\ \texttt{Assistant}:, \text{if } t > 1.
\end{cases}
\end{aligned}
\end{equation}
\begin{equation}
\Xmat_{\texttt{response}}^{i, t} = \Xmat_{\texttt{a}}^{i, t}\texttt{<SEP>}.
\end{equation}

We train the LLM on the prediction tokens using the auto-regressive training objective. Specifically, for a sequence of length $L$, we compute the probability of
generating target responses $\Xmat_{\texttt{response}}$ by:
\begin{equation}
\begin{aligned}
    p\left( \Xmat_{\texttt{response}} |  \Xmat_{\texttt{prompt}}\right) =\\
    \prod_{l=1}^{L} p_{\thetav} \left(  {\xv_l}
| \Xmat_{\texttt{prompt}, <l}, \Xmat_{\texttt{response}, <l}\right),
\end{aligned}
    \label{eq:auto_regressive}
\end{equation}
where $\thetav$ is the trainable parameters, $\Xmat_{\texttt{prompt}, <l}$ and $\Xmat_{\texttt{response}, <l}$ are prompt and response tokens in all turns before the current prediction token ${\xv_l}$.

\section{Related works}
\label{appendix:related_works}

Our work exploits image-text pairs to construct a dialogue dataset for instruction-tuning. Thus, we review related works on multimodal alignment datasets, multimodal dialogue datasets, and multimodal instruction tuning, primarily on natural images and text domains.

\paragraph{Multimodal alignment datasets.}
Various datasets, such as MSCOCO~\cite{lin2014microsoft}, Visual Genome~\cite{krishna2017visual}, Conceptual Captions~\cite{sharma2018conceptual}, Conceptual 12M \cite{changpinyo2021conceptual}, ALIGN~\cite{jia2021scaling} and LAION~\cite{schuhmann2021laion}, have been constructed to \textbf{align images with their corresponding descriptions}. These datasets have significantly contributed to the development of multimodal models for image-and-text generation~\cite{huang2021unifying,Rombach_2022_CVPR,li2023blip2} and understanding~\cite{jia2021scaling,huang2021seeing,radford2021learning}. We use these datasets in our data construction process.
Emerging trends include datasets featuring \textbf{interleaved images and text sequences from web corpora}, such as M3W~\cite{alayrac2022flamingo}, web and Wikipedia articles~\cite{aghajanyan2022cm3}, Common Crawl Interleaved data~\cite{huang2023language}, and the Multimodal C4 dataset~\cite{zhu2023multimodal}. These datasets extend conventional image-text alignment training by incorporating multiple images and sentences. When trained on these enriched datasets, models such as Flamingo~\cite{alayrac2022flamingo}, OpenFlamingo~\cite{awadalla2023openflamingo}, Kosmos-1~\cite{huang2023language}, and EMU~\cite{sun2023generative} can adapt to various image understanding tasks using multiple task-relevant image-text examples. However, these models often fall short in following intricate human instructions because they are trained to predict the next word on a large web dataset rather than perform the task the user wants~\cite{ouyang2022training}.

\paragraph{Multimodal dialogue datasets.}
Existing multimodal dialogue datasets broadly fall into two categories. The first comprises datasets where \textbf{conversations are heavily rooted in and driven by images}. Traditional datasets of this type are primarily generated by inviting crowd workers to engage in dialogues about a common image. Notable examples include Visual Dialog~\cite{das2017visual}, which emphasizes question-answering tasks within AI-human chat about visual content, and IGC~\cite{mostafazadeh-etal-2017-image}, a compilation of dialogues featuring an image, a corresponding textual description, and a conversation centered on the image. Image-Chat presents image-grounded dialogues crafted around given images~\cite{shuster2020image}. Recently, dialogue datasets, such as LLaVA~\cite{liu2023visual}, SVIT~\cite{zhao2023svit}, and LAMM~\cite{yin2023lamm}, created by LLMs alongside image annotations have surfaced. Each dialogue in these datasets begins with an inquiry about image attributes or factual knowledge, with responses expected to be brief within 50 words, which may not align with real-world scenarios requiring in-depth multi-image analysis.
The second category features \textbf{daily human conversations}, with images interspersed within multi-turn conversations sparsely. For example, OpenViDial~\cite{meng2020openvidial,wang2021openvidial} is sourced from dialogues in movies and TV series, whereas PhotoChat~\cite{zang2021photochat} is a human-human dialogue dataset developed through crowdsourcing and features photo-sharing. Other datasets, such as DialogCC~\cite{lee2022dialogcc}, MultiModalDialogue~\cite{lee2021constructing}, and IMAD~\cite{viktor2023imad} enhance text-only dialogues by incorporating semantically relevant images. In addition, MMChat~\cite{zheng2022MMChat} and MMDialog~\cite{feng-etal-2023-mmdialog} encompass image-grounded dialogues derived from social media interactions. However, these datasets, not being designed for user-assistant interactions, struggle with instructive, problem-solving dialogue requirements.

\paragraph{Multimodal instruction tuning.}
Multimodal instruction tuning has grown substantially with the advent of multimodal instruction datasets. For instance, MultiInstruct~\cite{xu2022multiinstruct} offers a benchmark comprising 62 diverse multimodal tasks unified in a seq-to-seq format. InstructBLIP~\cite{dai2023instructblip} extended the scope by transforming 26 datasets into instruction-tuning form. Otter~\cite{li2023otter} is trained on MIMIC-IT~\cite{li2023mimic}, a multimodal in-context instruction tuning dataset constructed by grouping multiple similar instructions into a contextual example.
To better align with user intentions, MiniGPT-4 is fine-tuned on a small dataset of detailed image descriptions ~\cite{zhu2023minigpt4} and PF-1M~\cite{chen2023visual} rewrites image annotations in a human-like style across 37 vision-language datasets.
Furthermore, techniques such as LLaVA~\cite{liu2023visual}, SVIT~\cite{zhao2023svit}, LRV-Instruction~\cite{liu2023aligning}, and LAMM~\cite{yin2023lamm} have emerged. These methods leverage language-only APIs such as OpenAI's GPT-4~\cite{openai2023gpt4} and self-instruction methods~\cite{wang2022self} to interpret image annotations (e.g., image captions, region descriptions, object bounding boxes, attributes, and relationships), and generate responses in various forms (i.e., short conversations, image captioning, and visual reasoning).
Models such as mPLUG-Owl~\cite{ye2023mplug}, PandaGPT~\cite{su2023pandagpt}, LLaMAAdapter V2~\cite{gao2023llama}, and Multimodal-GPT~\cite{gong2023multimodal} further extended this area, incorporating both language-only and vision-language instruction data.
These developments are a valuable foundation for our work. Our dataset, \OurData{}, is inspired by GPT-assisted data construction techniques and explores the interactions between multiple images and word-level textual content. Training our model, \OurModel{}, on this dataset unlocks the capability of multimodal models to interpret complex image-text interactions.

\section{Implementation details}\label{appendix:implementation_details}
\OurModel{} can be based on different architectures such as MiniGPT-4 or LLaVA.
To train \OurModel{} based on the MiniGPT-4 architecture, we built upon the official MiniGPT-4 codebase~\cite{zhu2023minigpt4}\footnote{\url{https://github.com/Vision-CAIR/MiniGPT-4}}. 
The language decoder, Vicuna~\cite{vicuna2023}, is based on the LLaMA framework~\cite{llama}, which can handle diverse language tasks. For image processing, we use the visual encoder from BLIP-2, combining a pretrained EVA-ViT in Vision Transformer (ViT) backbone with a pretrained Q-Former~\cite{li2023blip2,fang2022eva,dosovitskiy2021vit}.
Only the projection layer is trainable in the model while other vision and language components are frozen.
We refer to MiniGPT-4's efficient fine-tuning process and tune \OurModel{} using 1,500 training steps with a batch size of 8, based on MiniGPT-4's first-stage pretrained model. Our training data of \OurData{} is sampled with the same ratio from \OurDataCC{} and \OurDataVG{}, and with sampling ratios of 2 and 1 from the first and second turns of dialogues, respectively.

During instruction-tuning, we follow MiniGPT-4 to represent images with \texttt{<Img><ImageHere></Img>}. In practice, all tags of \texttt{<ImageHere>} are replaced by the visual features produced by a linear projection layer. Tags of \texttt{<Img>} and \texttt{</Img>} are language tokens that serve as signals for the start and end of images. A system message $\Xmat_{\texttt{system}}$ is appended to the beginning of each prompt. We also append \texttt{Human:} and \texttt{Assistant:} before each user and assistant messages to equip the model with conversation capability. System, user, and assistant messages are separated by a separator \texttt{<SEP>}. The system message $\Xmat_{\texttt{system}}$ = \texttt{\small Give the following image: <Img>ImageContent</Img>. You will be able to see the image once I provide it to you. Please answer my questions.} The separator \texttt{<SEP>} = \texttt{\#\#\#}.
\autoref{tab:input_sequence} illustrates the unified format for two-turn dialogue training sequences.

To train \OurModel{} based on the LLaVA-v1.5 architecture, we built upon the official LLaVA codebase\footnote{\url{https://github.com/haotian-liu/LLaVA/}}~\cite{liu2023visual,liu2023improved} and adapted it to accept multiple images. We adopted a learning rate of 2e-5 and a batch size of 4 per GPU. The training was conducted over 2,000 steps across four GPUs, employing the low-resource technique LoRA to save memory and time. The language model components are based on the 7-billion parameter size of the Vicuna architecture~\cite{vicuna2023}.

In our evaluation, we configure the parameters as follows: \texttt{temperature} is set to 1.0, \texttt{top\_p} to 0.9, and \texttt{max\_new\_tokens} to 300, with both \texttt{repetition\_penalty} and \texttt{length\_penalty} at 1.0. For demonstration cases, the \texttt{beam\_size} is 2; for all other evaluations, it is 1.

\begin{table*}[ht]
\caption{{Prompt and response sequence formats used to train \OurModel{}.} The first and the second conversation turns are illustrated here. The model is trained to predict the assistant answers, and thus only {\color{forestgreen} green sequence} are used to compute the loss in the auto-regressive model. We do not compute the regression loss for the prompt $\Xmat_{\texttt{prompt}}$ since the prompt is provided by users in real-world applications, making it unnecessary for the model to make predictions in this context.}
\label{tab:input_sequence}
\centering
\small
\begin{minipage}{\linewidth}\vspace{0mm}
\centering
\begin{tcolorbox}[colback=blue!0.7!white,colframe=blue!30!black]
\centering
\hspace{+10mm}
\begin{tabular}{p{0.95\columnwidth} c}

\VarSty{ {\bf Dialogue Turn One} } &\\
\textcolor{brickred}{$ \Xmat_{\texttt{prompt}}^{i,1} $} = $\Xmat_{\texttt{system}}${\texttt{<SEP>}}$\texttt{Human}$: $\Xmat_{\texttt{q}}^{i,1}${\texttt{<SEP>}}$\texttt{Assistant}$: \\

\textcolor{brickred}{$ \Xmat_{\texttt{response}}^{i,1} $} = \PredSty{$\Xmat_{\texttt{a} }^{i,1}$}\PredSty{\texttt{<SEP>}}

\hrulefill & \\

\VarSty{ {\bf Dialogue Turn Two} } &\\
\textcolor{brickred}{$ \Xmat_{\texttt{prompt}}^{i,2} $} =
$\Xmat_{\texttt{system}}$  {\texttt{<SEP>}}
$\texttt{Human}$: $\Xmat_{\texttt{q}}^{i,1}${\texttt{<SEP>}}
$\texttt{Assistant}$: 
{$\Xmat_{\texttt{a} }^{i,1}$}{\texttt{<SEP>}}
$\texttt{Human}$: $\Xmat_{\texttt{q}}^{i,2}${\texttt{<SEP>}}
$\texttt{Assistant}$: \\

\textcolor{brickred}{$ \Xmat_{\texttt{response}}^{i,2} $} = \PredSty{$\Xmat_{\texttt{a} }^{i,2}$}\PredSty{\texttt{<SEP>}}
\end{tabular}
\end{tcolorbox}
\end{minipage}
\end{table*}

We tailored the OpenAI's GPT-4 API (\texttt{gpt-4-0613}) parameters to balance diversity and quality for data construction. We set the \texttt{temperature} and \texttt{top\_p} parameters to 1.0, the \texttt{max\_tokens} parameter to 2,048, and both the \texttt{frequency\_penalty} and \texttt{presence\_penalty} to 0.0. In each query to the GPT-4 API, the ``system'' role was allocated the default instruction \texttt{You are a helpful assistant.} As of July 2023, the cost for generating 1,000 tokens was \$0.06 for outputs and \$0.03 for inputs within an 8K context\footnote{\url{https://openai.com/pricing}}, leading to a total dataset generation cost of approximately \$500.
The cost of evaluating a model on \OurEval{} is approximately \$1.4 and \$14 using \texttt{gpt-3.5-turbo-0613} and \texttt{gpt-4-0613}, respectively.

\section{Detailed model comparison on multi-image evaluation}\label{appendix:detail_model_compare_multiple_image}
We further investigate training models on different data sources, including detailed descriptions, complex reasoning, and dialogue data. Results are shown in \autoref{tab:evaluation_compare}.
When \OurModel{} is trained on reasoning data adapted from LLaVA~\cite{liu2023visual}, it achieves improved performance over models trained on description data on all metrics. This emphasizes the importance of reasoning ability.
Its highest scores in both the first and second turns across all criteria indicate its superior ability in image understanding and reasoning, maintaining cross-image and cross-turn coherence, and generating relevant and complete responses. In comparison, models trained on description and reasoning data approximate scores of 3 and 6.71, respectively. GPT-4 scores the highest at 9.26, largely attributed to its utilization of detailed ground-truth annotations. \OurModel{}'s score is about 92\% of that of GPT-4, highlighting \OurModel{}'s conversational competence across images and dialogue turns.

\begin{table*}[t]
  \small
  \caption{Model comparison on BISON, NLVR2 and \OurEval{}. We investigate training models on different data sources, including detailed descriptions, complex reasoning, and dialogue data.
  The evaluation metric is accuracy for BISON and NLVR2; \OurEval{} is rated 1-10. 
Description and reasoning datasets from LLaVA are adapted using formats similar to \OurData{}, with overlapping samples removed between train and evaluation sets.
  }
  \label{tab:evaluation_compare}
\tabcolsep=0.07cm
  \centering
\begin{tabular}{ccccccccccccc}
\toprule
\multirow{3}{*}{\textbf{Architecture}} & \multirow{3}{*}{\textbf{Tuning Data}} &  \multirow{3}{*}{\textbf{BISON}} & \multirow{3}{*}{\textbf{NLVR2}} & \multicolumn{9}{c}{\textbf{\OurEval{}}}  \\ \cmidrule(lr){5-13} 
                &  &  &  & \multirow{2}{*}{\textbf{Score}} & \multicolumn{4}{c}{\textbf{Turn one}} & \multicolumn{4}{c}{\textbf{Turn two}} \\ \cmidrule(lr){6-9} \cmidrule(lr){10-13}
                &  &  &  &                        & \textbf{A1}   & C1   & C2   & C3   & {\textbf{A2}}   & C1   & C2   & C3   \\ \midrule
GPT-4 & - & - & - & \demph{9.26} & \demph{9.26} & \demph{9.23} & \demph{9.18} & \demph{9.38} & \demph{9.26} & \demph{9.25} & \demph{9.15} & \demph{9.38} \\ \hline

\multirow{4}{*}{MiniGPT-4} 
  & MiniGPT-4 description  & 46.0\%  & 51.3\% & 3.91 & 3.55 & 3.67 & 3.53 & 3.44 & 4.28 & 4.38 & 4.21 & 4.23 \\
 & LLaVA description  & 52.0\% & 48.0\% & 3.06 & 2.64 & 2.79 & 2.67 & 2.46 & 3.48 & 3.76  & 3.40 & 3.29 \\
 & LLaVA reasoning & 52.7\% & 54.0\% & 6.71 & 6.55 & 6.63 & 6.42 & 6.59 & 6.87 & 6.89  & 6.73 & 6.98 \\
 &  \textbf{\OurData{}} & {56.7\%} & \textbf{58.0\%} & \textbf{8.56} & \textbf{8.76} & \textbf{8.81} & \textbf{8.67} & \textbf{8.81} & \textbf{8.35} & \textbf{8.37} & {8.28} & \textbf{8.41} \\ 

\hline

\multirow{2}{*}{LLaVA-v1.5} 
 & {Mixture (665K)}  & {52.7\%}  & {53.3\%} & {2.75} & {2.80} & {2.74} & {2.94} & {2.71} & {2.69} & {2.69} & {2.70} & {2.68} \\
 &  {\textbf{\OurData{}}} & {\textbf{65.3\%}} & {56.7\%} & {7.93} & {7.54} & {7.37} & {7.73} & {7.51} & {8.32} & {8.21} & {\textbf{8.36}} & {8.39} \\ 

\bottomrule
\end{tabular}
\end{table*}

\begin{table*}[t]
  \small
  \caption{Ablation studies on BISON, NLVR2, and \OurEval{} analyzing the effects of training \OurModel{} with variations of \OurData{} on dialogue turn ratios and different subsets.
  The evaluation metric is accuracy for BISON and NLVR2; \OurEval{} is rated 1-10.
  }
  \label{tab:ablation_detail}
\tabcolsep=0.1cm
  \centering
\begin{tabular}{ccccccccccccc}
\toprule
\multirow{3}{*}{\textbf{Data}} &
  \multirow{3}{*}{\textbf{\begin{tabular}[c]{@{}c@{}}Turns\\ Ratio\end{tabular}}} &
  \multirow{3}{*}{\textbf{BISON}} &
  \multirow{3}{*}{\textbf{NLVR2}} &
  \multicolumn{9}{c}{\textbf{\OurEval{}}} \\ \cline{5-13} 
                                          &     &      &      & \multirow{2}{*}{\textbf{Score}} & \multicolumn{4}{c}{\textbf{Turn one}} & \multicolumn{4}{c}{\textbf{Turn two}} \\ \cline{6-13} 
                                          &     &      &      &                                 & \textbf{A1}  & C1    & C2   & C3   & \textbf{A2}  & C1    & C2   & C3   \\ \midrule
\multirow{6}{*}{\begin{tabular}[c]{@{}c@{}}SparklesDialogue\\ (CC+VG)\end{tabular}} 
  & 1:0 & \underline{57.3\%} & \underline{55.3\%} & 8.50 & 8.65 & 8.70 & 8.52 & 8.73 & 8.35 & 8.38 & 8.24 & 8.44 \\
  & 0:1 & 50.7\% & 46.7\% & 8.24 & 8.24 & 8.23 & 8.18 & 8.32 & 8.24 & 8.24 & 8.15 & 8.32 \\ 
  & 1:1 & \textbf{59.3\%} & 51.3\% & \textbf{8.73} & \textbf{8.80} & \textbf{8.83} & \underline{8.66} & \textbf{8.91} & \textbf{8.65} & \textbf{8.62} & \textbf{8.55} & \textbf{8.79} \\
  & 1:2 & 49.3\% & 51.3\% & 8.43 & 8.54 & 8.57 & 8.43 & 8.63 & 8.31 & 8.28 & 8.21 & 8.43 \\
  & \textbf{2:1} & 56.7\% & \textbf{58.0\%} & 8.56 & \underline{8.76} & \underline{8.81} & \textbf{8.67} & 8.81 & 8.35 & 8.37 & 8.28 & 8.41 \\
  & 3:1 & 50.7\% & 48.7\% & 8.45 & 8.69 & 8.74 & 8.52 & \underline{8.83} & 8.20 & 8.18 & 8.08 & 8.33 \\ \hline
SparklesDialogueCC & 2:1 & 44.7\% & 53.3\% & 8.18 & 8.26 & 8.29  & 8.16 & 8.33 & 8.10 & 8.10 & 8.00 & 8.20 \\
SparklesDialogueVG & 2:1 & 54.7\% & 52.0\% & \underline{8.59} & 8.71 & 8.76  & 8.60 & 8.78 & \underline{8.47} & \underline{8.47} & \underline{8.35} & \underline{8.60} \\ \bottomrule
\end{tabular}
\end{table*}

\section{Detailed Results of Ablation Studies}\label{section:ablation_studies_detail}
We have investigated how \OurData{}'s dialogue turn ratios and subsets impact \OurModel{}'s performance in \autoref{subsection:ablation_studies}. \autoref{tab:ablation_detail} provides the detailed results on \OurEval{}, including two dialogue turn scores and three criteria scores.

\section{\OurEval{} Details}\label{appendix:sparkleseval}

The three criteria of GPT-assisted evaluation on \OurEval{} are as follows:

\textbf{Image understanding and reasoning score} C1: Assess the assistant's proficiency in accurately identifying and describing objects, contexts, and relationships within and across the images.\\
\textbf{Cross-image and cross-turn coherence score} C2: Evaluate the assistant's ability to maintain consistent understanding across multiple images and dialogue turns.\\
\textbf{Relevance and completeness of responses score} C3: Determine the extent to which the assistant's responses are directly related to the user's inquiries and the images' content, and whether the responses provide comprehensive and detailed answers.

Following this, we ask GPT models to assign a combined score for each turn. For each model's evaluation results, we gather scores for three criteria across two turns. First, we compute the mean scores for all criteria over evaluation samples. Next, we calculate the combined scores \textbf{A1} and \textbf{A2} by averaging their respective criteria scores, namely $\textbf{A1}=mean(\text{C1},\text{C2},\text{C3})$ for the first turn and $A2=mean(\text{C1},\text{C2},\text{C3})$ for the second turn. We refrain from using the \textbf{A1} and \textbf{A2} scores provided by judge models, as their calculations may be inaccurate. Ultimately, we derive an overall \textbf{score} by averaging \textbf{A1} and \textbf{A2}.
Through this methodology, our evaluation is more holistic and interpretable.

The prompt template of GPT-assisted evaluation on \OurEval{} is presented in \autoref{tab:prompt_oureval}.

\begin{table*}[htbp]
\centering
\caption{{Prompt format for \OurEval{} evaluation}.}
\label{tab:prompt_oureval}
\begin{minipage}{\linewidth}
\centering
\begin{tcolorbox}[colback=blue!0.7!white,colframe=blue!30!black]
\centering
\small 
\begin{tabular}{p{0.95\columnwidth} c}

Users will interact with a conversational assistant. The assistant is designed to understand, analyze, and reason about multiple images across two turns of conversation. The assistant is expected to provide highly helpful and exceptionally detailed answers providing comprehensive reasoning regarding the visual content of the images. &\\
 &\\
Below are images represented by their image IDs and captions (delimited by triple quotes):
\begin{lstlisting}
```json
\end{lstlisting}
\textcolor{brickred}{\texttt{\{Target Image Descriptions\}}}
\begin{lstlisting}
```
\end{lstlisting}
 &\\
Next is a dialogue between a user and the assistant regarding the images above:
\begin{lstlisting}
```
\end{lstlisting}
\#\#\#User Q1: &\\
\textcolor{brickred}{\texttt{\{Q1\}}}
&\\
&\\
\#\#\#Assistant A1: &\\
\textcolor{brickred}{\texttt{\{A1\}}}
&\\
&\\
\#\#\#User Q2: &\\
\textcolor{brickred}{\texttt{\{Q2\}}}
&\\
&\\
\#\#\#Assistant A2: &\\
\textcolor{brickred}{\texttt{\{A2\}}}
\begin{lstlisting}
```
\end{lstlisting}
 &\\
Your task as an impartial judge is to evaluate the responses (A1 and A2) provided by the assistant to the user's questions. &\\
Please rate the following three criteria C1, C2, and C3 on a scale of 1-10 for A1 and A2 separately, where a higher score indicates better overall performance: &\\

(C1) Image Understanding and Reasoning: This measures the assistant's ability to accurately identify and describe objects, context, and relationships within and between the images. &\\
(C2) Cross-Image and Cross-Turn Coherence: This evaluates the assistant's ability to maintain a consistent understanding across multiple images and dialogue turns. &\\
(C3) Relevance and Completeness of Responses: This assesses whether the assistant's responses are directly related to the user's inquiries and the images' content, and whether the responses provide thorough, detailed answers. &\\
 &\\

Begin your evaluation by providing a short explanation for each criterion. Be as objective as possible. After providing your explanation, rate the response on a scale of 1 to 10 by strictly following the format below (note that "5" and "..." are placeholders):
\begin{lstlisting}
```
\end{lstlisting}
* Evaluating A1 &\\
- (C1) Explanation: "..." Rating: [[5]] &\\
- (C2) Explanation: "..." Rating: [[5]] &\\
- (C3) Explanation: "..." Rating: [[5]] &\\
Therefore, the overall rating of A1 is [[5]] &\\
&\\
* Evaluating A2 &\\
- (C1) Explanation: "..." Rating: [[5]] &\\
- (C2) Explanation: "..." Rating: [[5]] &\\
- (C3) Explanation: "..." Rating: [[5]] &\\
Therefore, the overall rating of A2 is [[5]]
\begin{lstlisting}
```
\end{lstlisting}

\end{tabular}
\end{tcolorbox}
\end{minipage}
\end{table*}

\section{Zero-shot evaluation on vision-language tasks}\label{appendix:evaluation_vltasks}
\begin{figure*}[t]
\centering
\includegraphics[width=0.9\textwidth]{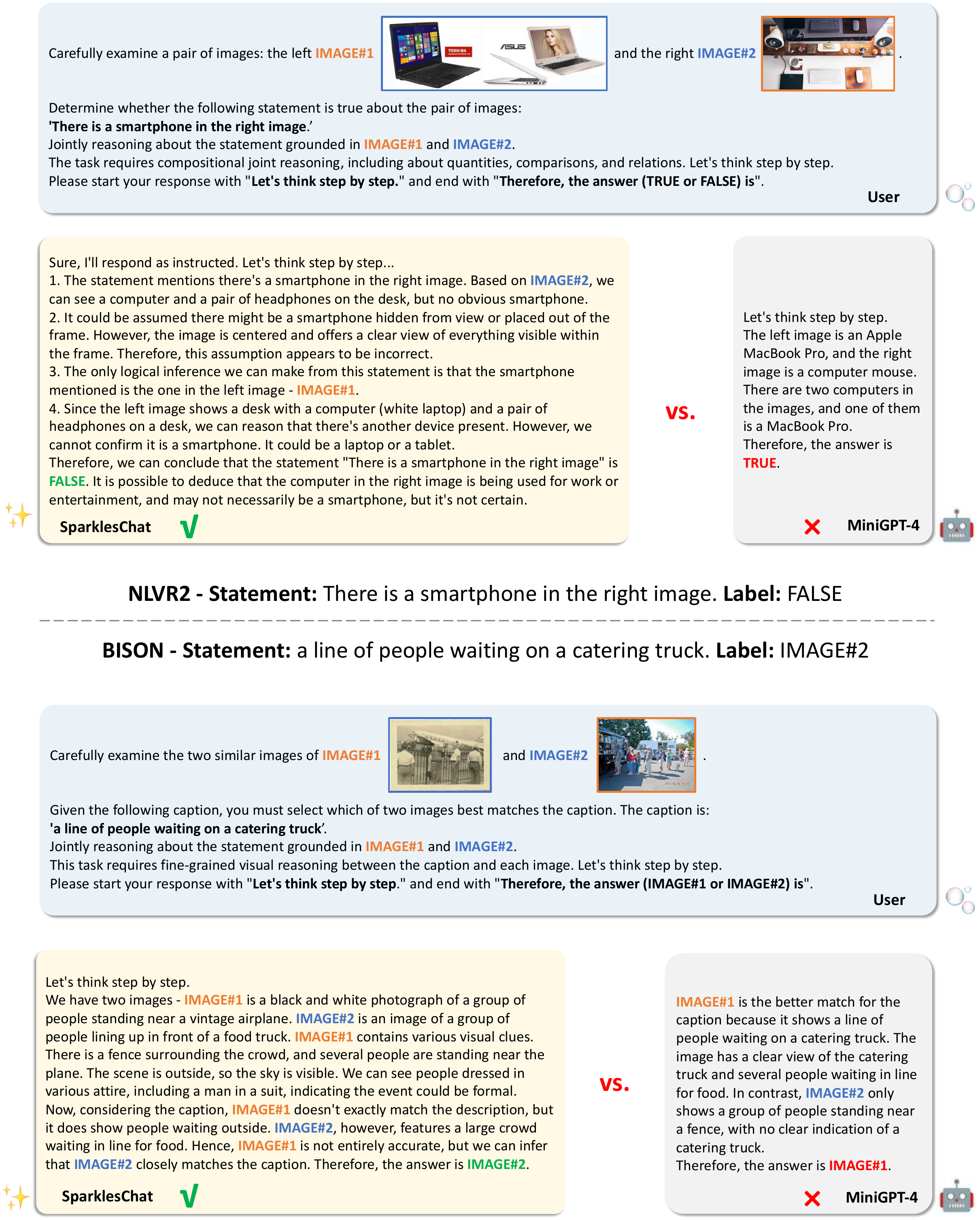}
\caption{{Comparison between \OurModel{} and MiniGPT-4 on examples of NLVR2 and BISON.}}
\label{fig:demo_VLtasks}
\end{figure*}

We chose two vision-language tasks, binary image selection and visual reasoning, to evaluate zero-shot understanding and reasoning capabilities over multiple images.
For both tasks, the evaluation metric is accuracy.
Side-by-side comparisons of example outputs for \OurModel{} and MiniGPT-4 on BISON and NLVR2 can be found in \autoref{fig:demo_VLtasks}.

\paragraph{Binary image selection on BISON.}\label{task_bison}
The Binary Image Selection task measures a model's ability to select the correct image from a pair given a text query that describes one of them~\cite{hu2019evaluating}. The model's performance is assessed in terms of binary classification accuracy. For this task, 150 examples were randomly sampled from the COCO-BISON dataset\footnote{\url{https://github.com/facebookresearch/binary-image-selection/blob/main/annotations/bison_annotations.cocoval2014.json}}.
The image source of COCO-BISON is COCO images. The image source of \OurDataCC{} is Conceptual Captions, which should have no overlap with COCO. However, our \OurDataVG{} originates from the Visual Genome, which includes a subset of COCO images. We carefully eliminate any overlapping images to ensure no overlap between the training and evaluation data.

\paragraph{Visual reasoning with natural language on NLVR2.}\label{task_nlvr}
The evaluation of the Visual Reasoning with Natural Language task assesses the model's ability to predict whether a sentence is true about a pair of images~\cite{suhr2019corpus}. This task addresses the challenge of compositional visual reasoning on relations, comparisons, and quantities. 
The NLVR2 dataset~\cite{suhr2019corpus} was used for this evaluation, with 150 examples randomly sampled from the public balanced test set\footnote{\url{https://github.com/lil-lab/nlvr/blob/master/nlvr2/data/balanced/balanced_test1.json}}.
The images in the NLVR2 dataset are sourced from Google Images, distinct from our \OurDataVG{}'s image source of the Visual Genome~\cite{krishna2017visual} and primarily feature images from Flickr.

\paragraph{Evaluation protocol and prompt design.}\label{protocol_prompt_design}
Models are evaluated on these tasks without any additional training. 
Inspired by \cite{kojima2022large}, we used a simple prompt, \textit{"Let's think step by step"}, to facilitate step-by-step reasoning before answering each question. We used the phrase \textit{"Therefore, the answer is"} to prompt the answer. Instead of using a two-stage prompting as in \cite{kojima2022large}, we combined the reasoning extraction and answer extraction stages into a single prompt: \textit{"Please start your response with 'Let's think step by step.' and end with 'Therefore, the answer is'"}. The full evaluation prompt templates to evaluate NLVR2 and BISON datasets are presented in \autoref{tab:prompt_evaluate_bison_nlvr}. We regenerated the response if the model failed to follow the instructions to output responses in the specified format. This approach ensures an unambiguous response and allows us to extract a potential answer from the text following the last occurrence of \textit{"Therefore"}.

\begin{table*}[htbp]
\centering
\caption{{Prompt formats to evaluate NLVR2 and BISON datasets.} }
\label{tab:prompt_evaluate_bison_nlvr}
\small 
\centering
\begin{tcolorbox}[colback=blue!0.7!white,colframe=blue!30!black]
\centering
\hspace{+10mm}
\begin{tabular}{p{0.95\columnwidth} c}

\VarSty{ {\bf NLVR2} } &\\

Carefully examine a pair of images: the left IMAGE\#1<Img><ImageHere></Img> and the right IMAGE\#2<Img><ImageHere></Img>. &\\
Determine whether the following statement is true about the pair of images: &\\
\textcolor{brickred}{\texttt{'\{Statement\}'}} &\\
Jointly reasoning about the statement grounded in IMAGE\#1 and IMAGE\#2. &\\
The task requires compositional joint reasoning, including quantities, comparisons, and relations. Let's think step by step. &\\
Please start your response with "Let's think step by step." and end with "Therefore, the answer (TRUE or FALSE) is". &\\

\hrulefill & \\

\VarSty{ {\bf BISON} } &\\

Carefully examine the two similar images of IMAGE\#1<Img><ImageHere></Img> and IMAGE\#2<Img><ImageHere></Img>. &\\
Given the following caption, you must select which of two images best matches the caption. &\\
The caption is: \textcolor{brickred}{\texttt{'\{Caption\}'}}. &\\
This task requires fine-grained visual reasoning between the caption and each image. Let's think step by step. &\\
Please start your response with "Let's think step by step." and end with "Therefore, the answer (IMAGE\#1 or IMAGE\#2) is". &\\

\end{tabular}
\end{tcolorbox}
\end{table*}

\section{Demonstrations and applications}\label{appendix:demo}
\begin{figure*}[htbp]
\centering
\includegraphics[width=1.0\textwidth]{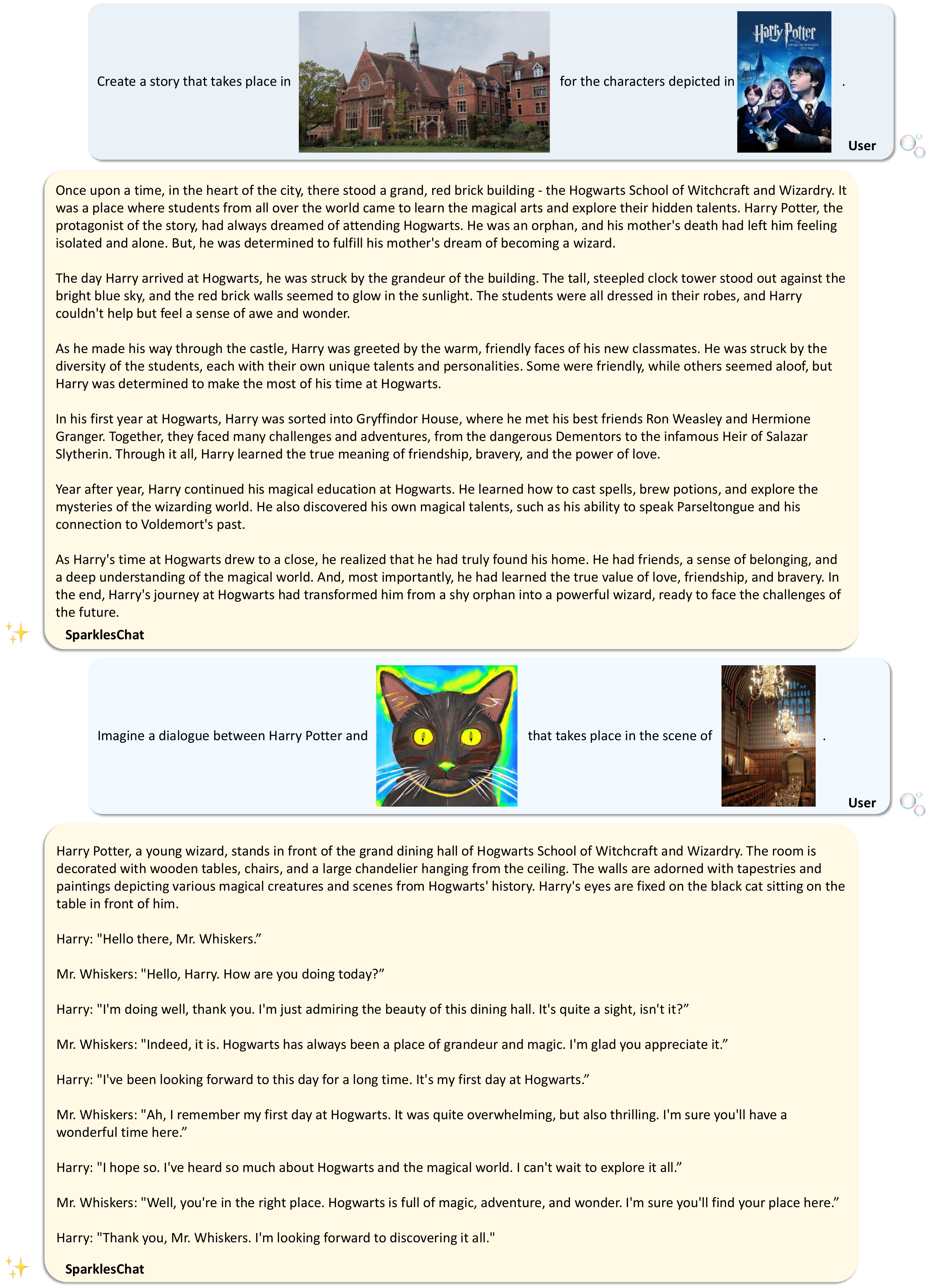}
\caption{{Demonstration of \OurModel{} to create a story and a dialogue that connects places and characters.}}
\label{fig:demo_case1}
\end{figure*}

\begin{figure*}[htbp]
\centering
\includegraphics[width=1.0\textwidth]{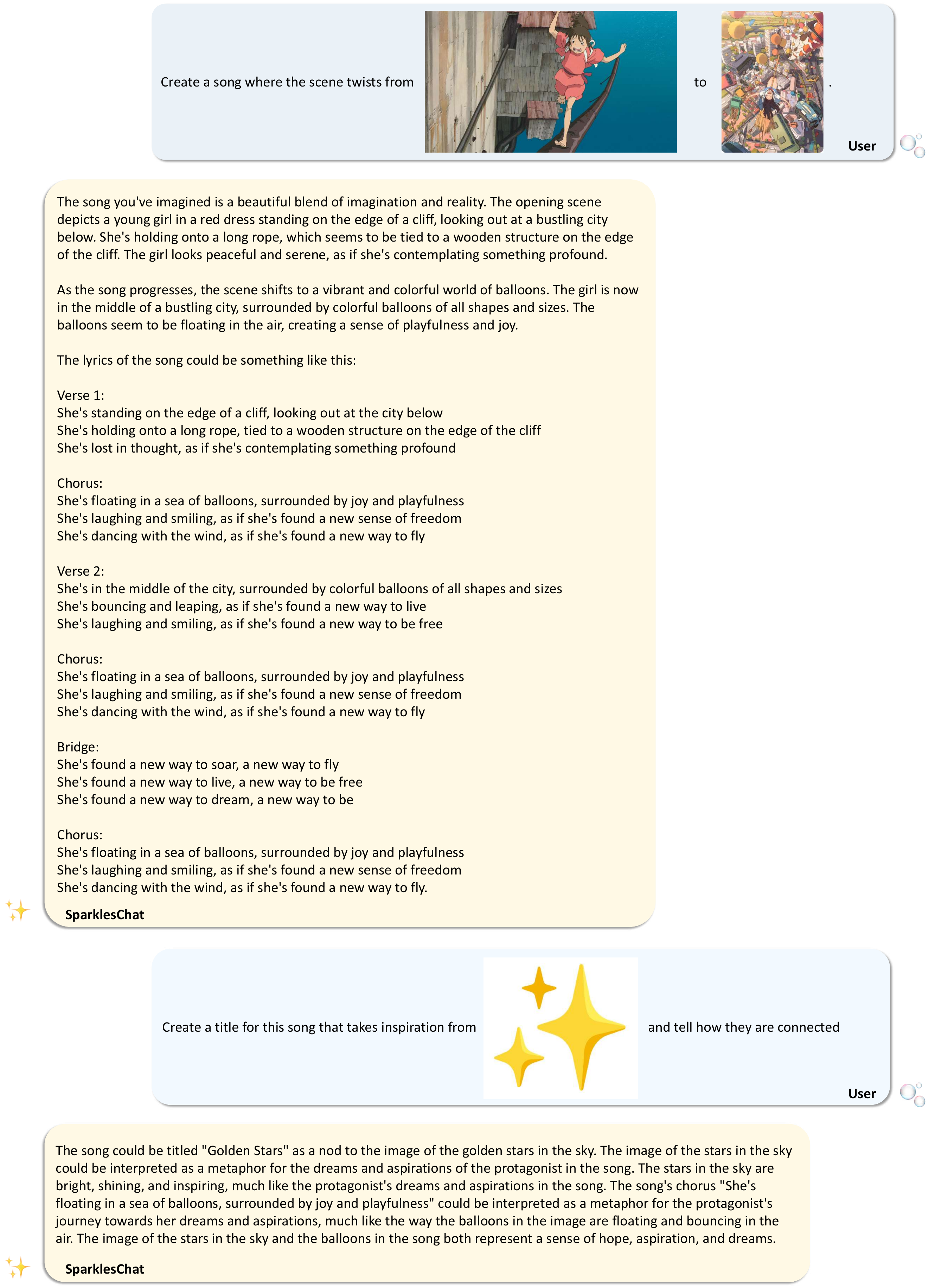}
\caption{{Demonstration of \OurModel{} to compose a song containing two scenes and generate a song title inspired by another image.}}
\label{fig:demo_case2}
\end{figure*}

\begin{figure*}[htbp]
\centering
\includegraphics[width=1.0\textwidth]{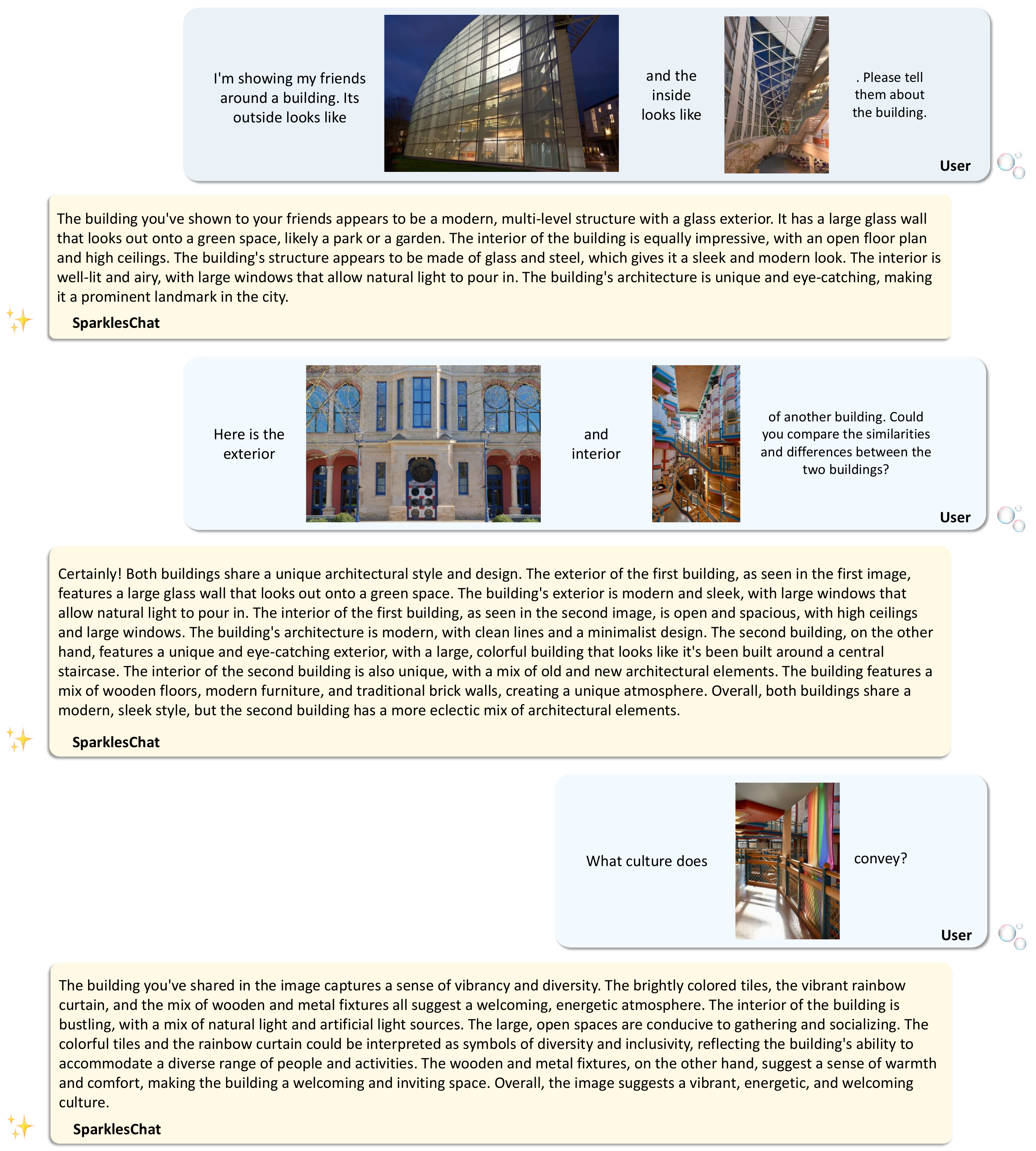}
\caption{{Demonstration of \OurModel{} to describe and reason about different groups of images.}}
\label{fig:demo_case3}
\end{figure*}

\begin{figure*}[t]
\centering
\includegraphics[width=0.98\textwidth]{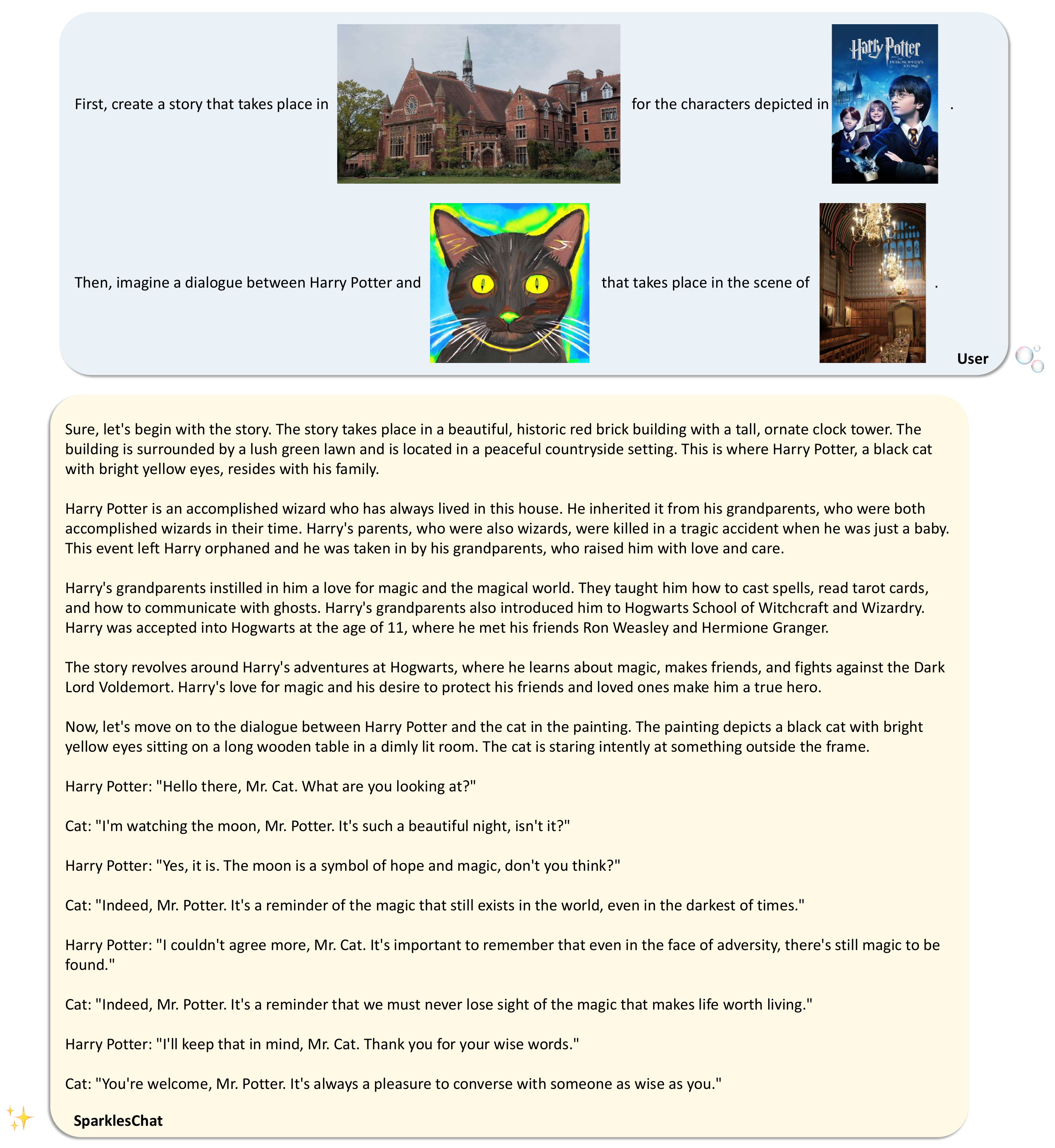}
\caption{{ Demonstration of SparklesChat to respond to a question with four images to create a story and a dialogue that connects places and characters.}}
\label{fig:demo_SparklesChat1_merge}
\end{figure*}

\begin{figure*}[t]
\centering
\includegraphics[width=0.98\textwidth]{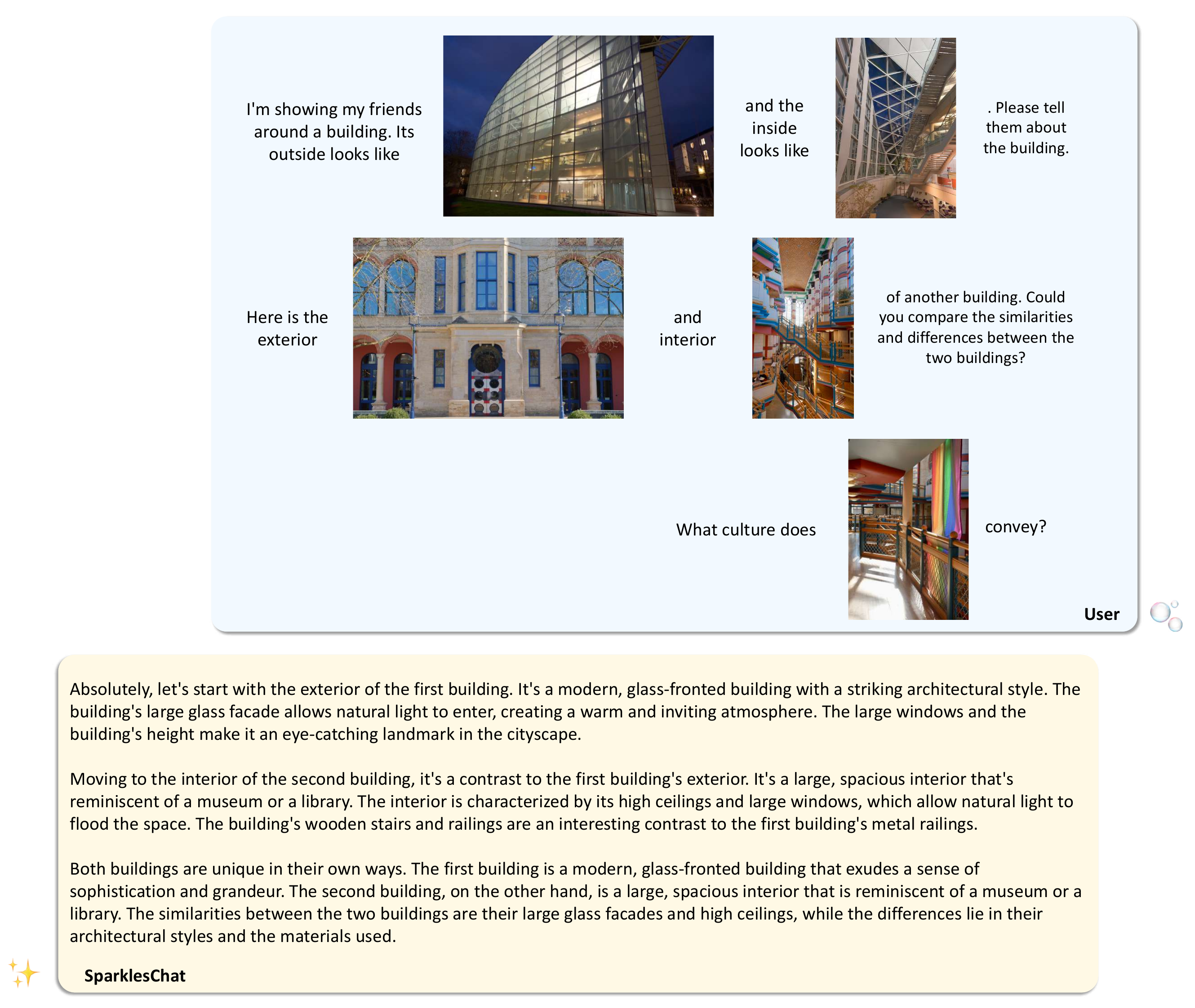}
\caption{{ Demonstration of SparklesChat to respond to a question with five images to describe and reason about different groups of images.}}
\label{fig:demo_SparklesChat3_merge}
\end{figure*}

We conducted qualitative demonstrations to showcase the model's wide applications in free-form scenarios by asking questions such as: 
``Create a story that takes place in \staricon for the characters depicted in \staricon.'', ``Imagine a dialogue between Harry Potter and \staricon that takes place in the scene of \staricon.'', ``Create a song where the scene twists from \staricon to \staricon.'', ``Create a title for this song that takes inspiration from \staricon.''.
Examples in \autoref{fig:demo_case1}, \autoref{fig:demo_case2}, and \autoref{fig:demo_case3} demonstrate two or three-turn dialogues, with each turn involving one or two images. SparklesChat is capable of distinguishing different images interleaving with text within and across dialogue turns. To analyze SparklesChat's ability to generalize in scenarios involving a larger number of images, we consolidated several questions from a dialogue into a single query. This was done to generate responses for queries involving four or five images, as shown in \autoref{fig:demo_SparklesChat1_merge} and \autoref{fig:demo_SparklesChat3_merge}, where the model effectively distinguishes between the images and adheres to complex queries. In the case of \autoref{fig:demo_SparklesChat3_merge}, where three questions involving five images are concatenated into one query, the model tends to ignore the final question and only responds to the first two. We believe this limitation arises from the absence of similar patterns in the training data. A potential solution could involve incorporating multiple turns into each training dialogue to enhance the model's ability to handle such complex scenarios.

\section{Judging with different GPT models}\label{appendix:judge_model_version}
As of July 2023, while it is widely recognized that employing \texttt{gpt-4} as a judge model outperforms alternatives such as \texttt{gpt-3.5-turbo}, the cost of using \texttt{gpt-4} is significantly higher. Therefore, we also provide scores generated by \texttt{gpt-3.5-turbo} as a reader reference, although we strongly recommend utilizing \texttt{gpt-4} or more advanced future models as reliable judges. The version of \texttt{gpt-4-0701} refers to API version \texttt{2023-07-01-preview} for the GPT-4 model. We adopt the latest version \texttt{gpt-4-0701} as our default judge model.

We evaluate GPT-4, MiniGPT-4, and \OurModel{} using \OurEval{}, leveraging three versions of judge models, as presented in \autoref{tab:judge_models_sparkleseval}. Both MiniGPT-4 and \OurModel{} generate responses based on the question and accompanying visual image. At the same time, GPT-4 is a reference LLM that only uses textual information, including the question, the ground-truth bounding boxes, and captions.
From the table, we observe that the more advanced judge models - \texttt{gpt-4-0613} and \texttt{gpt-4-0701} - provide higher scores compared to the older \texttt{gpt-3.5-turbo-0613} when assessing both GPT-4 and OurModel (approximately nine versus eight). However, these advanced judge models yield considerably lower scores for MiniGPT-4 (about three versus five).
GPT-4 achieves the highest score of 9.26 out of 10 when evaluated by the default \texttt{gpt-4-0701} mainly due to its use of detailed ground-truth annotations. Nevertheless, it's worth noting LLM judge models may display a self-enhancement bias, favoring the responses they generate~\cite{zheng2023judging}.
In contrast, MiniGPT-4 performs behind with a score of just 3.91. 
\OurModel{} achieves a score of 8.56 - about 92\% of the GPT-4 score - demonstrating \OurModel{}'s efficacy in generating responses that are not only relevant and complete but also exhibit cross-image and cross-turn coherence.

\begin{table*}[t]
  \small
  \centering
  \caption{Evaluation results on \OurEval{} with different judge models.}
  \label{tab:judge_models_sparkleseval}
\begin{tabular}{ccccccccccc}
\toprule
\multirow{2}{*}{\textbf{Model}} &
  \multirow{2}{*}{\textbf{\begin{tabular}[c]{@{}c@{}}Judge Model\\ Version\end{tabular}}} &
  \multirow{2}{*}{\textbf{Score}} &
  \multicolumn{4}{c}{\textbf{Turn one}} &
  \multicolumn{4}{c}{\textbf{Turn two}} \\ \cline{4-11} 
                           &                      &      & \textbf{A1} & C1   & C2   & C3   & \textbf{A2} & C1   & C2   & C3   \\ \midrule
\multirow{3}{*}{GPT-4}     & gpt-3.5-turbo-0613   & 8.48 & 8.61        & 8.69 & 8.18 & 8.95 & 8.35        & 8.27 & 8.08 & 8.68 \\
                           & gpt-4-0613           & 9.51 & 9.50        & 9.53 & 9.37 & 9.60 & 9.53        & 9.49 & 9.46 & 9.64 \\
                           & \textbf{gpt-4-0701} & 9.26 & 9.26        & 9.23 & 9.18 & 9.38 & 9.26        & 9.25 & 9.15 & 9.38 \\ \hline
\multirow{3}{*}{MiniGPT-4} & gpt-3.5-turbo-0613   & 5.51 & 5.46        & 6.11 & 4.78 & 5.48 & 5.55        & 5.92 & 5.23 & 5.51 \\
                           & gpt-4-0613           & 3.31 & 3.11        & 3.12 & 3.09 & 3.10 & 3.51        & 3.57 & 3.40 & 3.56 \\
                           & \textbf{gpt-4-0701} & 3.91 & 3.55        & 3.67 & 3.53 & 3.44 & 4.28        & 4.38 & 4.21 & 4.23 \\ \hline
\multirow{3}{*}{\OurModel{}} & gpt-3.5-turbo-0613   & 8.37 & 8.51        & 8.59 & 8.04 & 8.89 & 8.24        & 8.16 & 7.92 & 8.64 \\
                           & gpt-4-0613           & 8.82 & 8.75        & 8.78 & 8.59 & 8.89 & 8.88        & 8.89 & 8.79 & 8.97 \\
                           & \textbf{gpt-4-0701} & 8.56 & 8.76        & 8.81 & 8.67 & 8.81 & 8.35        & 8.37 & 8.28 & 8.41 \\ \bottomrule
\end{tabular}
\end{table*}

\section{Characteristics and verb-noun distribution analysis}\label{appendix:turn_level_verb_noun_distribution}
\begin{figure*}[t]
\small
\centering
	\begin{tabular}{c c}
\includegraphics[height=6.5cm]{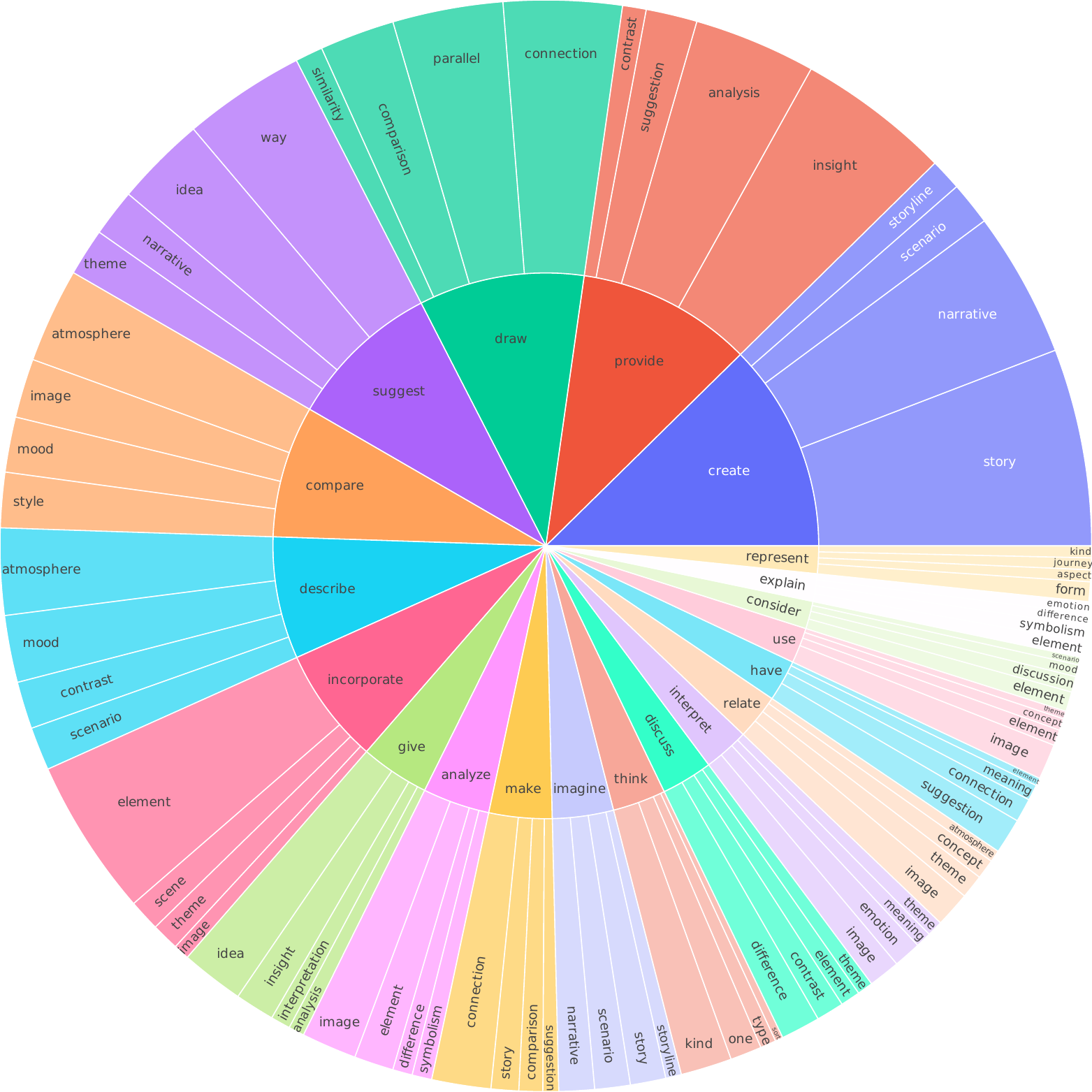} & 
\includegraphics[height=5.5cm]{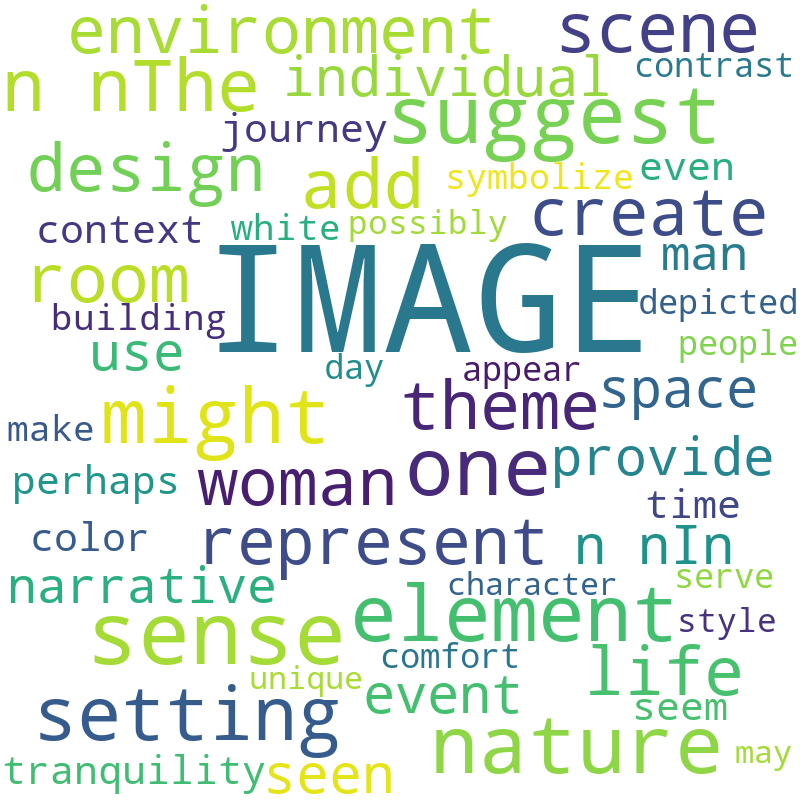}  \\
		(a) Root verb-noun pairs in user messages.
		&
		(b) Word cloud of assistant messages.  \\
\\
\includegraphics[height=4.0cm]{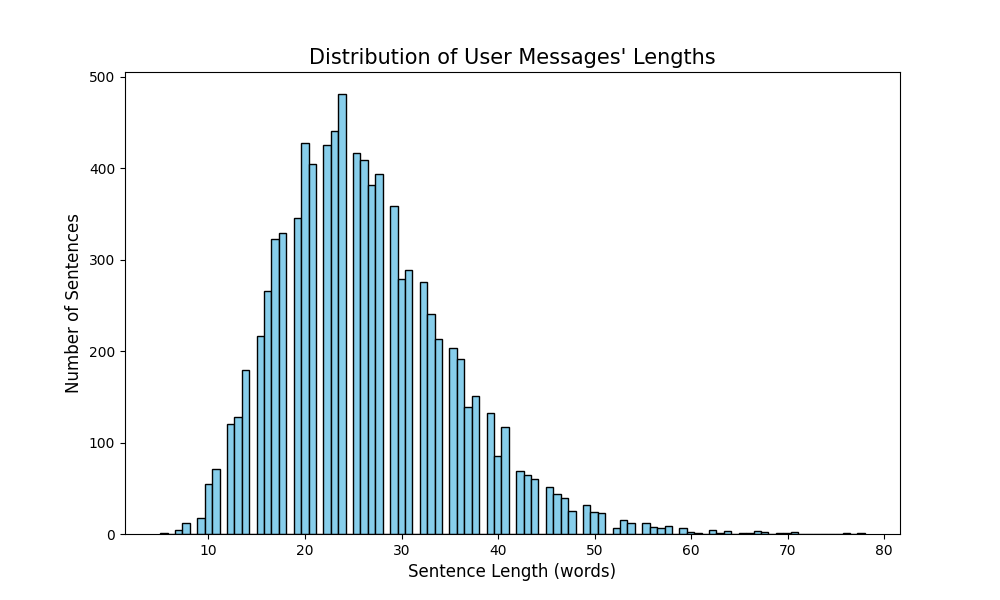} & 
\includegraphics[height=4.0cm]{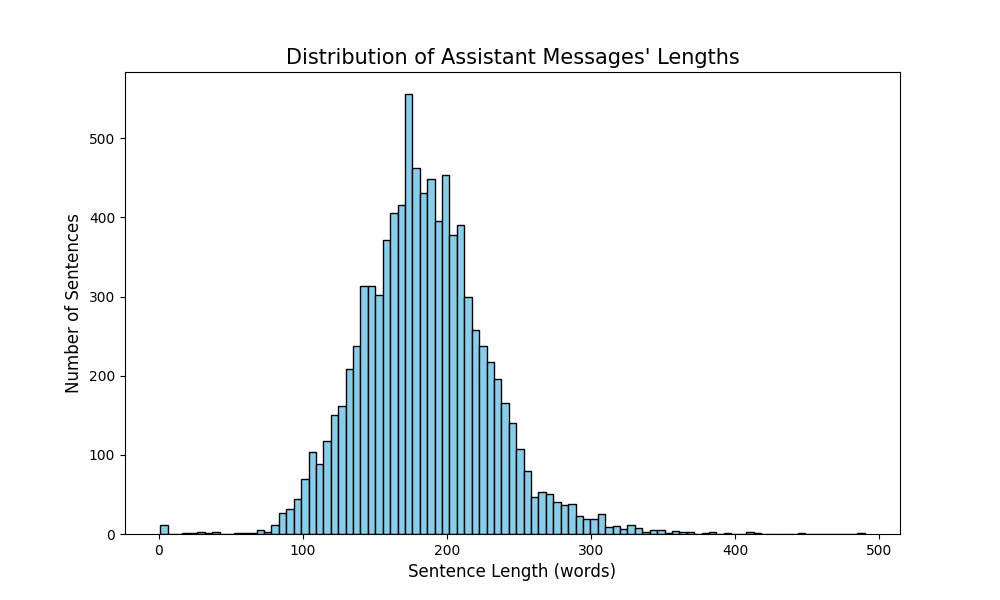}  \\
		(c) Distribution of word lengths in user
		&
		(d) Distribution of word lengths in assistant \\ messages with an average of 26.3 words.&
   messages with an average of 184.6 words. \\
  & 
   \end{tabular}
    \caption{{Characteristics of \OurDataCC{}.}}
    \label{fig:cc_data_stats}
\end{figure*}

\begin{figure*}[htbp]
	\begin{tabular}{c c}
		\hspace{-3mm}
\includegraphics[height=6.0cm]{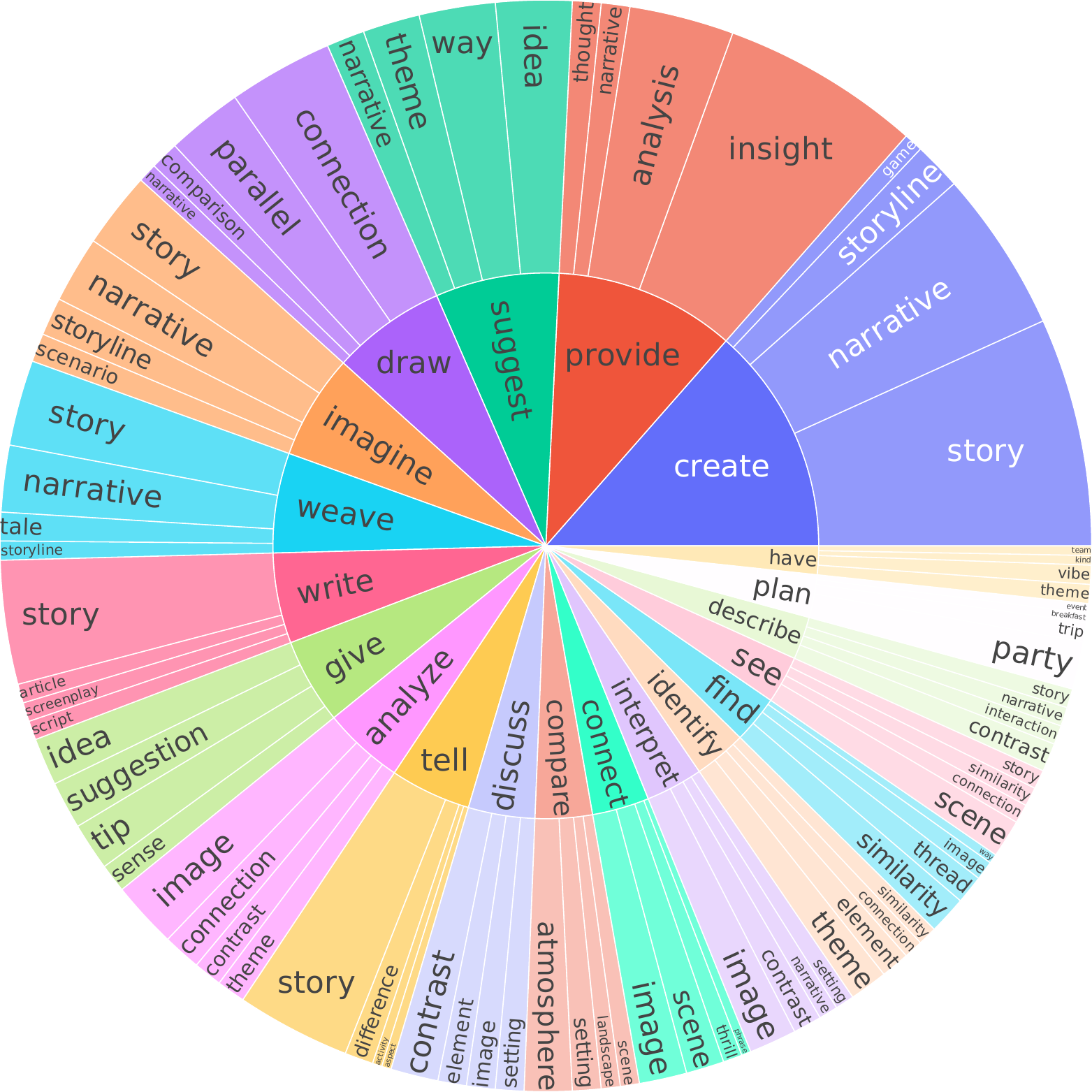} & 
\includegraphics[height=6.0cm]{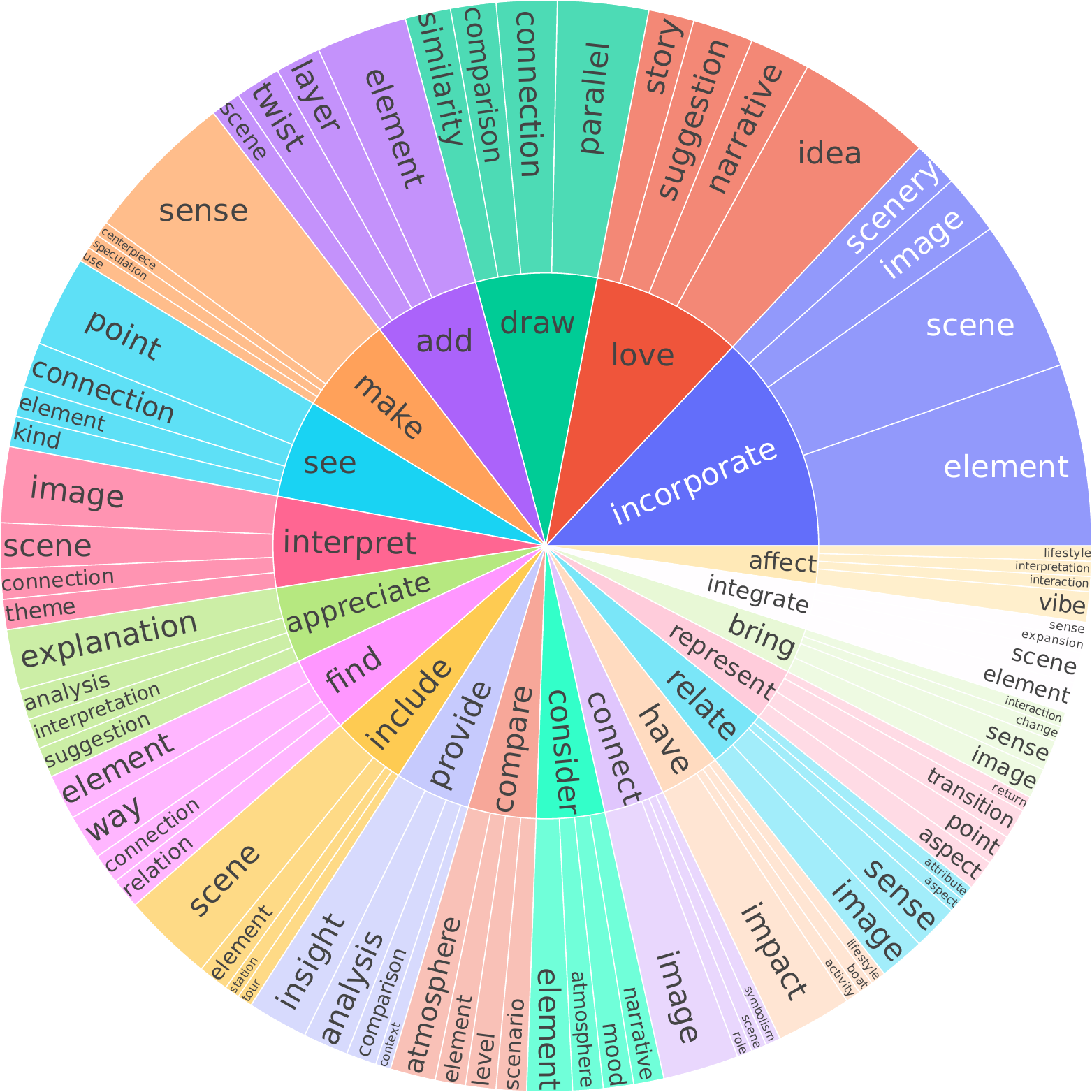}  \\
		(a) The first turn
		&
		(b) The second turn \vspace{-0mm} \\

\includegraphics[height=6.0cm]{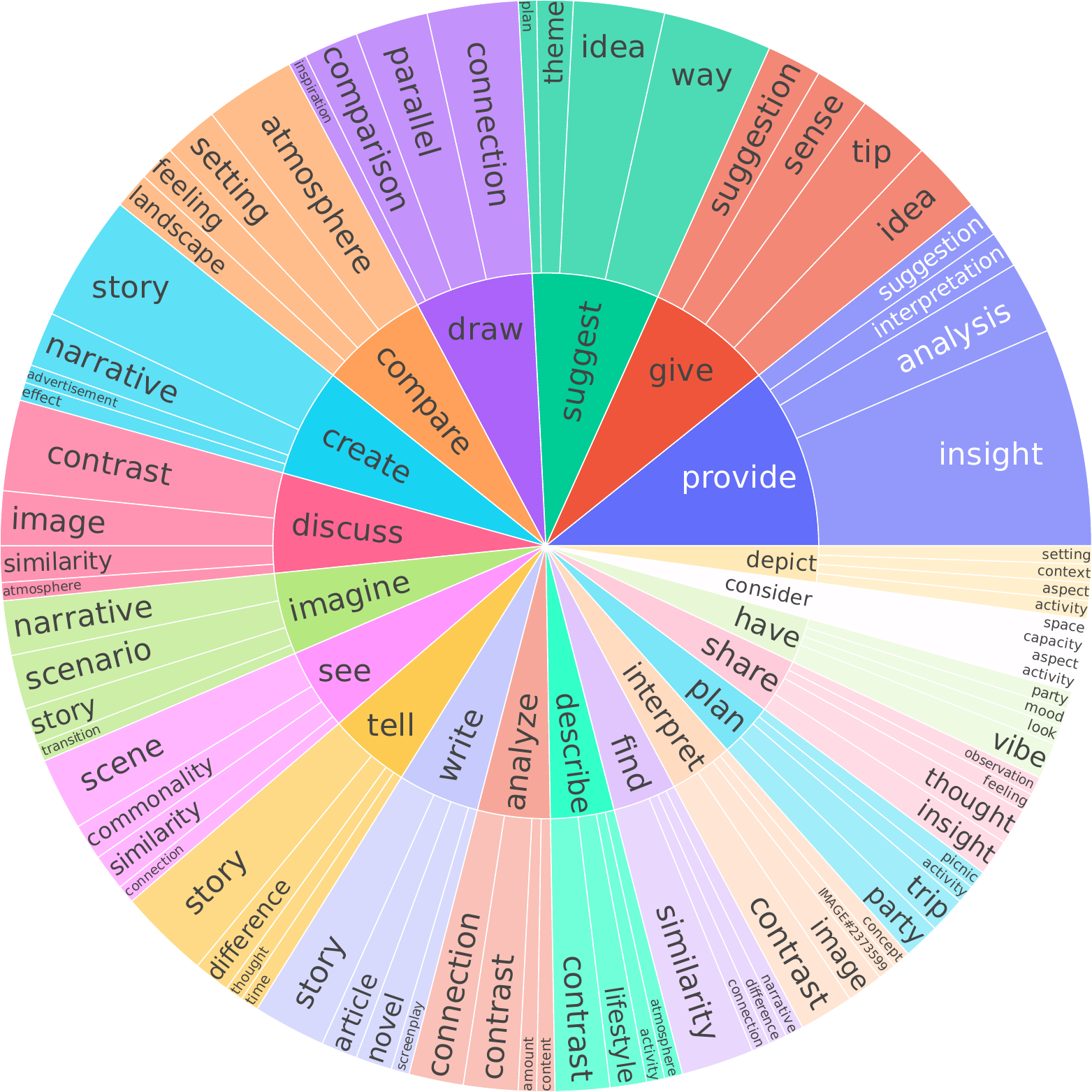} & 
\includegraphics[height=6.0cm]{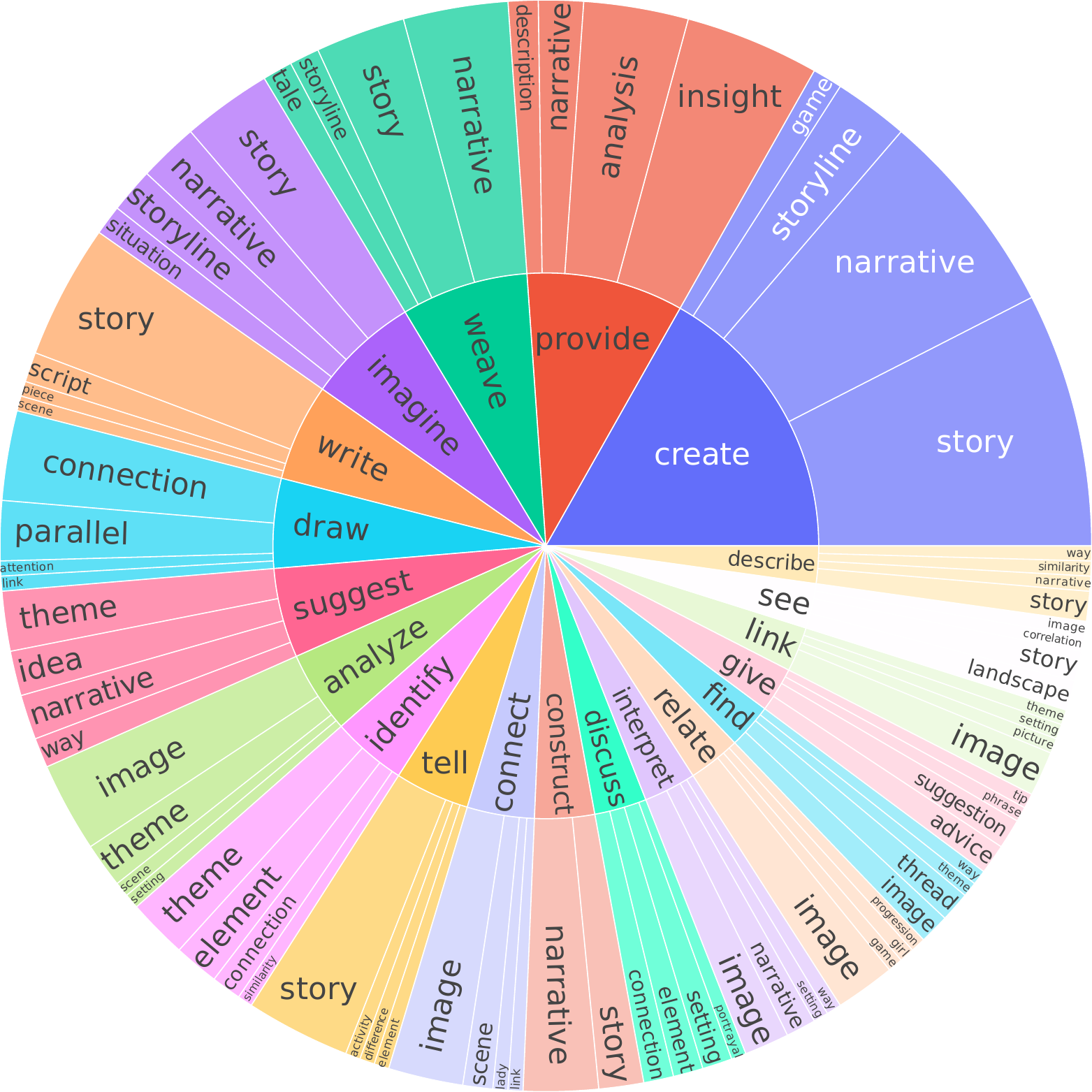}  \\
		(c) The first turn, two images
		&
		(d) The first turn, three images \vspace{-0mm} \\
  & 
   \end{tabular}
    \caption{{Root verb-noun distributions of \OurDataVG{}.}}
    \label{fig:vg_verb_noun_dist}
\end{figure*}

\begin{figure*}[htbp]
	\begin{tabular}{c c}
		\hspace{-3mm}
\includegraphics[height=6.0cm]{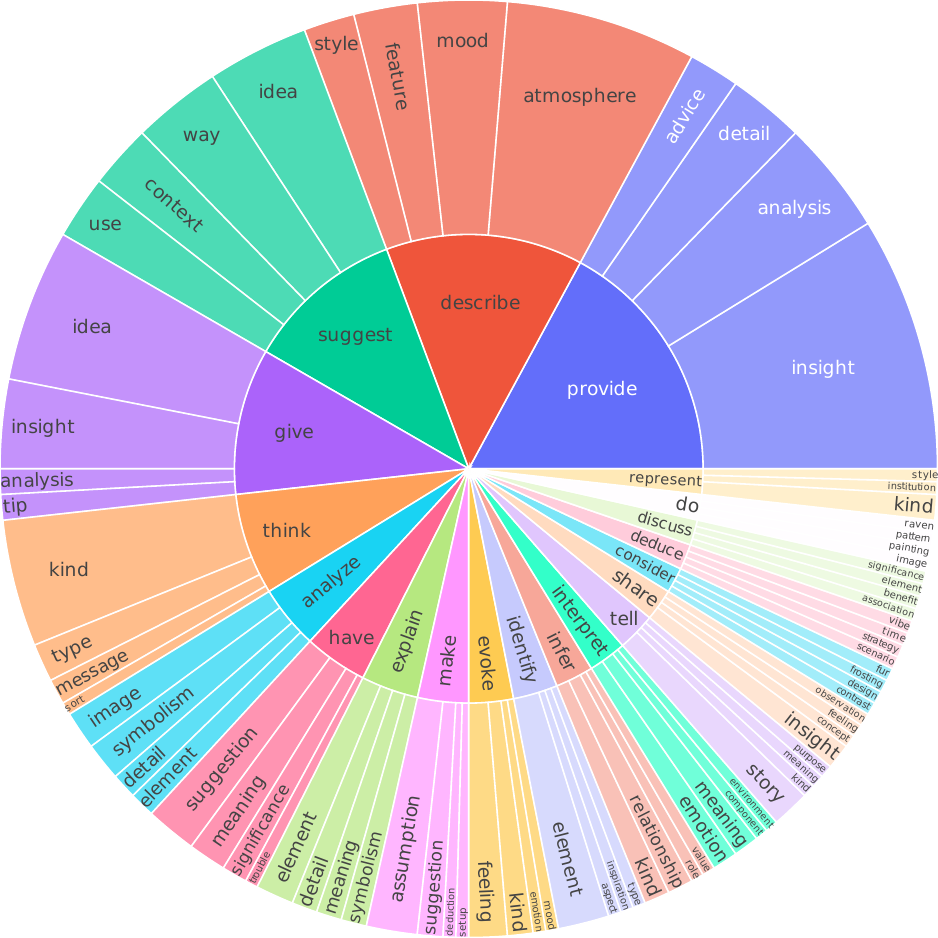} & 
\includegraphics[height=6.0cm]{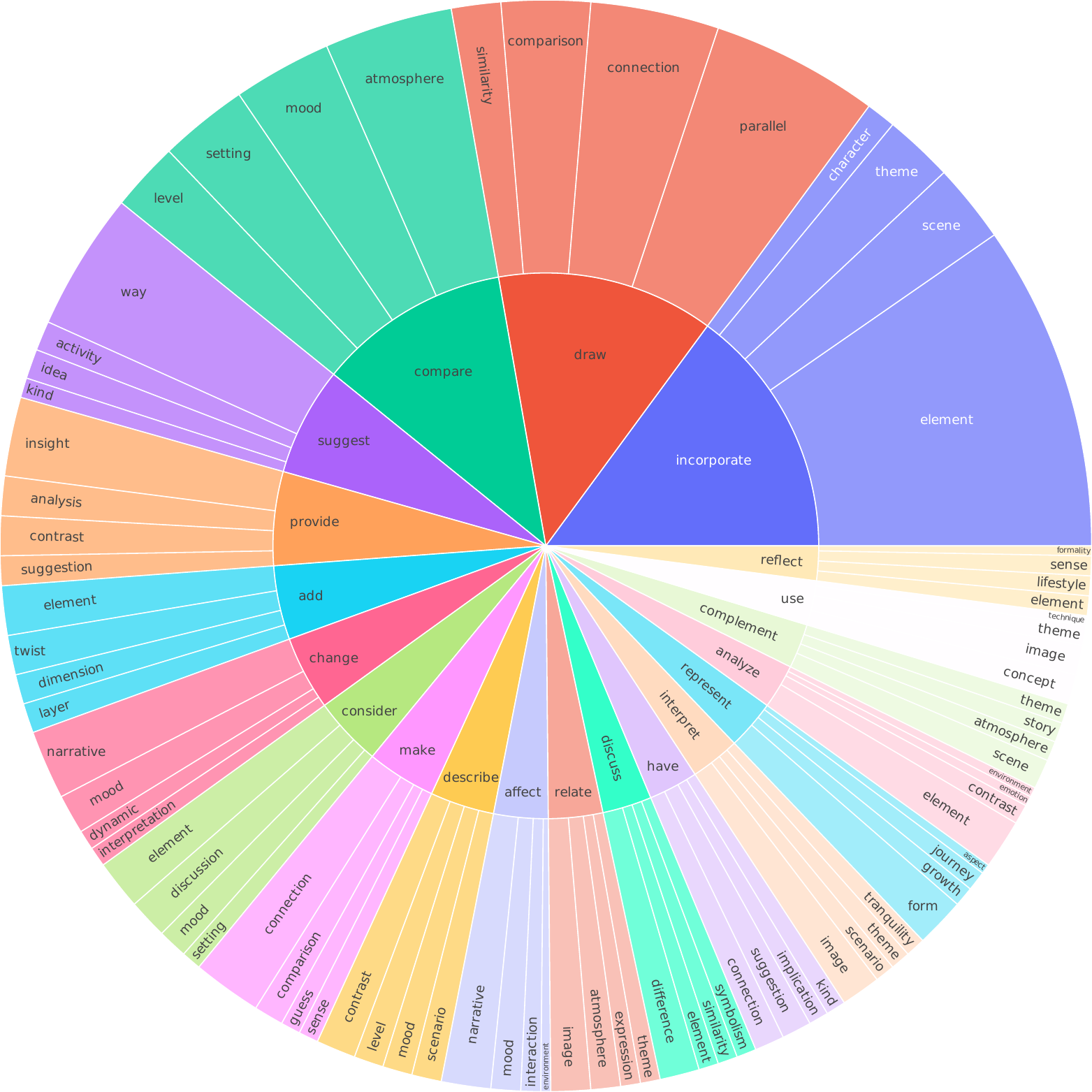}  \\
		(a) The first turn, one image
		&
		(b) The second turn \vspace{-0mm} \\

\includegraphics[height=6.0cm]{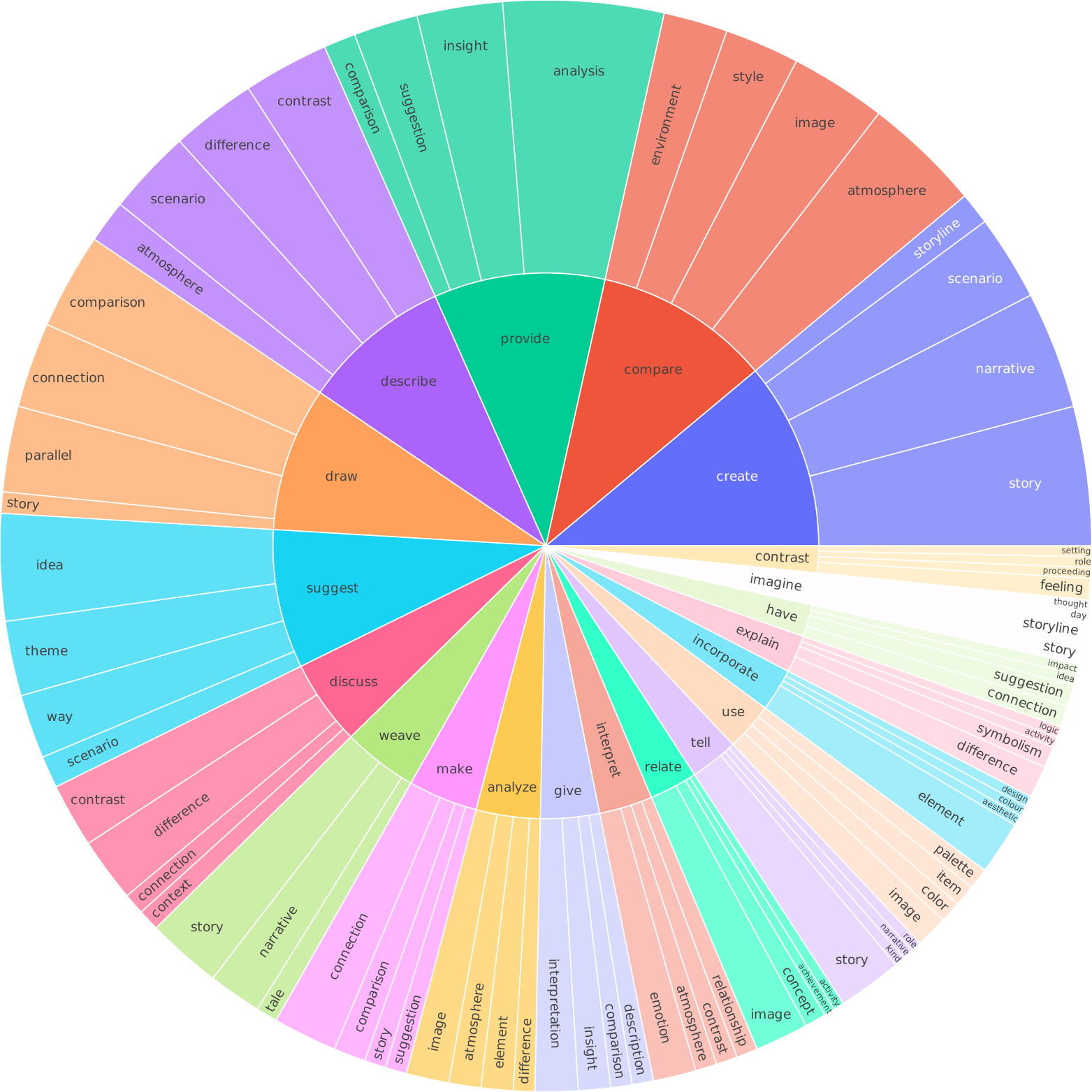} & 
\includegraphics[height=6.0cm]{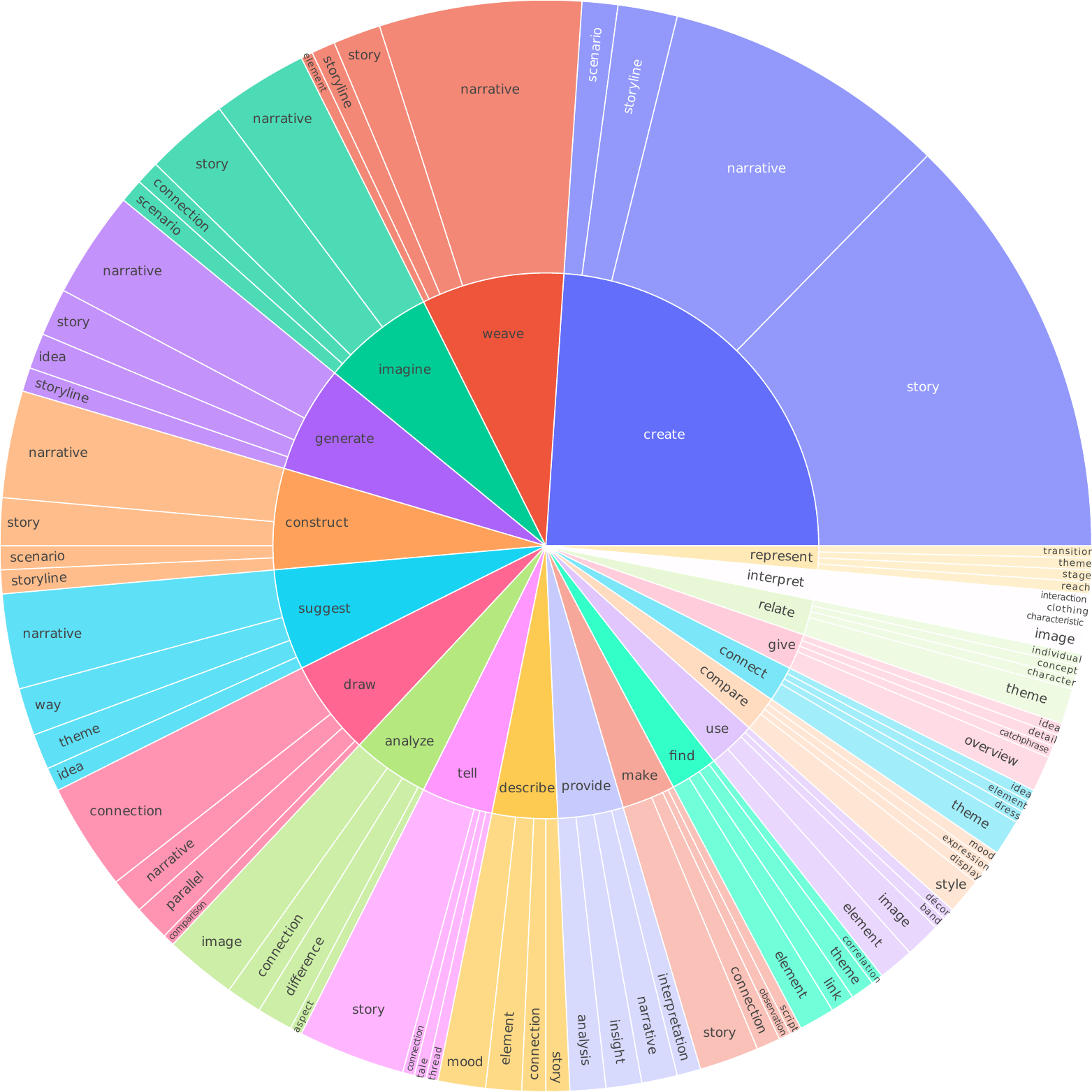}  \\
		(c) The first turn, two images
		&
		(d) The first turn, three images \vspace{-0mm} \\
  & 
   \end{tabular}
    \caption{{Root verb-noun distributions of \OurDataCC{}.}}
    \label{fig:cc_verb_noun_dist}
\end{figure*}

The characteristics of \OurDataCC{} are shown in \autoref{fig:cc_data_stats}.
For verb-noun distribution, we follow Self-instruct~\cite{wang2022self} to extract the verb closest to the root and its first direct noun object and plot the top 20 most common root verbs and their top 4 direct noun objects. 
We use the Berkeley Neural Parser\footnote{\url{https://parser.kitaev.io/}} \cite{kitaev-klein-2018-constituency} to parse user messages. We mainly focus on the last sentence of each message because it usually contains the question. If we can't extract the verb-noun pair from it, we look at the first sentence instead.
For \OurDataVG{} and \OurDataCC{}, we visualize the verb-noun distributions regarding different numbers of images in each turn in \autoref{fig:vg_verb_noun_dist} and \autoref{fig:cc_verb_noun_dist} respectively.

\section{GPT-assisted dialogue generation}\label{appendix:gpt_assisted_dialogue_generation}
\subsection{Single dialogue generation for \OurDataVG{}}\label{appendix:single_dialogue_generation}
For \OurDataVG{}, we generate one two-turn dialogue at a time, with the first turn incorporating two or three images.
We derive the demonstration dialogues from \OurDataCC{} to encourage diversity. However, to minimize redundancy, we retain only those dialogues with unique verb-noun combinations in the user questions. This results in pools of 661 and 441 demonstration dialogues for conversations incorporating two or three images in the first turn, respectively. We pull from an expansive collection of roughly 100,000 image-text pairs for this dataset. We randomly select four candidates each time, and they are not reused by excluding them from future selections.

We first present our designed \textbf{prompt} for GPT-assisted Single Dialogue Generation to generate \OurDataVG{} in \autoref{tab:prompt_single_dialogue_generation}. Then, we show a case of the \textcolor{brickred}{\texttt{Dialogue Demonstration}} and \textcolor{brickred}{\texttt{Candidate Image Descriptions}} to construct the prompt. Finally, we show the corresponding \textbf{generated dialogue} using the example prompt.

\begin{table*}[htbp]
\centering
\caption{{Prompt for GPT-assisted single dialogue generation}.}
\label{tab:prompt_single_dialogue_generation}
\begin{minipage}{\linewidth}
\centering
\begin{tcolorbox}[colback=blue!0.7!white,colframe=blue!30!black]
\centering
\small 
\begin{tabular}{p{0.95\columnwidth} c}

\VarSty{ {\bf System: You are a helpful assistant.} } &\\
Users will interact with a conversational assistant that has advanced capabilities of understanding, analyzing, and reasoning about images. This includes discussing a variety of real-world concepts, objects, and entities, generating a range of text materials, seeking advice, guidance, or assistance, and much more. &\\
 &\\
Below is an illustrative dialogue presented in a JSON format. The dialogue represents a meaningful conversation between a "user" and the "assistant" regarding multiple images. Each "user" message contains an "image\_ids" field recording the IDs of newly selected images. The images are referred to in the "content" field as IMAGE\#image\_id.

\begin{lstlisting}
```json
\end{lstlisting}
\textcolor{brickred}{\texttt{\{Dialogue Demonstration\}}}
\begin{lstlisting}
```
\end{lstlisting}
Please note that the user contents in the JSON above may be a counterexample that reveals the content of images and can be answered without looking at the images. Please make sure not to reveal the content of the images or describe the images in the user messages in the conversation that follows. &\\
 &\\
Please note that the specific "image\_ids" and "content" in the JSON above are for illustrative purposes only. The actual candidate images are shown below delimited by triple quotes, each accompanied by an image ID and a caption. Avoid using phrases similar to 'caption' and 'description' in your dialogue as if the user and the assistant have visual capabilities.

\begin{lstlisting}
```json
\end{lstlisting}
\textcolor{brickred}{\texttt{\{Candidate Image Descriptions\}}}
\begin{lstlisting}
```
\end{lstlisting}

Each dialogue consists of four messages: & \\
1. A user examines all candidate images, selects \textcolor{brickred}{\texttt{\{Number of Images\}}} highly relevant images, and sends a reasonable and creative message to the assistant. & \\
2. Once the images are provided, the assistant thoroughly perceives and comprehends them, responding with highly helpful and exceptionally detailed answers that provide comprehensive reasoning regarding the visual content of the images. & \\
3. Considering the past dialogue, the user chooses other candidate images for further inquiry. The user should refer to both the newly selected images and those mentioned earlier in the same dialogue. & \\
4. The assistant provides a highly helpful and exceptionally detailed answer providing comprehensive reasoning regarding the visual content of the images. & \\
 & \\
The following is a dialogue between the user and the assistant, adhering to the given JSON format.& \\
Make sure to formulate accurate and diverse "content" that does not follow the illustrative dialogues. And remember to develop the last "content" even though it is shown as "..." in the JSON format provided above.& \\

\hrulefill & \\

\textcolor{brickred}{\texttt{Dialogue Demonstration}}
\begin{lstlisting}[escapeinside={(*}{*)}]
[[{'role': 'user', 'image_ids': (*$\Imat^{1,1}$*), 'content': (*$\Xmat_{\texttt{q}}^{1,1}$*)}, 
  {'role': 'assistant', 'content': (*$\Xmat_{\texttt{a}}^{1,1}$*)},
  {'role': 'user', 'image_ids': (*$\Imat^{1,2}$*), 'content': (*$\Xmat_{\texttt{q}}^{1,2}$*)},
  {'role': 'assistant', 'content': '...'}]]
\end{lstlisting}

\textcolor{brickred}{\texttt{Candidate Image Descriptions}}
\begin{lstlisting}[escapeinside={(*}{*)}]
[{'image_id': (*$\Jmat^{1}$*), 'caption': (*$\Cmat^{1}$*)},
 ......
 {'image_id': (*$\Jmat^{4}$*), 'caption': (*$\Cmat^{4}$*)}]
\end{lstlisting}

\textcolor{brickred}{\texttt{Number of Images}} is "two" or "three".

\end{tabular}
\end{tcolorbox}
\end{minipage}
\end{table*}

\begin{figure*}[ht]
    \centering
    \begin{tabular}{ccc}
    \includegraphics[width=0.3\textwidth]{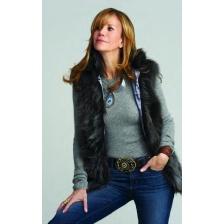} & \includegraphics[width=0.3\textwidth]{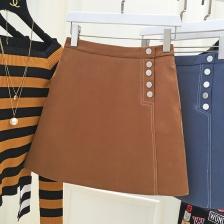} & \includegraphics[width=0.3\textwidth]{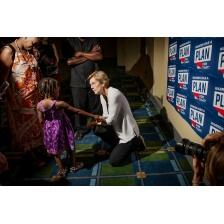} \\
    2675 & 1081 & 2679
    \end{tabular}
    \caption{Reference images corresponding to the image IDs in the demonstration dialogues in \autoref{tab:demo_dialog_in_prompt_single_dialog_gen}. These images were not sent to GPT-4 for data generation.}
    \label{fig:reference_images_in_single_demo_dialog}
\end{figure*}

\paragraph{Example of dialogue demonstration.}
\label{tab:demo_dialog_in_prompt_single_dialog_gen}
We visualize the images corresponding to image IDs in the dialogues in \autoref{fig:reference_images_in_single_demo_dialog} for reference, while these visual images were not sent to GPT-4 for data generation.
Note that we abbreviate the message content of the assistant in the second turn as ``...'' to save space, considering that the previous message contents have provided enough demonstrations. 

\begin{lstlisting}[language=json,escapechar=@]
```json
[
[
    {'role': '@\color{darkcyan}{user}@', 'image_ids': [2675, 1081], 'content': 'How would you connect the woman in IMAGE#2675 with the clothing items in IMAGE#1081?'}, 
    {'role': '@\color{forestgreen}{assistant}@', 'content': 'In IMAGE#2675, we see a woman dressed in fashionable attire consisting of a grey turtleneck sweater, blue jeans, a black leather belt, a black leather jacket with fur detailing and black ankle boots. Her overall outfit looks stylish, suggesting she has good fashion sense.\\\\n\\\\nIMAGE#1081 represents a group of clothing items hanging on a rack. It shows an outfit comprising a striped shirt, a brown skirt, and a pair of black shoes. The clothes and shoes have a sophisticated touch to them.\\\\n\\\\nConnecting these images, one could infer that the clothing items could belong to the same woman. Maybe she's someone who likes keeping her wardrobe organized or could be in the fashion industry focusing on outfit planning. Her style in IMAGE#2675 is modern and chic, while the ensemble in IMAGE#1081 is equally trendy, with classic combinations of stripes and brown, which many consider a style staple, and the elegant black shoes. Therefore, we could imagine that the outfit from IMAGE#1081 may be another stylish option that she might opt for. The clothing items could be a part of her personal style or they might signal her work attire, given the formal vibe the striped shirt and brown skirt ensemble radiate.'}, 
    {'role': '@\color{darkcyan}{user}@', 'image_ids': [2679], 'content': 'Looking at IMAGE#2679, how would you weave it into the narrative of this stylish woman from IMAGE#2675 and the clothes in IMAGE#1081?'}, 
    {'role': '@\color{forestgreen}{assistant}@', 'content': '...'}
]
]
```
\end{lstlisting}

\begin{figure*}[ht]
    \centering
    \begin{tabular}{cccc}
    \includegraphics[width=0.24\textwidth]{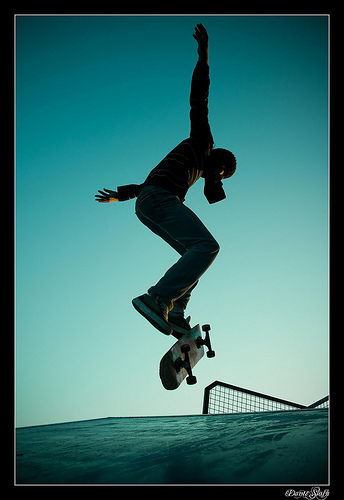} & \includegraphics[width=0.24\textwidth]{assets/images/2408549.jpg} & \includegraphics[width=0.24\textwidth]{assets/images/2330601.jpg} &
    \includegraphics[width=0.24\textwidth]{assets/images/2331159.jpg} \\
    2315390 & 2408549 & 2330601 & 2331159 \\
    \end{tabular}
    \caption{Candidate images corresponding to the image IDs in the dialogues generation process in \autoref{tab:candidate_images_descriptions_in_prompt_single_dialog_gen}. These images were not sent to GPT-4 for data generation.}
    \label{fig:candidate_images_in_single_demo_dialog}
\end{figure*}

\paragraph{Example of candidate image descriptions.}
The image IDs in the dialogues refer to the images in \autoref{fig:candidate_images_in_single_demo_dialog}.
\label{tab:candidate_images_descriptions_in_prompt_single_dialog_gen}

\begin{lstlisting}[language=json,escapechar=@]
```json
[
    {'image_id': 2315390, 'caption': 'The image captures a dynamic scene at a skate park, specifically a boy mid-air, performing a trick on his skateboard. The boy appears to be a bit on the dark side, silhouetted against the lighter sky behind him. He's adorned in casual skate attire: a sweater, blue jeans, and athletic shoes. His arms are raised in the air, which adds to the intensity of the trick he's executing. Interestingly, he is not wearing a helmet.\\nThe skateboard itself is flipped, and it appears to be on its side. This unusual positioning gives the impression that the skater is performing an intricate and complex trick. The skateboard has multiple wheels, described as black, and is noticeably detailed in the image.\\nThere's another person present in the scene, presumably a photographer or a spectator, positioned toward the right corner of the frame. However, this individual is located quite close to the edge, suggesting that they are not the main focus of the photograph.\\nThe skate park consists of a grey concrete ramp that the boy is using for his tricks. It stretches across the majority of the bottom part of the picture, a hard, flat contrast to the dynamic motion taking place above it. There's also a metal gate visible in the scene, possibly part of the boundary or safety measures at the park.\\nThe backdrop is a vibrant blue sky with clusters of white clouds scattered across it. It seems to be a bright, clear day, perfect for outdoor activities. Lastly, a safety net in the distance lends an additional element of safety to the environment.\\nOverall, the photograph encapsulates an exhilarating moment of skill, action, and athleticism at a bustling skate park, set against a serene, blue-skied day.'},
    {'image_id': 2408549, 'caption': "This image captures a dynamic scene of a large blue train moving rapidly on railroad tracks. The train's hue stands in beautiful contrast with the clear, blue sky overhead. The train is quite long, stretching almost the entire width of the image, and it appears to be well maintained, with grey stripes highlighting its design. The train's lower half is primarily filled with windows and double doors. Three windows are clearly visible, each reflecting the bright sunlight.\\nWithin the train, passengers can be seen through the windows. Notably, one person dressed in a white t-shirt is looking out of the window, taking in the scenery or perhaps observing the vehicle whose side mirror is captured in the frame. The double doors, one on the left and the other on the right, stand out on the body of the train. Each door has a number '2' inscribed on it.\\nInterestingly, in the right section of the image, the side mirror of a car is in the frame, reflecting a blurry image of another vehicle, further contributing to the sense of movement in this scene.\\nThe foreground of the image is filled with a wide expanse of green grass that contrasts nicely with the railroad tracks and a nearby road. To the right of the train, there's a tall pole that rises high into the image, likely used for mounting signs. In this case, the pole hosts a railroad crossing sign with lights and a large X on top. There is also a triangular sign with three lights underneath the X sign, providing important safety information for approaching vehicles.\\nBehind the pole, a red metal barrier is barely visible. It appears to be part of the infrastructure that surrounds the tracks. With the beautiful sunny sky overhead, this picture seems to represent a typical day with normal hustle and bustle at this railroad crossing. The sunlight reflecting off the train windows adds a stunning glow to the scene.\\nDespite the fast motion of the train, details such as the wheels and even the driver's side view mirror are captured in the image, emphasizing the skill of the photographer in capturing this dynamic and detailed snapshot of a moment in time."},
    {'image_id': 2330601, 'caption': 'This image depicts an exciting scene of a man dressed in a blue and black racing suit, riding a dirt bike on a muddy track. The man is prominently positioned in the image, seeming to occupy a considerable portion of it from left to right. His blue helmet, matching his attire, is clearly visible.\\nThe motorcycle he's racing is intricately detailed. Its prominent front and back black wheels kick up wet mud as they tear through the track, while the metallic shimmer of the exhaust and the sturdy grey frame suggest its rugged durability. A number, black in color, stands out on the side of the bike, and there's a patch of blue at the bike's back that contributes to the cohesive color scheme.\\nThe rider's attire stands out as well. Apart from the matching helmet, he's wearing a blue and orange shirt, black pants, and blue and yellow shoes. A black visor on his helmet and black gloves further accessorize his ensemble. His coat, in shades of blue and grey, fits snugly, outlining his physique.\\nThe scene around the bike is as dynamic as the rider. The track underneath is a dark brown, most likely a mix of dirt and water, suggestive of recent rain or the challenging conditions of a dirt bike race. Patches of water and water spots can be seen at various locations, indicating the wetness of the track and the splashing caused by the bike speeding through.\\nMoreover, there's an evident sense of motion in the image with water splashing up from the bike and wet sand scattering in its wake. The ground can be seen in patches, displaying its dark brown color. Amidst all this action, the bike stands as a striking subject in the image, catching the eye with its blue frame and detailings, while the rider, dressed in coordinating colors, charges forward.\\nAll of these elements combined create an image that is full of life and action, capturing a thrilling moment of a dirt bike race in progress.'},
    {'image_id': 2331159, 'caption': 'The image is lively, filled with people gathered possibly for a party or a social event. In the center of the image, a woman dressed elegantly draws attention. She stands prominently, making a distinct statement with her long, dark hair. Her face, sharply defined, features a noticeably distinct nose. She is holding a white plastic spoon in one hand, which also showcases a black wristwatch. As she raises the spoon, it's clear that she is indulging in a delicious treat, a piece of cake resting on a small plate.\\nThis cake is a stunning creation, white with red frosting. It's adorned with a delightful mix of red strawberries and an array of white candles. On the cake, there seems to be a flag as well, perhaps signifying a special occasion. An unused serving knife rests in the cake, and it appears that the woman has just served herself a piece.\\nAround her, numerous other faces peer out, all engaged in their individual conversations. Most of them appear to be men, some notable for their long hair and glasses. There's an interesting mix of attire in the scene, from yellow and white-striped shirts to red and black plaid ones. One Asian man to the right seems to be focused on the woman with the cake, adding to the collective sense of attention directed towards her.\\nIn the background, several intriguing details pop out. For instance, the twinkling lights to the right catch the eye, likely part of the party's decoration. Additionally, there is a mirror behind the woman, reflecting the attendees and amplifying the sense of a bustling crowd. A lamp stands beside a wall, casting a warm glow, while the silhouettes of patrons in the dark restaurant form an atmospheric backdrop. A window lets in some additional light, illuminating a vacant chair.\\nThe overall atmosphere conveys the joyous, friendly nature of the gathering. You can almost hear the buzz of conversation and feel the warmth of shared laughter. It's clear that this is an occasion of happiness and togetherness.'}
]
```
\end{lstlisting}

\begin{figure*}[ht]
\centering
\includegraphics[width=0.9\textwidth]{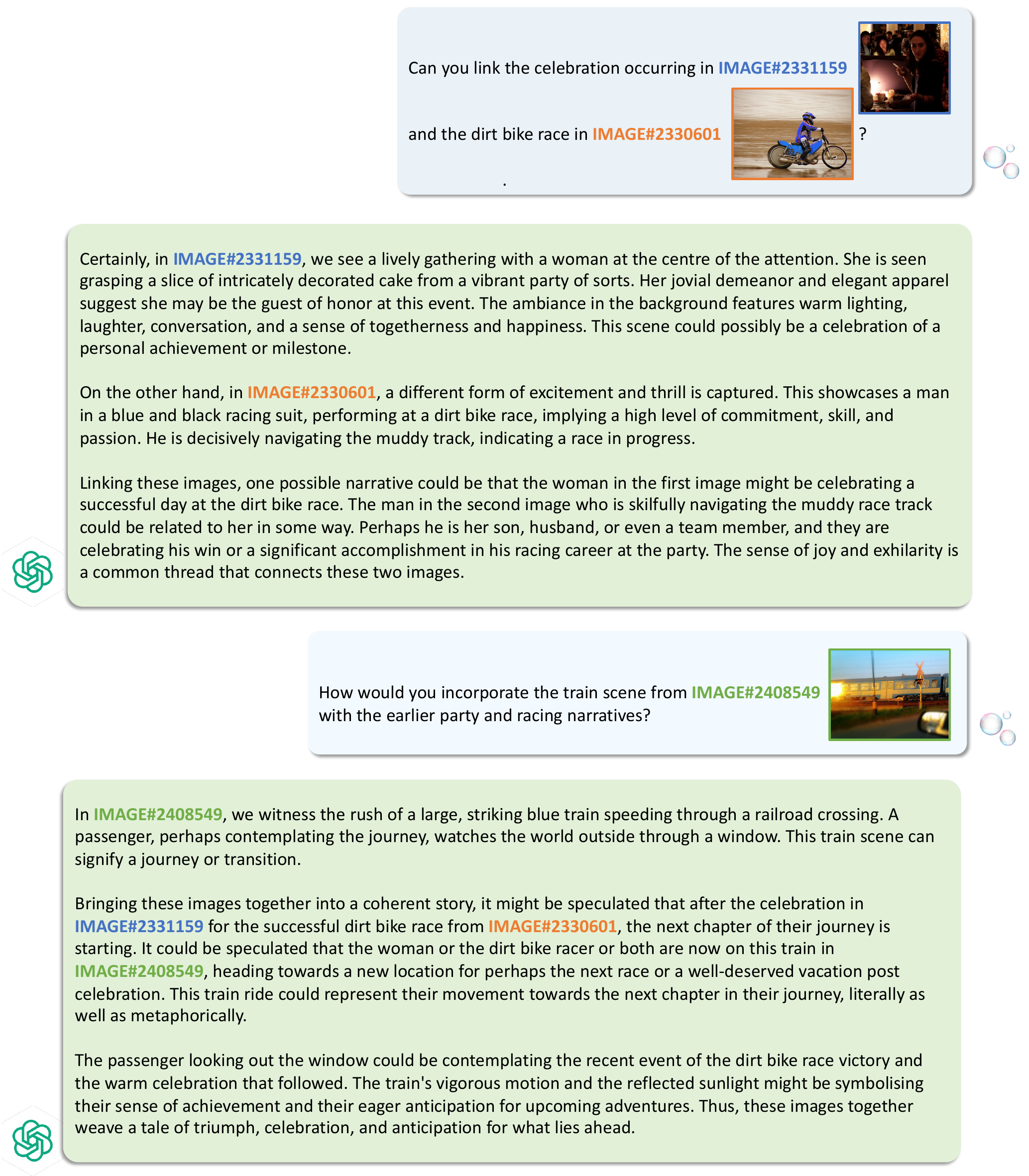}
\caption{{Dialogue example from \OurDataVG{}.} Visual images were not provided to GPT-4 during data generation but will be incorporated during \OurModel{} training.}
\label{fig:demo_SparklesDialogueVG}
\end{figure*}

\paragraph{Dialogue example from \OurDataVG{}.}
\label{tab:gen_dialog_in_prompt_single_dialog_gen}
The generated dialogue is visualized in \autoref{fig:demo_SparklesDialogueVG}. The raw text is shown as follows.
The image IDs in the dialogue refer to the images in \autoref{fig:candidate_images_in_single_demo_dialog}.

\begin{lstlisting}[language=json,escapechar=@]
```json
[
[
     {'role': '@\color{darkcyan}{user}@', 'image_ids': [2331159, 2330601], 'content': 'Can you link the celebration occurring in IMAGE#2331159 and the dirt bike race in IMAGE#2330601?'},
     {'role': '@\color{forestgreen}{assistant}@', 'content': 'Certainly, in IMAGE#2331159, we see a lively gathering with a woman at the centre of the attention. She is seen grasping a slice of intricately decorated cake from a vibrant party of sorts. Her jovial demeanor and elegant apparel suggest she may be the guest of honor at this event. The ambiance in the background features warm lighting, laughter, conversation, and a sense of togetherness and happiness. This scene could possibly be a celebration of a personal achievement or milestone.\\n\\nOn the other hand, in IMAGE#2330601, a different form of excitement and thrill is captured. This showcases a man in a blue and black racing suit, performing at a dirt bike race, implying a high level of commitment, skill, and passion. He is decisively navigating the muddy track, indicating a race in progress.\\n\\nLinking these images, one possible narrative could be that the woman in the first image might be celebrating a successful day at the dirt bike race. The man in the second image who is skilfully navigating the muddy race track could be related to her in some way. Perhaps he is her son, husband, or even a team member, and they are celebrating his win or a significant accomplishment in his racing career at the party. The sense of joy and exhilarity is a common thread that connects these two images.'},
     {'role': '@\color{darkcyan}{user}@', 'image_ids': [2408549], 'content': 'How would you incorporate the train scene from IMAGE#2408549 with the earlier party and racing narratives?'},
     {'role': '@\color{forestgreen}{assistant}@', 'content': 'In IMAGE#2408549, we witness the rush of a large, striking blue train speeding through a railroad crossing. A passenger, perhaps contemplating the journey, watches the world outside through a window. This train scene can signify a journey or transition.\\n\\nBringing these images together into a coherent story, it might be speculated that after the celebration in IMAGE#2331159 for the successful dirt bike race from IMAGE#2330601, the next chapter of their journey is starting. It could be speculated that the woman or the dirt bike racer or both are now on this train in IMAGE#2408549, heading towards a new location for perhaps the next race or a well-deserved vacation post celebration. This train ride could represent their movement towards the next chapter in their journey, literally as well as metaphorically.\\n\\nThe passenger looking out the window could be contemplating the recent event of the dirt bike race victory and the warm celebration that followed. The train's vigorous motion and the reflected sunlight might be symbolising their sense of achievement and their eager anticipation for upcoming adventures. Thus, these images together weave a tale of triumph, celebration, and anticipation for what lies ahead.'}]
]
```
\end{lstlisting}

% ---------------------------------------------------------------------------

\subsection{Multiple dialogues generation for \OurDataCC{}}\label{appendix:multiple_dialogues_generation}
For \OurDataCC{}, we prompt GPT-4 to generate three dialogues in a single response. These dialogues incorporate one, two, and three images in the first turn and a single image in the second. Each prompt includes three demonstration dialogues and nine candidate image descriptions to facilitate this. We curated 150 demonstration dialogues, evenly split with 50 dialogues for each type. The complete image-text dataset comprises about 3,500 pairs.

We first present our designed \textbf{prompt} for GPT-assisted Multiple Dialogues Generation to generate \OurDataCC{} in \autoref{tab:prompt_dialogue_generation}. Then, we show a case of the \textcolor{brickred}{\texttt{Dialogue Demonstrations}} and \textcolor{brickred}{\texttt{Candidate Image Descriptions}} to construct the prompt. Finally, we show the corresponding \textbf{generated dialogues} using the example prompt.

\begin{table*}[htbp]
\centering
\caption{{Prompt for GPT-assisted multiple dialogues generation}.}
\label{tab:prompt_dialogue_generation}
\begin{minipage}{\linewidth}
\centering
\begin{tcolorbox}[colback=blue!0.7!white,colframe=blue!30!black]
\centering
\small 
\begin{tabular}{p{0.95\columnwidth} c}

% \VarSty{ {\bf System: You are a helpful assistant.} } &\\
Users will interact with a conversational assistant that has advanced capabilities of understanding, analyzing, and reasoning about images. This includes discussing a variety of real-world concepts, objects, and entities, generating a range of text materials, seeking advice, guidance, or assistance, and much more. &\\
 &\\
Below are three illustrative dialogues presented in a JSON format. Each one represents a self-contained conversation between a "user" and the "assistant" regarding multiple images. Each "user" message contains an "image\_ids" field recording the IDs of newly selected images. The images are referred to in the "content" field as IMAGE\#image\_id.

\begin{lstlisting}
```json
\end{lstlisting}
\textcolor{brickred}{\texttt{\{Dialogue Demonstrations\}}}
\begin{lstlisting}
```
\end{lstlisting}

Please note that the specific "image\_ids" and "content" in the JSON above are for illustrative purposes only. The actual candidate images are shown below delimited by triple quotes, each accompanied by an image ID and a caption. Avoid using phrases similar to 'caption' and 'description' in your dialogue as if the user and the assistant have visual capabilities.

\begin{lstlisting}
```json
\end{lstlisting}
\textcolor{brickred}{\texttt{\{Candidate Image Descriptions\}}}
\begin{lstlisting}
```
\end{lstlisting}

Each dialogue consists of four messages: & \\
1. A user examines all candidate images, selects highly relevant ones, and sends a reasonable and creative message to the assistant. & \\
2. Once the images are provided, the assistant thoroughly perceives and comprehends them, responding with highly helpful and exceptionally detailed answers that provide comprehensive reasoning. & \\
3. Considering the past dialogue, the user chooses another candidate image for further inquiry. The user should refer to both the newly selected image and those mentioned earlier in the same dialogue. & \\
4. The assistant provides a highly helpful and exceptionally detailed answer providing comprehensive reasoning regarding the visual content of the images. & \\
 & \\
The following are three independent dialogues between the user and the assistant, adhering to the given JSON format. In this format, the first message in the three dialogues includes 1, 2, and 3 image IDs respectively. & \\
Make sure to formulate accurate and diverse "content" that does not strictly follow the illustrative dialogues. And remember to develop the last "content" even though it is shown as "..." in the JSON format provided above. & \\

\hrulefill & \\

\textcolor{brickred}{\texttt{Dialogue Demonstrations}}
\begin{lstlisting}[escapeinside={(*}{*)}]
[[{'role': 'user', 'image_ids': (*$\Imat^{1,1}$*), 'content': (*$\Xmat_{\texttt{q}}^{1,1}$*)}, 
  {'role': 'assistant', 'content': (*$\Xmat_{\texttt{a}}^{1,1}$*)},
  {'role': 'user', 'image_ids': (*$\Imat^{1,2}$*), 'content': (*$\Xmat_{\texttt{q}}^{1,2}$*)},
  {'role': 'assistant', 'content': '...'}],
 ......
 [{'role': 'user', 'image_ids': (*$\Imat^{3,1}$*), 'content': (*$\Xmat_{\texttt{q}}^{3,1}$*)},
  {'role': 'assistant', 'content': (*$\Xmat_{\texttt{a}}^{3,1}$*)},
  {'role': 'user', 'image_ids': (*$\Imat^{3,2}$*), 'content': (*$\Xmat_{\texttt{q}}^{3,2}$*)},
  {'role': 'assistant', 'content': '...'}]]
\end{lstlisting}

\textcolor{brickred}{\texttt{Candidate Image Descriptions}}
\begin{lstlisting}[escapeinside={(*}{*)}]
[{'image_id': (*$\Jmat^{1}$*), 'caption': (*$\Cmat^{1}$*)},
 ......
 {'image_id': (*$\Jmat^{9}$*), 'caption': (*$\Cmat^{9}$*)}]
\end{lstlisting}

\end{tabular}
\end{tcolorbox}
\end{minipage}
\end{table*}

\begin{figure*}[ht]
    \centering
    \begin{tabular}{ccc}
    \includegraphics[width=0.3\textwidth]{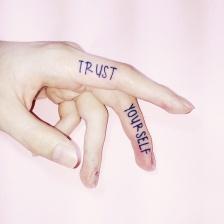} & \includegraphics[width=0.3\textwidth]{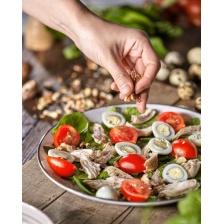} & \includegraphics[width=0.3\textwidth]{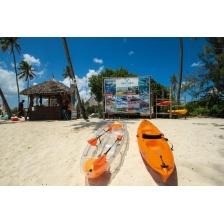} \\
    3775 & 4799 & 301 \\
    \includegraphics[width=0.3\textwidth]{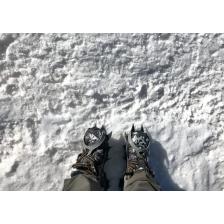} & \includegraphics[width=0.3\textwidth]{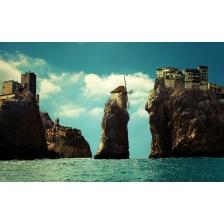} & \includegraphics[width=0.3\textwidth]{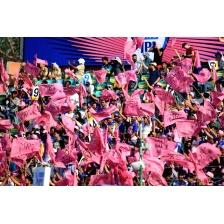} \\
    3122 & 204 & 3045 \\
    \includegraphics[width=0.3\textwidth]{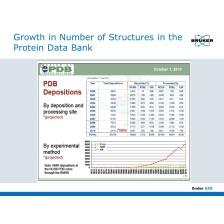} & \includegraphics[width=0.3\textwidth]{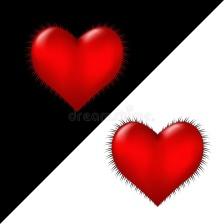} & \includegraphics[width=0.3\textwidth]{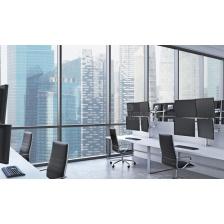} \\
    1863 & 1336 & 3302 \\
    \end{tabular}
    \caption{Reference images corresponding to the image IDs in the demonstration dialogues in \autoref{tab:demo_dialog_in_prompt_dialog_gen}. These images were not sent to GPT-4 for data generation.}
    \label{fig:reference_images_in_demo_dialog}
\end{figure*}

\paragraph{Example of dialogue demonstrations.}
\label{tab:demo_dialog_in_prompt_dialog_gen}
We visualize the images corresponding to image IDs in the dialogues in \autoref{fig:reference_images_in_demo_dialog} for reference, while these visual images were not sent to GPT-4 for data generation.
Note that we abbreviate the message content of the assistant in the second turn as ``...'' to save space, considering that the previous message contents have provided enough demonstrations. 

\begin{lstlisting}[language=json,escapechar=@]
```json
[
[
    {'role': '@\color{darkcyan}{user}@', 'image_ids': ['3775'], 'content': 'What kind of ink is usually used for such tattoos that we see in IMAGE#3775, and how long can we expect it to last?'}, 
    {'role': '@\color{forestgreen}{assistant}@', 'content': 'Tattoo artists typically use a specific type of ink known as tattoo ink for their designs, like the one shown in IMAGE#3775. Tattoo ink is usually made from metal salts, plastic-based pigments, or heavy metals for creating different colors, although black ink, like the one used in the "trust yourself" tattoo in the image, is generally carbon-based.\n\nHow long the tattoo lasts in its crisp form depends on a series of factors such as the depth of the ink placement, the type of ink used, how well it\'s cared for during the healing process, and how it\'s protected from the sun and aging. \n\nGenerally, tattoos are permanent because the human body\'s immune system isn\'t capable of dealing with the large molecules of tattoo ink. Over several years, a tattoo can fade slightly due to the natural regeneration cycle of skin cells, but it won\'t disappear completely. The good thing about black tattoos, as seen in IMAGE#3775, is that black pigment is absorbed well by all skin types and tends to last longer without fading. However, the specific longevity of a tattoo will differ from person to person. If one wants their tattoo\'s health to be maintained, they should keep it moisturized, use sunscreen when out in harsh sun, and consider touch-ups from a professional tattoo artist when necessary.'}, 
    {'role': '@\color{darkcyan}{user}@', 'image_ids': ['4799'], 'content': 'Can the ink used in the tattoo from IMAGE#3775 potentially cause any skin irritations or allergies, similar to how some people have reactions to certain foods like in IMAGE#4799?'}, 
    {'role': '@\color{forestgreen}{assistant}@', 'content': '...'}
], 
[
    {'role': '@\color{darkcyan}{user}@', 'image_ids': ['301', '3122'], 'content': 'What kind of environment is depicted in IMAGE#301 and how does it contrast with the one in IMAGE#3122?'}, 
    {'role': '@\color{forestgreen}{assistant}@', 'content': "IMAGE#301 depicts a tropical, beach environment. In the image, there are two kayaks resting on a sandy beach, and this detail hints at a location that's likely in a warmer climate, suitable for water sports and recreational activities. The presence of palm trees further supports this, as they are commonly found in regions that are hot and sunny most of the year. The blue sky and the small building that looks like a beach hut or kiosk suggest a relaxed, leisure-oriented environment.\n\nOn the contrary, IMAGE#3122 portrays a snowy, mountainous environment. The presence of snow suggests a colder climate, likely during winter. The pair of black rubber snowshoes with metal spikes on the bottom implies activities related to snow, such as hiking or mountain climbing. The person wearing a hooded jacket and grey pants with their hands in their pockets further indicates the cold nature of this environment. The trees and mountains visible through the snow complete the impression of a wilderness setting, possibly remote and less populated.\n\nIn summary, the two images contrast significantly - one depicts a warm, tropical beach setting that is likely oriented towards leisure and relaxation, while the other depicts a colder, mountainous setting associated with more strenuous outdoor activities. This contrast goes beyond just the climatic conditions; it also extends to the kind of activities and lifestyle one might associate with each setting."}, 
    {'role': '@\color{darkcyan}{user}@', 'image_ids': ['204'], 'content': 'Does the scene in IMAGE#204 remind you of the environment depicted in IMAGE#301, or are they entirely different?'}, 
    {'role': '@\color{forestgreen}{assistant}@', 'content': '...'}
],
[
    {'role': '@\color{darkcyan}{user}@', 'image_ids': ['3045', '1863', '1336'], 'content': 'Generate a promotional text that could fit for a sports event based on IMAGE#3045. Also, please include elements from IMAGE#1863 and IMAGE#1336 to give it some unique angles.'},
    {'role': '@\color{forestgreen}{assistant}@', 'content': "Join us for an electrifying display of sportsmanship like no other! Just as the fans in IMAGE#3045, cheer on your favorite team in an outdoor stadium filled with a roaring crowd, all decked out in their team's gear, waving pink flags in unity and excitement. Experience an atmosphere buzzing with energy beneath the open sky, surrounded by the vibrant greenery.\n\nDrawing inspiration from IMAGE#1863, watch as the team's performance graph soars high, much like the increasing deposits in a bank account. Witness the thrilling trajectory of action unfold, as each second, each decision, contributes to an upward trend in their game. Be a part of this journey, as they race not just to maintain the pace but to set a new norm, akin to the decreasing withdrawals mark, showcasing improvement and growth over time.\n\nIn the spirit of IMAGE#1336, let your hearts beat as one, with the passion of the red and the harmony of the black and white. Let the dual celebrations of vibrant love for the game and steadfast unity resound in the air. Let the duality of a heartbeat stir a symphony of balance - just like the yin and yang of competitiveness and sportsmanship.\n\nSo, let's delve into this exciting world, where not just the players, but even the spectators embark on a journey of passion, unity, growth, and harmony. Let the games begin!"},
    {'role': '@\color{darkcyan}{user}@', 'image_ids': ['3302'], 'content': 'Could the atmosphere and energy captured in IMAGE#3045 somehow be invoked in the office space depicted in IMAGE#3302 to make it more lively? Any suggestions?'}, 
    {'role': '@\color{forestgreen}{assistant}@', 'content': '...'}
]
]
```
\end{lstlisting}

\begin{figure*}[ht]
    \centering
    \begin{tabular}{ccc}
    \includegraphics[width=0.3\textwidth]{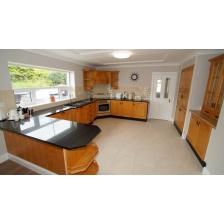} & \includegraphics[width=0.3\textwidth]{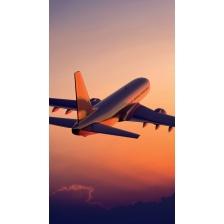} & \includegraphics[width=0.3\textwidth]{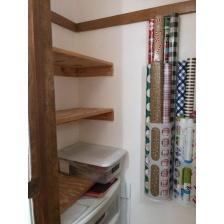} \\
    2439 & 3065 & 1093 \\
    \includegraphics[width=0.3\textwidth]{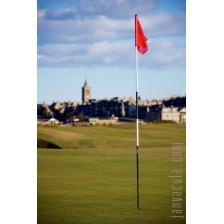} & \includegraphics[width=0.3\textwidth]{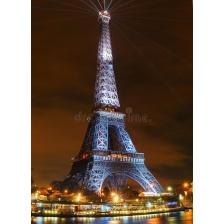} & \includegraphics[width=0.3\textwidth]{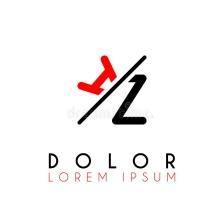} \\
    4704 & 1438 & 3120 \\
    \includegraphics[width=0.3\textwidth]{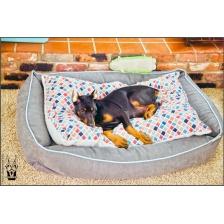} & \includegraphics[width=0.3\textwidth]{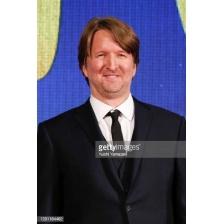} & \includegraphics[width=0.3\textwidth]{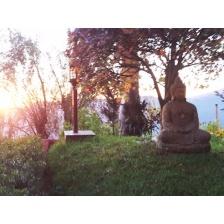} \\
    2630 & 2071 & 2966 \\
    \end{tabular}
    \caption{Candidate images corresponding to the image IDs in the dialogues generation process in \autoref{tab:candidate_images_descriptions_in_prompt_dialog_gen}. These images were not sent to GPT-4 for data generation.}
    \label{fig:candidate_images_in_demo_dialog}
\end{figure*}

\paragraph{Example of candidate image descriptions.}
An example of Candidate Image Descriptions is shown below, and their corresponding source images are shown in \autoref{fig:candidate_images_in_demo_dialog} for reference (they are not sent to GPT-4).
\label{tab:candidate_images_descriptions_in_prompt_dialog_gen}

\begin{lstlisting}[language=json,escapechar=@]
```json
[
    {'image_id': '2439', 'caption': 'This image shows a kitchen with wooden cabinets, black countertops, and white appliances. The floor is made of tiles and the walls are painted white. There is a large window above the sink that lets in plenty of natural light. The room is spacious and well lit.'},
    {'image_id': '3065', 'caption': "This is an image of an airplane flying in the sky at sunset. The plane is a large, commercial jet with a white body and red and blue stripes on the tail. It is flying low in the sky, with the sun setting behind it, casting a warm orange glow on the left side of the image and a blue glow on the right. The plane's engines are visible at the bottom of the image, with smoke coming from them. The sky is a deep blue, with clouds in the distance that are tinged with pink from the sunset."},
    {'image_id': '1093', 'caption': 'The image shows a small room with a wooden shelf on the wall, several rolls of wrapping paper stacked on it, a door on the right side, and a window on the left side. The walls are painted white and there is a wooden floor.'},
    {'image_id': '4704', 'caption': 'The image shows a view of a golf course with a red flag on the green. In the background, there is a city skyline with buildings and a church steeple. The grass on the course is lush and green, and there are trees on either side of the fairway. The sky is clear and blue, and there are a few clouds in the distance. The flag on the green is a small, red flag with a white pole. It is standing upright in the middle of the green, and it looks like it is blowing in the wind. The city skyline in the background is quite impressive, with several tall buildings and a church steeple visible. The church steeple is quite tall and has a pointed top.'},
    {'image_id': '1438', 'caption': 'The Eiffel Tower is a famous landmark in Paris, France. It is a wrought iron lattice tower that was built in 1889 to commemorate the centenary of the French Revolution. The tower stands 324 meters tall and is located on the Champ de Mars in the heart of Paris. It is one of the most visited tourist attractions in the world, with millions of people visiting it every year. The tower has become an iconic symbol of Paris and France. The tower is painted in blue, white, and red, the colors of the French flag. The lights on the tower create a beautiful and magical atmosphere, making it a popular spot for romantic walks and photos. The tower is surrounded by water, with a river running underneath it.'},
    {'image_id': '3120', 'caption': 'This image is a logo for a company or brand. The logo consists of the letters "z" and "l" in a red and black color scheme. The letters are connected by a diagonal line through the center of the image, creating a simple and modern design suitable for various businesses and industries.'},
    {'image_id': '2630', 'caption': "The image shows a black and tan dog lying on a large, gray dog bed. The bed has a geometric pattern on it and is placed in front of a brick fireplace. The dog's head is resting on its paws, and it appears to be sleeping. The image is well lit and the details of the dog and the bed are clear."},
    {'image_id': '2071', 'caption': "The image shows a man in a black suit and tie standing in front of a blue and yellow background. The man is smiling and has his hands in his pockets.\n\nThe man is wearing a black suit and tie, with a white shirt underneath. He has short, dark hair and a well-groomed beard. He is standing in front of a large blue and yellow background that looks like a cartoon character's face. The character's eyes and mouth are wide open, and it looks like the man is standing in front of the character's face.\n\nThe background looks like a cartoon character's face, with blue and yellow colors. The character's eyes and mouth are wide open, and it looks like the man is standing in front of the character's face. The character's eyes and mouth are wide open, and it looks like the man is standing in front of the character's face."},
    {'image_id': '2966', 'caption': 'The image shows a garden with a stone statue of a person sitting under a tree. The statue has a serene expression on its face, and the sun is setting behind the trees, casting a warm orange light on the scene. The foreground grass is green and well-maintained, and there are no other objects or people in the image.'}
]
```
\end{lstlisting}

\begin{figure*}[ht]
\centering
\includegraphics[width=0.9\textwidth]{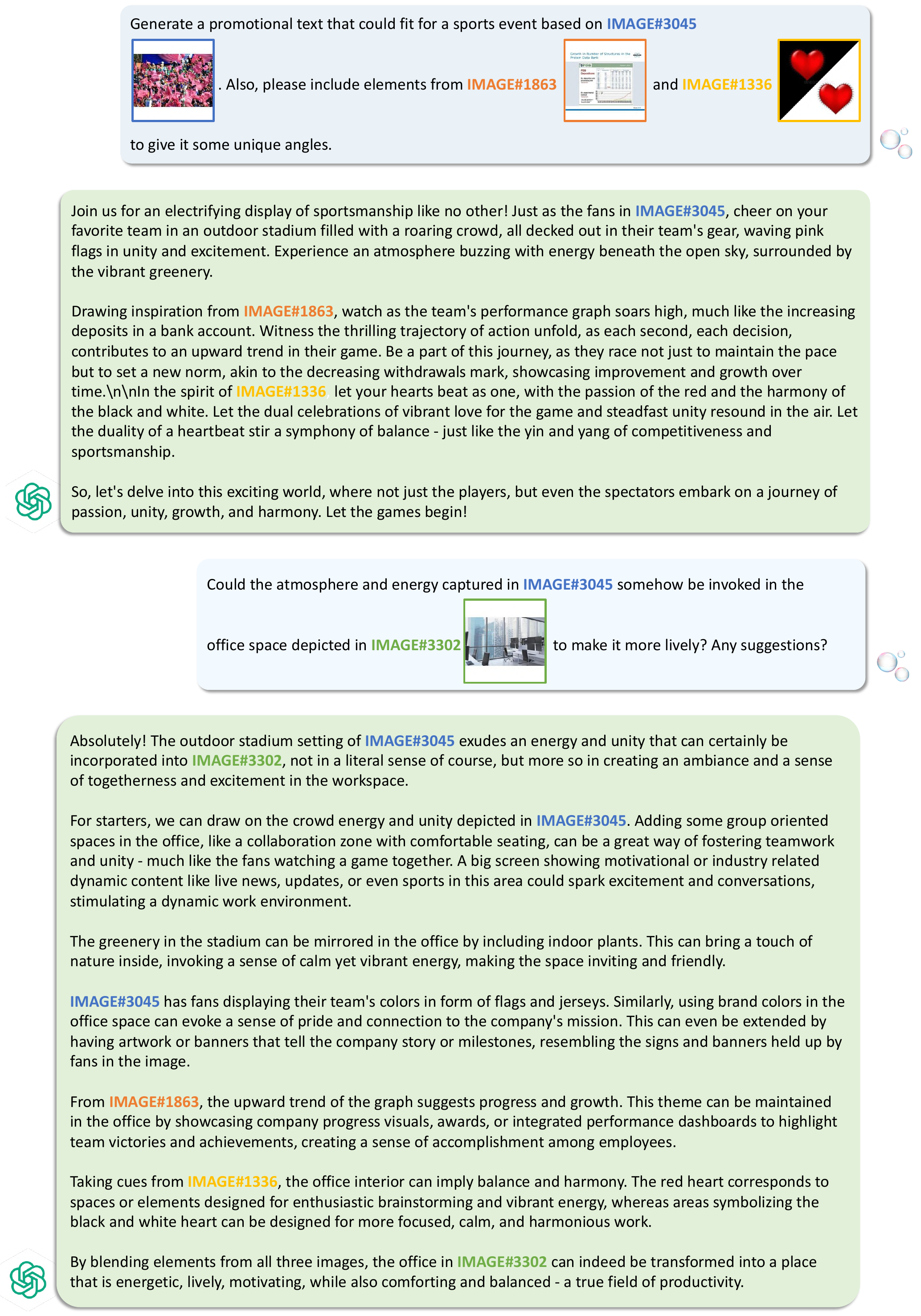}
\caption{{Dialogue example from \OurDataCC{}.} Visual images were not provided to GPT-4 during data generation but will be incorporated during \OurModel{} training.}
\label{fig:demo_SparklesDialogueCC}
\end{figure*}

\paragraph{Dialogue examples from \OurDataCC{}.}
\label{tab:gen_dialog_in_prompt_dialog_gen}
The generated dialogue is visualized in \autoref{fig:demo_SparklesDialogueCC}. The raw text is shown as follows.
The image IDs in the dialogues refer to the images in \autoref{fig:candidate_images_in_demo_dialog}.

\begin{lstlisting}[language=json,escapechar=@]
```json
[
[
     {'role': '@\color{darkcyan}{user}@', 'image_ids': ['3120'], 'content': 'Could you tell me what the design of the logo in IMAGE#3120 might symbolize or represent?'},
     {'role': '@\color{forestgreen}{assistant}@', 'content': 'The logo in IMAGE#3120, with the combined letters "z" and "l" connected by a diagonal line, seems to convey simplicity, modernity, and approachability. The use of "z" and "l" could represent the initials of a company or individual\'s name, while the diagonal line may symbolize a link or connection, emphasizing unity or collaboration. The choice of red and black colors could be conveying a sense of power and sleek sophistication, respectively. Black often symbolizes solidity and formality, while red can represent passion and energy. However, the specific meaning could vary as logo interpretations often depend on the organization\'s mission, brand values, and the industry it\'s part of.'},
     {'role': '@\color{darkcyan}{user}@', 'image_ids': ['1438'], 'content': 'Considering the design and message one might infer from the logo in IMAGE#3120, would it be a good fit for the place depicted in IMAGE#1438? How would they complement each other?'},
     {'role': '@\color{forestgreen}{assistant}@', 'content': 'The logo from IMAGE#3120 can bring a modern, energetic, and connected vibe to the traditional and iconic setting of the Eiffel Tower in IMAGE#1438. This can create an interesting juxtaposition between historic and contemporary themes.\n\nThe simplicity and modernity of the logo design would sharply contrast yet harmonize with the complexity and intricacy of the Eiffel Tower. This could appeal to audiences who appreciate both the historical and futuristic aspects of a brand. Similarly, the vibrancy and energy indicated by the red color in the logo may align well with the romantic and lively atmosphere associated with Paris, the city where Eiffel Tower is situated.\n\nIf this was for a brand that intended to evoke a sense of modern connection in a place steeped with history, or perhaps a company that combines technology with tourism in Paris, this logo could indeed complement the place effectively.'}], 
[
     {'role': '@\color{darkcyan}{user}@', 'image_ids': ['2630', '3065'], 'content': 'Can you imagine a story that might link the peaceful dog lounging in IMAGE#2630 and the airplane soaring in the sunset sky of IMAGE#3065?'},
     {'role': '@\color{forestgreen}{assistant}@', 'content': 'Certainly! Here\'s a story possibility: The dog in IMAGE#2630, named Boomer, belongs to a dedicated pilot who often takes on long haul flights, like the airplane depicted in IMAGE#3065. Whenever the pilot sets off for a lengthy trip, Boomer retreats to his comfortable dog bed in front of the cozy fireplace, waiting for his owner\'s return.\n\nThe pilot, while often away, never fails to think about Boomer. From high above the earth, amidst beautiful sunsets, the canine\'s owner counts the hours until they can be reunited. Just the thought of Boomer's serene expression on the tastefully patterned dog bed provides a sense of peace and comfort even thousands of miles away.\n\nSo despite their geographical separation depicted in these images, there's an underlying connection of loyalty and companionship between Boomer and his pilot owner, which is a heartwarming staple in their world filled with frequent departures and heartfelt reunions.'},
     {'role': '@\color{darkcyan}{user}@', 'image_ids': ['4704'], 'content': 'How would the atmosphere in IMAGE#4704 contribute to the story of our pilot-owner and Boomer from IMAGE#2630 and IMAGE#3065?'},
     {'role': '@\color{forestgreen}{assistant}@', 'content': 'The setting in IMAGE#4704 could introduce another layer to our narrative. Imagine this: When off-duty, our pilot-owner is also an avid golfer who cherishes the peacefulness and serenity found on the golf course, like the one seen in IMAGE#4704. After returning from a long trip, the pilot usually heads to the fairway, and Boomer eagerly accompanies him.\n\nThe manicured greens serve as a perfect place for Boomer to roam freely, chase golf balls, and feel the wind in his fur. The distant city skyline seen from the golf course acts as a reminder of the urban life our pilot leaves behind when they find solace in the peaceful embrace of the game and their cherished canine companion.\n\nSubsequently, the lush green golf course fills in a missing piece of the puzzle, acting as a mediating space that connects the frequent travels of the pilot and Boomer\'s waiting. It becomes a place of joy, reconnection, and shared experiences between the duo, marking the happy ending to their waiting and the start of their time together until the next flight.'}], 
[
    {'role': '@\color{darkcyan}{user}@', 'image_ids': ['2439', '1093', '2071'], 'content': 'Could you help me design a scenario the place in IMAGE#1093 is being used as a temporary office for the man in IMAGE#2071, who usually works in the kitchen in IMAGE#2439?'},
    {'role': '@\color{forestgreen}{assistant}@', 'content': 'Sure, here\'s a possible scenario: In IMAGE#2071, we see a man who typically works from his spacious kitchen shown in IMAGE#2439. The kitchen, with its wooden cabinets and black countertops, doubles as his office. Its large windows provide ample natural light, creating a conducive and productive work environment. It is here that he plans his strategies, makes important business decisions, and leads virtual meetings.\n\nHowever, for a change of scenario - perhaps due to a home renovation or family visiting - the man decides to shift his workspace to the small room shown in IMAGE#1093 for a while. The room is likely a storage space or craft room with wrapping paper stacked on a wooden shelf, indicating creativity and resourcefulness.\n\nHe sets up his temporary office here, positioning his desk near the window to replicate the natural light he enjoys in his kitchen-office. The surrounding array of vibrant wrapping paper inspires creativity and brings a touch of liveliness to his work routine. Also, the closed door ensures that he has quiet and privacy for his business calls.\n\nSo in this unusual but charming workspace, our businessman continues his work, exhibiting that all you need to be productive is a bit of adaptability, irrespective of where you are.'},
    {'role': '@\color{darkcyan}{user}@', 'image_ids': ['2966'], 'content': 'Taking into account all the previous images, how could we incorporate the scene from IMAGE#2966 into the story of the man changing his workspace?'},
    {'role': '@\color{forestgreen}{assistant}@', 'content': 'The serene garden scene in IMAGE#2966 could be an excellent addition to the narrative. Let\'s say, towards the back of the house where both the kitchen-office and the temporary workspace reside, there's this quiet garden depicted in the image, featuring a stone statue sitting under a tree.\n\nAt the end of each workday, the man enjoys taking a stroll in the garden to unwind. The setting sun casts a warm orange light on the scene creating a peaceful atmosphere, and he often finds himself sitting next to the stone statue, reflecting on his day. The tranquility of the garden effectively helps him transition from work mode, and it provides him with a sense of balance that complements his adaptable work life.\n\nTherefore, despite the change in his workspace, the man maintains his evening ritual in the garden, capturing the idea that while our surroundings may change, there's comfort in keeping certain routines and finding peace in nature.'}
]
]
```
\end{lstlisting}

{ \section{Dialogue Demonstrations' contribution to data quality and diversity}\label{appendix:dialogue_demonstration_effect}
{\texttt{Dialogue Demonstrations}} act as contextual learning examples, guiding GPT-4 to produce responses that are both well-formatted and diverse. To demonstrate this effect, we modified the original \textbf{prompt} used for GPT-assisted Multiple Dialogues Generation, as detailed in \autoref{tab:prompt_dialogue_generation}, by removing content relating to demonstration dialogues. We then employed the same \textcolor{brickred}{\texttt{Candidate Image Descriptions}} as in \autoref{fig:candidate_images_in_single_demo_dialog} to create a new prompt and generate a response. The resulting response was inferior in quality, failing to meet the desired formatting criteria, such as assigning image IDs, specifying the number of images per dialogue turn, and incorporating new images in subsequent turns. Furthermore, the response lacked the diversity that its dialogues typically ask for more detailed descriptions of images but not specifying particular aspects. In conclusion, dialogue demonstrations are crucial not only for enhancing data quality by providing formatting guidelines but also for increasing diversity by conditioning different demonstrations.
}

% \begin{markdown}
% ---
\noindent\rule{\textwidth}{0.4pt} % Horizontal line
% **Dialogue 1:**  
\textbf{Dialogue 1:}
   
1. \textbf{User:} I'm working on a home renovation project and need some kitchen inspiration. The image with ID 2439 caught my eye; could you give a more detailed account of it?  
   
2. \textbf{Assistant:} Absolutely, I'd be happy to! The kitchen space shown in image 2439 exhibits a clean and bright style, with an airy feel. The cabinets, constructed of wood, have an earthy and natural charm. They complement the black countertops well, adding a layer of contrast. The appliances are white, offering a sense of cleanliness and simplicity.   
   The floor is tiled, a practical choice for a room that often grapples with spills and frequent foot traffic. A substantial window draws in an ample amount of natural light, brightening the room and making it appear spacious. The white paint on the walls enhances this light, reflecting it to confer an even more airy ambience.  
   This type of kitchen design is both efficient and welcoming, making it an ideal space for cooking and entertaining.  
   
3. \textbf{User:} Great, that was very helpful! One more thing - is there anything in this kitchen that could be improved or updated?  
   
4. \textbf{Assistant:} Well, any suggestions for improvement would largely depend on your personal style and functionality needs. However, here are a couple of general ideas:  
   Whilst the window lets in a large amount of natural light, at night the room might need more illumination. Consider upgrading the lighting fixtures for a brighter and more evenly lit kitchen after sunset. Depending on the design scheme, you could select modern pendant lights, under-cabinet lights, or recessed lights.   
   Additionally, the countertops appear to be clear of clutter, which is great, but there doesn't seem to be much in the way of storage solutions on them. Maybe you could use some stylish canisters for everyday utensils or ingredients, providing storage while keeping items within easy reach.   
   Remember, home renovation is personal and subjective, so it's always a good idea to take inspiration from different sources, but ultimately stick to what will make you and the other users of the kitchen most content and comfortable.  
   
% ---  
\noindent\rule{\textwidth}{0.4pt} % Horizontal line
   
\textbf{Dialogue 2: }
   
1. \textbf{User:} I'm curious about the photos with IDs 3065 and 2630. Could you explain these in more detail?  
   
2. \textbf{Assistant:} Sure thing! In image 3065, we see an impressive dramatization of human achievement and the beauty of nature. It captures a large commercial airplane in the midst of flight under a sunset. The aircraft, distinguished by a white body and red and blue stripes on the tail, is low flying in the sky. Rumbling engines can be seen at the bottom, belching smoke. The setting sun irradiates the left side of the image with a warm, orange glow, while the right side basks in a cooler hue of blue. The sky's depth is accentuated by its rich blue tones, and the distantly scattered pink-tinged clouds add a touch of romanticism to the scene.  
   Image 2630 is an adorable and intimate snapshot of a companion animal, a black and tan dog lying on a sizable gray dog bed. The bed's geometric pattern and substantial size suggest its comfort. Located before a brick fireplace, the setting seems cozy and warm. It appears as if the dog is sleeping, its snout resting atop its folded front legs. The clarity and lighting in the image allow us to appreciate the serene and comfortable atmosphere it represents.  
   
3. \textbf{User:} I find those details on image 3065 very helpful. I'm working on an art project that involves illustrating a story about a family dog. How can I incorporate elements from image 2630 into a scene involving a plane flying at sunset as in image 3065?  
   
4. \textbf{Assistant:} That sounds like a captivating art project! If you're trying to harmonize both images into one scene, here are a few suggestions:  
   You could keep the breathtaking sunset sky with the airplane from image 3065 as the background. It brings a sense of vastness and adventure. However, instead of showing the airplane up in the sky, you could bring it closer to the ground, perhaps just made it after a landing or just before taking off on a runway near the home with the dog.  
   In the foreground, you could portray a homely scene inspired by image 2630. The dog could be lying on its comfy bed, looking up at the airplane in the sky with curiosity, or perhaps excitement. To create a strong emotional connection, the placement of the dog and its bed could be strategic, maybe by a window that shows the plane in the sky, demonstrating that the two scenes are connected.  
   Finally, you might want to add elements that link the dog to the plane. Perhaps a family member waving goodbye in the distance near the plane, or a toy plane near the dog to show the connection between the domestic scene and the broader story. This could convey notions of longing, anticipation, or excitement.  
   
% ---  
\noindent\rule{\textwidth}{0.4pt} % Horizontal line
   
\textbf{Dialogue 3:}  
   
1. \textbf{User:} I've been assessing images 1438, 2630, and 3120 and would like some further clarification on them.   
  
2. \textbf{Assistant:} Of course, I'd be happy to help. In image 1438, we see the Eiffel Tower, an international symbol of Paris and France, standing tall. It's painted in blue, white, and beneath it runs a river. The tower's lights create a magical ambience that has made it a popular site for strolls and photos.   
   Image 2630 depicts a peaceful scene of a black and tan dog sleeping on a large, gray dog bed. The bed, sporting a geometric pattern, is placed in front of a brick fireplace, amplifying the homely and comfort vibes the picture transmits.  
   Image 3120 shows a distinctive logo for a company or brand. The logo comprises a ‘z’ and an ‘l’ in a red and black color scheme. The letters are uniquely connected by a diagonal line running through the image's center, rendering a modern and minimal design suitable for various businesses.  
   
3. \textbf{User:} I'm starting a pet-friendly accommodation hosting service in Paris and trying to come up with a logo. Could I blend elements from the images of the Eiffel Tower and the dog into the logo design like in image 3120?  
   
4. \textbf{Assistant:} Combining those elements sounds like a great idea! Here's a suggestion:  
   Use the 'z' and 'l' shape from the logo in image 3120, but perhaps integrate elements that signify your business's unique selling point, being pet-friendly and located in Paris. You could morph the 'z' into a simplified, abstract version of the Eiffel Tower, keeping the minimal and sharp edges. On the other hand, the 'l' could be transformed into an abstract, stylized rendition of a dog, keeping the balance between simplicity and recognition.  
   To make it clear that the accommodation is pet-friendly, consider incorporating paw prints as embellishments or as a background pattern.  
   The color scheme of red, white, and blue could be another subtle nod to your location in France, inheriting the theme from the Eiffel Tower's image.   
   Ultimately, the design should resonate with your target audience, so using familiar symbols (like the Eiffel Tower and a dog) in a clean, modern design could effectively communicate your brand's purpose and values.

% ---
\noindent\rule{\textwidth}{0.4pt} % Horizontal line

% \end{markdown}

\end{document}